\documentclass{article}

\usepackage[final,nonatbib]{neurips_2021}

\usepackage[utf8]{inputenc}
\usepackage[T1]{fontenc}
\usepackage{hyperref}
\usepackage{url}
\usepackage{booktabs}
\usepackage{amsfonts}
\usepackage{nicefrac}
\usepackage{microtype}
\usepackage[usenames,dvipsnames,table]{xcolor}

\usepackage[numbers]{natbib}
\title{Recursive Bayesian Networks:\\Generalising and Unifying\\Probabilistic Context-Free Grammars and\\Dynamic Bayesian Networks}

\author{%
  Robert Lieck\thanks{corresponding author, code at \url{https://github.com/robert-lieck/RBN}} \\
  Digital and Cognitive Musicology Lab \\
  École Polytechnique Fédérale de Lausanne \\
  1015 Lausanne, Switzerland \\
  \texttt{research@robert-lieck.com}
  \And
  Martin Rohrmeier \\
  Digital and Cognitive Musicology Lab \\
  École Polytechnique Fédérale de Lausanne \\
  1015 Lausanne, Switzerland \\
  \texttt{martin.rohrmeier@epfl.ch}
}

\usepackage[utf8]{inputenc}

\usepackage[]{graphicx}
\usepackage{tikz}
\usetikzlibrary{
  arrows,
  arrows.meta,
  calc,
  math,
  positioning,
  backgrounds,
  graphs,
  shapes.geometric,
  fit,
  decorations,
  decorations.markings,
  bayesnet,
}
\usepackage{pgfplots}
\usepgfplotslibrary{ternary}
\usepackage[export]{adjustbox}

\usepackage{pifont}
\newcommand{\xmark}{\ding{53}}%

\usepackage{tcolorbox}

\usepackage{xparse}
\usepackage{relsize}
\usepackage{enumitem}

\usepackage{longtable}
\usepackage{booktabs}
\newcommand{\otoprule}{\midrule[\heavyrulewidth]}
\usepackage{multirow}

\usepackage{url}

\usepackage{minitoc}
\usepackage{amsmath,amsfonts,amssymb,mathtools,centernot,amsthm}
\usepackage{bbold}

\usepackage[
acronym,
shortcuts,
]{glossaries}
\glsdisablehyper
\MFUhyphentrue
\newcommand*{\titleac}[1]{\glsentrytitlecase{#1}{long}}
\newcommand*{\titleacp}[1]{\glsentrytitlecase{#1}{longpl}}

\allowdisplaybreaks

\NewDocumentCommand{\incoming}{ O{} O{0.6} m }{\draw[edge,<-,densely dotted,#1] (#3) -- +(0,#2);}
\NewDocumentCommand{\outgoing}{ O{} O{-0.6} m }{\draw[edge,->,densely dotted,#1] (#3) -- +(0,#2);}

\NewDocumentCommand{\vcenterbox}{ m }{\raisebox{-0.5\height}{#1}}

\newcommand*{\NonTermVar}{x}
\newcommand*{\TermVar}{y}
\newcommand*{\StrucVar}{z}
\newcommand*{\NonTermOrTermVar}{v}

\newcommand*{\SetNonTermVars}{\mathcal{X}}
\newcommand*{\SetTermVars}{\mathcal{Y}}
\newcommand*{\SetStrucVars}{\mathcal{Z}}

\newcommand*{\NonTermVarBlock}{\vec{X}}
\newcommand*{\TermVarBlock}{\vec{Y}}
\newcommand*{\StrucVarBlock}{\vec{Z}}

\newcommand*{\SetTransDist}{\mathcal{T}}
\newcommand*{\SetStrucDist}{\mathcal{S}}
\newcommand*{\TransDist}{\tau}

\newcommand*{\PriorLabel}{\mathrm{P}}
\newcommand*{\NonTermLabel}{\mathrm{N}}
\newcommand*{\TermLabel}{\mathrm{T}}
\newcommand*{\StrucLabel}{\mathrm{S}}

\newcommand*{\PriorDist}{p_\PriorLabel}
\newcommand*{\NonTermTrans}{p_\NonTermLabel}
\newcommand*{\TermTrans}{p_\TermLabel}
\newcommand*{\StrucDist}{p_\StrucLabel}

\NewDocumentCommand{\Inside}{ o o }{%
  \IfNoValueTF{#1}{%
    \beta
  }{%
    \IfNoValueTF{#2}{%
      \beta(\NonTermVar_{#1})
    }{%
      \beta(#2_{#1})
    }%
  }%
}
\NewDocumentCommand{\Outside}{ o o }{%
  \IfNoValueTF{#1}{%
    \alpha
  }{%
    \IfNoValueTF{#2}{%
      \alpha(\NonTermVar_{#1})
    }{%
      \alpha(#2_{#1})
    }%
  }%
}

\newcommand*{\TermProb}{p_{\text{term}}}

\newcommand*{\TransMat}{T}
\newcommand*{\TransVar}{\tau}

\newcommand*{\Arity}{\eta}

\NewDocumentCommand{\sumint}{ O{} O{} O{-0.3em} }{%
  \mathop{\text{\makebox[0.5\width]{\phantom{$\sum$}}} \mathclap{\sum}\mathclap{\;\int_{\text{\raisebox{#3}{$\mathrlap{#1}$}}}^{\mathrlap{#2}}} \text{\makebox[0.5\width]{\phantom{$\sum$}}}}%
}

\newcommand*{\multisum}{\mathop{\sum\hspace{-1ex}\cdots\hspace{-0.5ex}\sum}}

\NewDocumentCommand{\shortminus}{ O{0.6} }{\text{\scalebox{#1}[1]{$-$}}}
\NewDocumentCommand{\shortplus}{ O{0.6} }{\raisebox{#1pt}{\text{\scalebox{#1}{$+$}}}}

\DeclarePairedDelimiter\parens{\lparen}{\rparen}
\DeclarePairedDelimiter\brackets{\lbrack}{\rbrack}

\DeclarePairedDelimiter\bars{|}{|}

\renewcommand{\det}{\bars}

\newcommand*{\NoIndentAfterEnv}[1]{%
  \AfterEndEnvironment{#1}{\noindent\ignorespaces}%
}

\theoremstyle{indented}

\newtheorem{theorem}{Theorem}

\newtheorem{condition}{Condition}
\newtheorem{subcondition}{Condition}[condition]

\newtheorem{definition}{Definition}
\newtheorem{subdefinition}{Definition}[definition]

\NoIndentAfterEnv{definition}
\NoIndentAfterEnv{subdefinition}

\NoIndentAfterEnv{example}

\makeatletter
\NewDocumentCommand{\Label}{ o }{%
  \IfNoValueTF{#1}{%
    &\refstepcounter{equation}(\theequation)&%
  }{%
    &\refstepcounter{equation}(\theequation)\ltx@label{#1}&%
  }%
}
\makeatother

\theoremstyle{proof}
\newtheorem*{myproof}{Proof}
\renewenvironment*{proof}{\begin{myproof}}{\hfill\qed\end{myproof}}
\NoIndentAfterEnv{proof}

\newacronym{sgd}{SGD}{stochastic gradient descent}
\newacronym{mcmc}{MCMC}{Markov chain Monte-Carlo}
\newacronym{hmc}{HMC}{Hamiltonian Markov chain Monte-Carlo}
\newacronym{hmm}{HMM}{hidden Markov model}
\newacronym{cyk}{CYK}{Cocke-Younger-Kasami}
\newacronym{cfg}{CFG}{context-free grammar}
\newacronym{rg}{RG}{regular grammar}
\newacronym{pcfg}{PCFG}{probabilistic context-free grammar}
\newacronym{bn}{BN}{Bayesian network}
\newacronym{dbn}{DBN}{dynamic Bayesian network}
\newacronym{rbn}{RBN}{Recursive Bayesian Network}
\newacronym{grbn}{GRBN}{Gaussian RBN}
\newacronym{map}{MAP}{maximum posterior}
\newacronym{cnf}{CNF}{Chomsky normal form}
\newacronym{ml}{ML}{maximum likelihood}
\newacronym{kld}{KLD}{Kullback-Leibler divergence}
\newacronym{cpd}{CPD}{change point detection}
\newacronym{pcd}{PCD}{pitch-class distribution}
\newacronym{gm}{GM}{Gaussian mixture}
\newacronym[longplural={Gaussian processes}]{gp}{GP}{Gaussian process}
\newacronym{hc}{HC}{hierarchical clustering}
\newacronym{nn}{NN}{neural network}
\newacronym{spn}{SPN}{sum-product network}
\newacronym{fgg}{FGG}{factor graph grammars}
\newacronym{lvg}{LVG}{latent variable grammar}
\newacronym{lveg}{LVeG}{latent vector grammar}

\newcommand*{\figref}[1]{Figure~\ref{fig:#1}}
\newcommand*{\tabref}[1]{Table~\ref{tab:#1}}

\newcommand*{\secref}[1]{Section~\ref{sec:#1}}
\newcommand*{\appref}[1]{Appendix~\ref{app:#1}}

\newcommand{\setvars}[2]{
  \pgfmathsetmacro{\c}{(#1 + #2)/2}
  \pgfmathsetmacro{\cc}{sqrt(2)*\c}
  \pgfmathsetmacro{\sqrttwo}{sqrt(2)/2}
  \pgfmathsetmacro{\w}{#2 - #1}
  \pgfmathsetmacro{\ww}{sqrt(2)*\w/2}
}

\newcommand*{\colA}[1]{{\color{BrickRed}#1}}
\newcommand*{\colB}[1]{{\color{OliveGreen}#1}}
\newcommand*{\colC}[1]{{\color{RoyalBlue}#1}}

\definecolor{myblue}{HTML}{2174e6}

\NewDocumentCommand{\newlineAlign}{ s O{10mm} O{} }{%
  \IfBooleanTF{#1}{%
    #3\\&#3\hspace*{#2}%
  }{%
    \ldots#3\\&#3\hspace*{#2}\ldots%
  }%
}

\newcommand*{\pprime}{{\prime\prime}}
\newcommand*{\ppprime}{{\prime\prime\prime}}

\newcommand*{\DKL}{D_{\mathrm{KL}}}

\NewDocumentCommand{\Norm}{ s o m m m O{;} O{,} }{%
  \IfBooleanTF{#1}{%
    \IfNoValueTF{#2}{%
      \mathcal{N}\parens*{#3#6#4#7#5}%
    }{%
      \mathcal{N}\parens*[#2]{#3#6#4#7#5}%
    }%
  }{%
    \IfNoValueTF{#2}{%
      \mathcal{N}\parens{#3#6#4#7#5}%
    }{%
      \mathcal{N}\parens[#2]{#3#6#4#7#5}%
    }%
  }%
}

\newcommand*{\cond}{\mathop{|}}

\newcommand*{\ccond}{\mathop{\|}}

\newcommand*{\expectcustom}[4][]{#1#2[\hspace{#3}#2[#4#2]\hspace{#3}#2]}
\newcommand*{\expect}[2][]{\expectcustom[#1]{}{-0.15em}{#2}}
\newcommand*{\expectbig}[2][]{\expectcustom[#1]{\big}{-0.29em}{#2}}
\newcommand*{\expectBig}[2][]{\expectcustom[#1]{\Big}{-0.32em}{#2}}
\newcommand*{\expectbigg}[2][]{\expectcustom[#1]{\bigg}{-0.38em}{#2}}
\newcommand*{\expectBigg}[2][]{\expectcustom[#1]{\Bigg}{-0.42em}{#2}}

\NewDocumentCommand{\iverson}{ m O{1} }{\expect{\scalebox{#2}{$#1$}}}
\NewDocumentCommand{\iversonbig}{ m O{1} }{\expectbig{\scalebox{#2}{$#1$}}}
\NewDocumentCommand{\iversonBig}{ m O{1} }{\expectBig{\scalebox{#2}{$#1$}}}
\NewDocumentCommand{\iversonbigg}{ m O{1} }{\expectbigg{\scalebox{#2}{$#1$}}}
\NewDocumentCommand{\iversonBigg}{ m O{1} }{\expectBigg{\scalebox{#2}{$#1$}}}

\renewcommand*{\vec}[1]{\boldsymbol{\mathbf{#1}}}

\begin{document}
\doparttoc
\faketableofcontents
\part{}

\maketitle

\begin{abstract}
  \Acp{pcfg} and \acp{dbn} are widely used sequence models with complementary strengths and limitations.
  While \acp{pcfg} allow for nested hierarchical dependencies (tree structures), their latent variables (non-terminal symbols) have to be discrete. In contrast, \acp{dbn} allow for continuous latent variables, but the dependencies are strictly sequential (chain structure).
  Therefore, neither can be applied if the latent variables are assumed to be continuous and \emph{also} to have a nested hierarchical dependency structure.
  In this paper, we present \acp{rbn}, which
  generalise and unify \acp{pcfg} and \acp{dbn}, combining their strengths and containing both as special cases.
  \Acp{rbn} define a joint distribution over tree-structured Bayesian networks with
  discrete or continuous
  latent variables.
  The main challenge lies in performing \emph{joint} inference over the exponential number of possible structures and the continuous variables.
  We provide two solutions: 1)~For arbitrary \acp{rbn}, we generalise inside and outside probabilities from \acp{pcfg} to the mixed discrete-continuous case, which allows for \acl{map} estimates of the continuous latent variables via gradient descent, while marginalising over network structures.
  2)~For Gaussian \acp{rbn}, we additionally derive an analytic approximation of the marginal data likelihood (evidence) and marginal posterior distribution, allowing for robust parameter optimisation and Bayesian inference.
  The capacity and diverse applications of \acp{rbn} are illustrated on two examples:
  In a quantitative evaluation on synthetic data, we demonstrate and discuss the advantage of \acp{rbn} for segmentation and tree induction
  from noisy sequences, compared to \acl{cpd} and \acl{hc}.
    In an application to musical data, we approach the unsolved problem of hierarchical music analysis from the raw note level and compare our results to expert annotations.
\end{abstract}

\begin{figure}[tp!]
  \centering
  \scalebox{0.83}{
    \begin{tikzpicture}
      \tikzstyle{arrow}=[->, >=stealth]
      \node[inner sep=0pt] (box) {
        \begin{tabular}{ccc|cc}
          &&& \multicolumn{2}{c}{\textbf{Structure\hspace*{1em}}} \\[1mm]
          &&& Linear/Chain & Hierarchical/Tree \\
          &&& \includegraphics[scale=0.05]{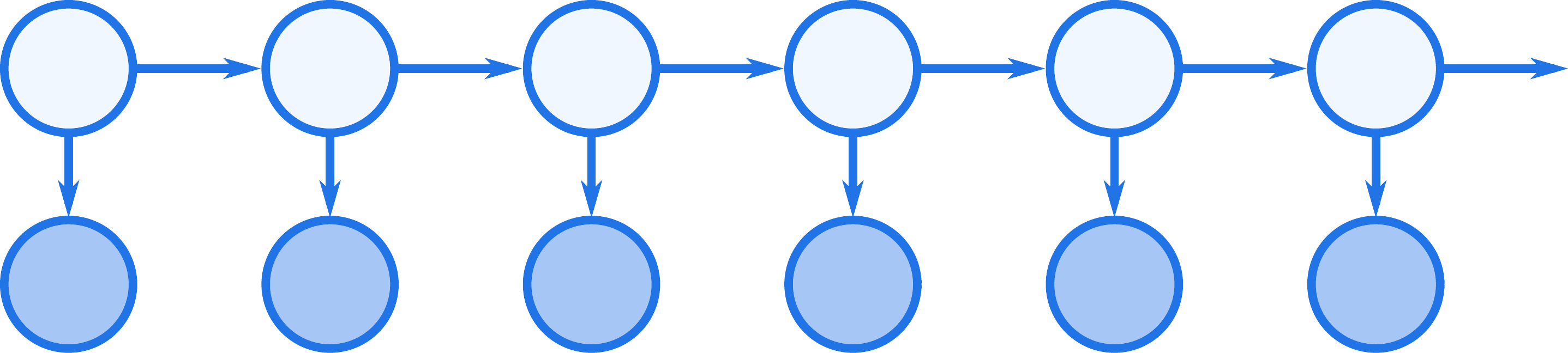} & \includegraphics[scale=0.05]{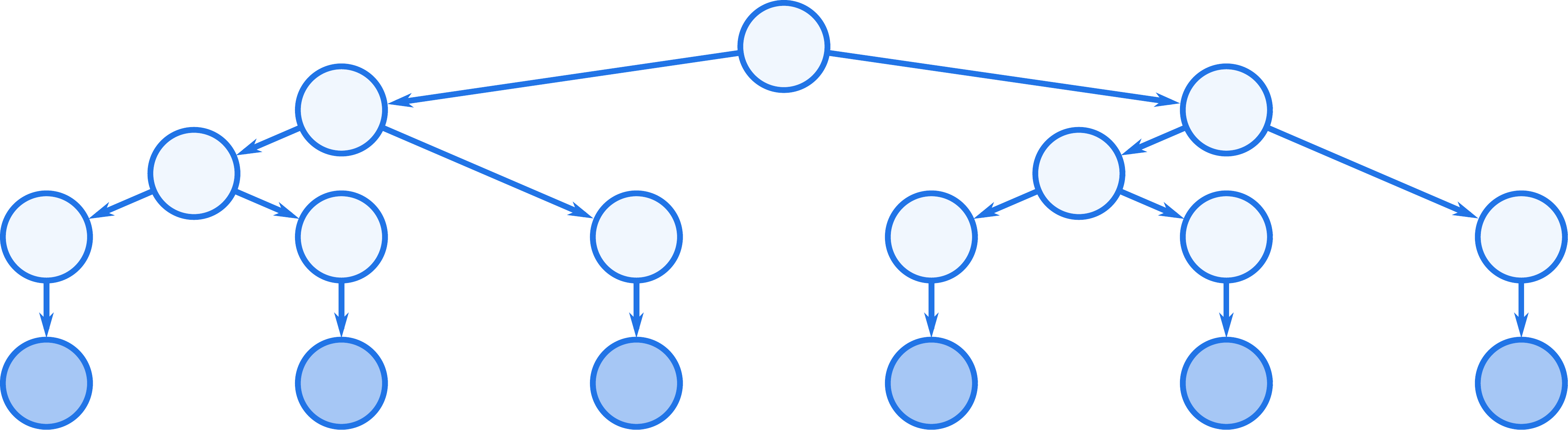} \\
          \hline
          \multirow{2}{1mm}{\rotatebox[origin=c]{90}{\textbf{Variables\hspace*{1em}}}}
          & \rotatebox[origin=c]{90}{\parbox{16mm}{\centering Only\\Discrete}} & \raisebox{-0.35\height}[0pt][0pt]{\includegraphics[scale=0.05]{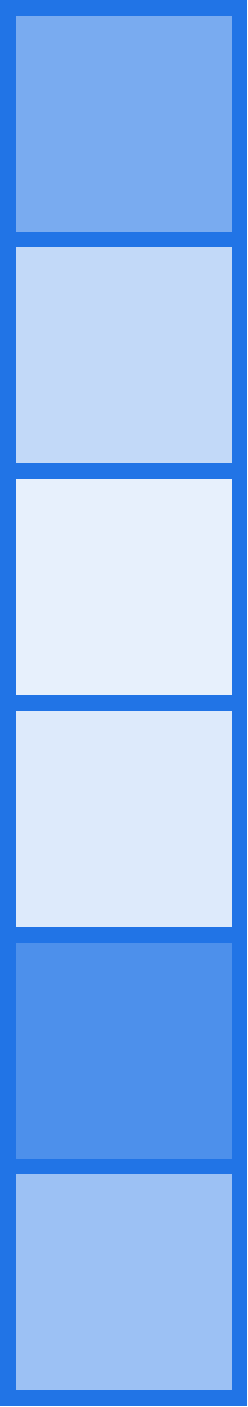}} & \parbox{3.5cm}{\centering \titleacp{rg}/\\\titleacp{hmm}
          } & \parbox{3.5cm}{\centering \titleacp{pcfg}}\\
          & \rotatebox[origin=c]{90}{\parbox{16mm}{\centering Discrete/\\Continuous}} & \raisebox{-0.5\height}[0pt][0pt]{\includegraphics[scale=0.05,angle=90]{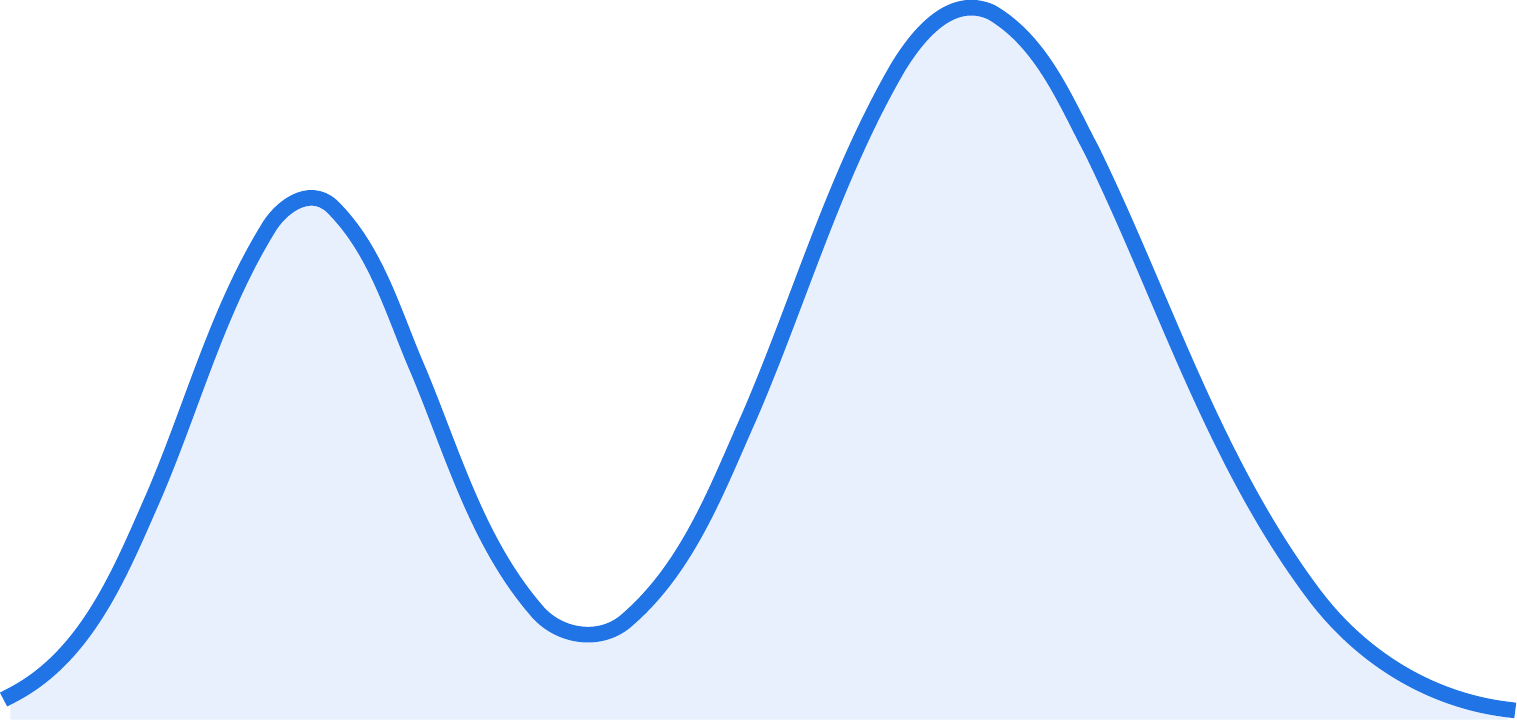}} & \parbox{3.3cm}{\centering Dynamic\\Bayesian Networks} & \parbox{3.8cm}{
                                                                                                                                                                                                                \begin{tcolorbox}[width=\linewidth,colframe={myblue},colback={white},opacityback=0.97,arc=1mm,boxrule=0.3mm,left=0mm,right=0mm,top=0mm,bottom=0mm,boxsep=4mm]
                                                                                                                                                                                                                \centering\textbf{Recursive\\Bayesian Networks}
                                                                                                                                                                                                                \end{tcolorbox}
          }
        \end{tabular}
      };
    \end{tikzpicture}
  }
  \caption{\Acsp{rbn} generalise \acsp{pcfg} by allowing for continuous latent variables and \acsp{dbn} by incorporating nested hierarchical dependencies.
  }
  \label{fig:q3aTdiGF}
\end{figure}%

\section{Introduction}

Long-term dependencies with a nested hierarchical structure are one of the major challenges in modelling sequential data. This type of dependencies is common in many domains, such as natural language \citep{jurafsky2000}, music \citep{lerdahl1983,rohrmeier2020}, or decision making \citep{barto03,ghallab_automated_2016}.
Two of the most widely used probabilistic models for sequential data are \glsreset{pcfg}\acp{pcfg} and \glsreset{dbn}\acp{dbn},
both having complementary strengths.

\Acp{pcfg} are well-established and widely used for modelling hierarchical long-term dependencies in symbolic data \citep{jurafsky2000,grune2007,geib2009,rohrmeier2011,jacquemard2015,ghallab_automated_2016,harasimGeneralized2018}.
They generalise local (Markov) transition models by
allowing for infinitely many levels of nested hierarchical dependencies and a flexible number of latent variables.
However, parsing methods such as the \ac{cyk} algorithm \citep{kasami1966efficient,younger1967recognition,goodman1999,manning1999,grune2007} rely on the discrete nature of the rules and variables.

In contrast, \acp{dbn} are sequential models with a fixed set of random variables that reoccur at each time step \citep{murphy2002,koller_probabilistic_2009}. The variables at each time step may be discrete or continuous, latent or observed, and may have an arbitrary non-cyclic dependency structure among each other, with additional links from the previous and to the next time slice. They comprise important model classes as special cases, such as \acp{hmm} if there is only a single discrete latent variable or linear dynamical systems
if all dependencies are linear Gaussians \citep{welch1995,bishop07,koller_probabilistic_2009}. However, \acp{dbn} only allow for a fixed chain of Markov dependencies between time slices and cannot represent nested hierarchical structures.

In this paper, we present \glsreset{rbn}\acp{rbn}, a novel class of probabilistic models that combines the strengths of \acp{pcfg} and \acp{dbn} by allowing for nested hierarchical dependencies in combination with arbitrary discrete or continuous random variables (\figref{q3aTdiGF}). Our main contributions are as follows:
\begin{enumerate}[nosep,itemsep=0.2em]
\item With \acp{rbn}, we provide a unified theoretical framework for a large class of important sequence models, including \acp{pcfg} and \acp{dbn}.
\item We generalise inside and outside probabilities from \acp{pcfg} to continuous latent variables, allowing for \ac{map} inference in arbitrary \acp{rbn}.
\item For \aclp{grbn}, we derive an analytic approximation for the marginal likelihood and marginal posterior distribution, allowing for robust parameter optimisation and Bayesian inference.
\item We provide a quantitative evaluation on synthetic data and an application to the challenging task of hierarchical music analysis.
\end{enumerate}

\subsection{Related Work}
\label{sec:related_work}

\Acp{pcfg} have a long tradition for modelling nested hierarchical dependencies in symbolic data with a variety of parsing algorithms for inferring the structure and variables' values \citep{goodman1999,grune2007,manning1999}. Beyond their application to sequential data, \acp{pcfg} have been generalised to graph structures \citep{drewes1997,engelfriet1997,golin1991,rozenberg1997}, which readily transfers to applications of \acp{rbn}.
\Acp{lveg} \citep{zhao2018} are an extension of \acp{lvg} \citep{socher2013a,cohen2017} with continuous latent states. As for \acp{rbn}, approximate parsing is possible in the Gaussian case. However, both \acp{lvg} and \acp{lveg} are special cases of \acp{rbn} and do not draw the connection to graphical models.
More recently, the availability of automatic differentiation libraries, such as \texttt{PyTorch} \citep{paszke2019}, has lead to a number of applications where gradients are propagated through the entire parsing process \citep{eisner2016,kim2019,zhang-etal-2020-fast,rush2020}.

The process of parsing a \ac{pcfg} or \ac{rbn} can be formally rewritten as a \acl{spn} \citep[\acs{spn}\glsunset{spn};][]{poon2011,molina2018,shao2020}.
\Acl{fgg} \citep[\acsp{fgg}\glsunset{fgg};][]{chiang2020} generalise \acp{pcfg}, case-factor diagrams \citep{mcallester2008} and \acp{spn} by using a hyperedge replacement graph grammar \citep{drewes1997} to describe a distribution over graph structures that is more general than that of \acp{rbn} (not only trees). However, none of the approaches addresses the problem of inference with continuous variables that we are facing in \acp{rbn} (exponentially many terms with exponentially many nested integrals).

A wide range of probabilistic and neural models operate with a fixed graphical structure and are loosely related to \acp{rbn}.
Hidden tree Markov models \citep{diligenti2003,bacciu2012} generalise \acp{hmm} from chain to fixed tree structures. They model data at each node as observations of a latent Markov process on the underlying tree, which is part of the input data. Additionally estimating the underlying tree structure has been addressed in \citep{anandkumar2011,choi2011,huang2020}.
Recursive neural tensor networks \citep{socher2013} use a \ac{pcfg} for parsing a given sequence of symbols to obtain a tree structure, which is then fixed and used as the backbone for a \acl{nn}.
More generally, there is a number of methods for inferring a fixed structure for graphical models \citep{heckerman1995,koller_probabilistic_2009,heckerman2013,drton2017}, \acp{spn} \citep{gens2013,lee2013,rooshenas2014,vergari2015,molina2018,vergari2019}, or graph \aclp{nn} \citep{franceschi2019,jin2020,yu2019}.
All these methods have in common that a fixed structure is either given or estimated but not treated in a probabilistic Bayesian manner.

Some approaches attempt a Bayesian treatment of the unknown structure of a graphical model or \ac{spn} via dynamic programming \citep{dash2004,meila2006} or \acl{mcmc} sampling \citep{grzegorczyk2008,eaton2007,trapp2019}. However, the structure is assumed to be independent of the latent variables (they only become dependent \emph{conditional} on the data) and
the latent variables cannot be used to \emph{control} the structure, as it is the case in \acp{rbn}. The challenge of continuous variables also remains unsolved.

\begin{figure}[t!]
  \begin{minipage}[b]{0.35\linewidth}
    \tikzstyle{var}=[circle,draw,fill=white,inner sep=0pt,minimum size=22pt]%
    \tikzstyle{obs}=[fill=black!5]%
    \tikzstyle{edge}=[->,>=stealth]%
    \tikzstyle{plate}=[draw,rounded corners,inner sep=13pt]%
    \tikzstyle{gate}=[draw,dashed,inner sep=5pt]%
    \tikzstyle{factor}=[draw,fill=black,inner sep=0pt,minimum size=5pt]%
    \tikzstyle{con}=[-*,shorten >=-2.5pt]%
    \def\scale{1.15}%
    \newcommand*{\plateindex}[2][1]{\scalebox{#1}{${}_{#2}$}}%
    \newcommand*{\rbncellSQ}{%
      \raisebox{-0.5\height}{\scalebox{\scale}{%
          \def\xscale{1.4}%
          \def\xscalecm{1.4cm}%
          \def\yscale{1.4}%
          \def\yscalecm{1.4cm}%
          \begin{tikzpicture}[xscale=\xscale,yscale=\yscale]%
            \node[var] (parent) at (0,0) {$\NonTermVar$};
            \node[var] (sigma) at (-1,0) {$\StrucVar$};
            \begin{scope}[xshift=-0.5cm]
              \node[var] (left_child) at (0,-1) {$\NonTermVar^\prime$};
              \node[var] (right_child) at (1,-1) {$\NonTermVar^\pprime$};
              \node[factor] (fac) at ($0.5*(left_child)+0.5*(right_child)+(0,0.3)$) {};
              \node[var,obs] (struc) at (2,-1) {$\TermVar$};
              \node[gate, anchor=west, minimum height=\yscalecm, minimum width=3*\xscalecm] (x_gate) at (-0.5,-1) {};
              \node[anchor=east, minimum height=\yscalecm, minimum width=\xscalecm] (y_gate) at (x_gate.east) {};
              \draw[gate] (y_gate.north west) -- (y_gate.south west);
              \node[anchor=south west] at (x_gate.south west) {\plateindex[0.8]{\StrucVar=\NonTermLabel}};
              \node[anchor=south west] at (y_gate.south west) {\plateindex[0.8]{\StrucVar=\TermLabel}};
            \end{scope}
            \incoming[][0.42]{parent}
            \outgoing[][-0.5]{left_child}
            \outgoing[][-0.5]{right_child}
            \draw[edge] (parent) -| (struc);
            \draw[edge] (parent) -- (fac);
            \draw[edge] (fac) -- (left_child);
            \draw[edge] (fac) -- (right_child);
            \draw[edge] (parent) -- (sigma);
            \draw[edge,con] (sigma) -- (x_gate.north west);
          \end{tikzpicture}%
        }}}%
    \hspace{-2ex}\rbncellSQ{}%
    \caption{\Ac{rbn} in \acl{cnf}. The
      non-terminal transition $\NonTermTrans(\NonTermVar^\prime, \NonTermVar^\pprime \cond \NonTermVar)$ and
      terminal transition $\TermTrans(\TermVar \cond \NonTermVar)$ are grouped into an \acs{rbn} cell
      with a structural distribution $\StrucDist(\StrucVar \cond \NonTermVar)$. We use gates \citep{minka2009} to describe structural distributions and extended factor graph notation \citep[black squares;][]{frey2003} for conditional joint distributions. Considering all possible ways how $n$ observations can be generated by recursively applying the \ac{rbn} cell produces an \acs{rbn} chart, as shown in \figref{zE70w8gY}.}
    \label{fig:vc9MBEI5}
  \end{minipage}
\hspace{1ex}
    \begin{minipage}[b]{0.63\linewidth}
    \def\n{7}
    \centering
    \tikzstyle{var}=[circle,draw,black!15,inner sep=0pt,minimum size=22pt]
    \tikzstyle{edge}=[->,>=stealth]
    \tikzstyle{emph}=[solid,RoyalBlue,line width=1pt,opacity=1]
    \tikzstyle{emphnode}=[emph,fill opacity=0,text=black,text opacity=1]
    \tikzstyle{box}=[black!10]
    \tikzstyle{emphbox}=[box,BurntOrange,opacity=1,line width=1pt]
    \tikzstyle{obsbox}=[emphbox,BrickRed,draw,minimum size=1.2cm]
    \newcommand*{\dosplit}[3]{
      \node[var,emphnode] at (alpha_#1_#3) {$\NonTermVar_{#1:#3}$};
      \draw[edge,emph] (alpha_#1_#3) -- (alpha_#1_#2);
      \draw[edge,emph] (alpha_#1_#3) -- (alpha_#2_#3);
    }
    \newcommand*{\doterm}[2]{
      \pgfmathsetmacro{\xstart}{int(#1+1)}
      \foreach \x in {\xstart,...,#2} {
        \draw[edge,emph] (alpha_#1_#2) -- (s\x);
        \node[var,emphnode] at (s\x) {};
      }
    }
    \newcommand*{\singleterm}[1]{
      \pgfmathsetmacro{\start}{int(#1-1)}
      \node[var,emphnode] at (alpha_\start_#1) {$\NonTermVar_{\start:#1}$};
      \node[var,emphnode] at (s#1) {$\TermVar_{#1}$};
      \draw[edge,emph] (alpha_\start_#1) -- (s#1);
    }
    \newcommand*{\drawsquare}{($(\cc-\sqrttwo, \ww)$) -- ($(\cc, \ww+\sqrttwo)$) -- ($(\cc+\sqrttwo, \ww)$) -- ($(\cc, \ww-\sqrttwo)$) -- cycle}
    \scalebox{0.9}{
      \begin{tikzpicture}[scale=1]
        \pgfmathsetmacro{\startend}{int(\n-1)}
        \foreach \s in {0,...,\startend} {
          \pgfmathsetmacro{\endstart}{int(\s+1)}
          \foreach \e in {\endstart,...,\n} {
            \setvars{\s}{\e}
            \node[var] (alpha_\s_\e) at (\cc, \ww) {$\NonTermVar_{\s:\e}$};
            \draw[box] \drawsquare;
          }
        }
        \foreach \s in {0,...,\startend} {
          \pgfmathsetmacro{\e}{int(\s+1)}
          \setvars{\s}{\e}
          \node[var] (s\e) at (\cc, \ww-1.5) {$\TermVar_{\e}$};
        }
        \foreach \x in {1,...,3} \node[obsbox] at (s\x) {};
        \foreach \e in {1,...,\n} {
          \setvars{0}{\e}
        }
        \foreach \s in {0,...,\startend} {
          \setvars{\s}{\n}
        }
        \pgfmathsetmacro{\e}{\n/2+0.5}
        \setvars{0}{\e}
        \pgfmathsetmacro{\s}{\n/2-0.5}
        \setvars{\s}{\n}
        \foreach \s/\e in {0/1, 0/2, 1/2, 1/3, 2/3} {
          \setvars{\s}{\e}
          \draw[emphbox] \drawsquare;
        }
        \dosplit{0}{1}{3}
        \dosplit{0}{3}{\n}
        \dosplit{1}{2}{3}
        \dosplit{3}{4}{5}
        \dosplit{3}{5}{\n}
        \dosplit{5}{6}{\n}
        \foreach \x in {1,...,\n} {
          \singleterm{\x}
        }
      \end{tikzpicture}
    }\\[-2mm]
    \caption{Chart of an \ac{rbn} in \ac{cnf} for sequential data of length $n=\n$. The network obtained by fixing one specific dependency structure is highlighted in blue; latent non-terminal variables that are not part of this particular structure are shown in grey. Orange and red boxes, respectively, indicate the subsets {\color{BurntOrange}$\NonTermVarBlock_{0:3}$} and {\color{BrickRed}$\TermVarBlock_{0:3}$} of latent non-terminal and observed terminal variables generated from $\NonTermVar_{0:3}$.
    }
    \label{fig:zE70w8gY}
  \end{minipage}
\end{figure}

\section{\Aclp{rbn}}

\Acp{rbn} are \emph{template-based} graphical models that define a joint distribution over network structures and variables' values. The number of template variables is fixed, but the number of instantiated variables, their connectivity and values are governed by the joint distribution.
As a rough analogy, \acp{rbn} can be thought of as \acp{dbn} that can not only be connected linearly to form a chain but also hierarchically to form a tree structure. Alternatively, they can be thought of as a \ac{pcfg} in which each symbol is a (possibly continuous) random variable.

\subsection{Definition}
\label{sec:definition}

\Acp{rbn} have three types of template variables: 1)~latent non-terminal variables (discrete or continuous), 2)~observed terminal variables (discrete or continuous), and 3)~latent structural variables (always discrete). In the simplest case, illustrated in \figref{vc9MBEI5}, an \ac{rbn} has one template variable of each type. Formally, an \ac{rbn} is defined as follows:
\begin{definition}[\titleac{rbn}]
  An \ac{rbn} is a tuple $(\SetNonTermVars, \SetTermVars, \SetStrucVars, \SetTransDist, \SetStrucDist, \PriorDist)$ with
  \begin{align}
    \SetNonTermVars \colon& \text{a set of latent non-terminal template variables} \\
    \SetTermVars \colon& \text{a set of observed terminal template variables} \\
    \SetStrucVars \colon& \text{a set of latent structural template variables, paired up with the non-terminal variables} \\
    \SetTransDist \colon& \text{a set of transition distributions } p(\NonTermOrTermVar_1,\ldots,\NonTermOrTermVar_\Arity \cond \NonTermVar) \text{ from a single non-terminal variable } \NonTermVar\in\SetNonTermVars \nonumber \\
    & \text{to a set of non-terminal and/or terminal variables } \NonTermOrTermVar_1,\ldots,\NonTermOrTermVar_\Arity\in\SetNonTermVars\cup\SetTermVars \label{eq:D1TXzdSD} \\
    \SetStrucDist \colon& \text{a set of structural distributions } p(\StrucVar \cond \NonTermVar) \text{, one for each non-terminal/structural pair} \\
    \PriorDist \colon& \text{a prior/start distribution for exactly one non-terminal variable.}
  \end{align}%
  The cardinality of a structural variable $\StrucVar\in\SetStrucVars$ corresponds to the number of possible transitions from the associated non-terminal variable $\NonTermVar\in\SetNonTermVars$; $\Arity$ in \eqref{eq:D1TXzdSD} is called the arity of the transition.
\end{definition}
Generating with an \ac{rbn} is straightforward. We start by sampling the value of the first non-terminal variable $\NonTermVar$ from the prior distribution $\PriorDist(\NonTermVar)$ and then repeat the following steps until no unprocessed non-terminal variables are left:
\begin{enumerate}
\item sample the value of the associated structural variable from $p(\StrucVar \cond \NonTermVar)$
\item choose a transition distribution $p(\NonTermOrTermVar_1,\ldots,\NonTermOrTermVar_\Arity \cond \NonTermVar)$ based on the structural variable's value
\item sample the variables $\NonTermOrTermVar_1,\ldots,\NonTermOrTermVar_\Arity$ from the transition distribution
\item for all newly generated non-terminal variables, go to step 1.
\end{enumerate}
The major challenge and focus of this paper is to perform joint inference over the latent structure and non-terminal variables' values conditional on a given set of observations.

\textbf{Chomsky Normal Form:}
In the simplest non-trivial case, an \ac{rbn} has one latent non-terminal, one observed terminal, and one latent structural template variable, with one non-terminal transition of arity $\Arity=2$ and one terminal transition of arity $\Arity=1$, as illustrated in Figures~\ref{fig:vc9MBEI5} and~\ref{fig:zE70w8gY}. It is defined by four distributions
\begin{align}
  \PriorDist(\NonTermVar) :{}& \text{prior/start distribution} \Label[eq:3Ejwxtzm]
  &
    \NonTermTrans(\NonTermVar^\prime, \NonTermVar^\pprime \cond \NonTermVar) :{}& \text{non-terminal transition} \\
  \TermTrans(\TermVar \cond \NonTermVar) :{}& \text{terminal transition} \Label
  &
    \StrucDist(\StrucVar \cond \NonTermVar) :{}& \text{termination probability}~. \label{eq:CJPJBcDP}
\end{align}
In analogy to \acp{pcfg}, we call this the \glsreset{cnf}\ac{cnf}. Any \ac{rbn} may be rewritten in \ac{cnf} (see \appref{W5B87VoS} for details).

\textbf{\Ac{rbn} Chart:}
During inference, we will make use of an \ac{rbn} \emph{chart}, similar to the parse chart for \acp{pcfg} \citep{grune2007}. Each non-terminal variable is associated to a layer in the chart. For discrete variables, they store the actual distributions, while for continuous variables they either hold the point estimate (for \ac{map} inference) or the parameters of the approximate distributions (for inference in \acl{grbn}).
Different instances of the same template variable are identified by a subscript indicating the span of data generated from them, which also corresponds to their position in the chart (see \figref{zE70w8gY}). Sets of variables that are generated from a specific latent non-terminal variable $\NonTermVar_{i:k}$ are denoted by a bold capital letter with a corresponding subscript ($\NonTermVarBlock_{i:k}$, $\TermVarBlock_{i:k}$, $\StrucVarBlock_{i:k}$); omitting the subscript refers to \emph{all} variables ($\NonTermVarBlock$, $\TermVarBlock$, $\StrucVarBlock$); for $\NonTermVarBlock$ and $\StrucVarBlock$ this also includes the root variables $\NonTermVar_{0:n}$ and $\StrucVar_{0:n}$, respectively. The subscripts are to be interpreted as time intervals, that is, $\TermVarBlock_{i:i}$ is empty, $\TermVarBlock_{0:1}=\TermVar_1$ is the first observation, $\TermVarBlock_{n\shortminus{}2:n}=(\TermVar_{n\shortminus{}1}, \TermVar_n)$ are the last two observations etc.

\textbf{Comparison to \acsp{pcfg}:}
Any \ac{pcfg} can be rewritten as an \ac{rbn} in two different ways, which we call \emph{abstraction} and \emph{expansion} (see \appref{PdZJXFH8} for details). Abstraction of a \ac{pcfg} produces a discrete \ac{rbn} with one latent non-terminal and one observed terminal variable. The resulting \ac{rbn} is exactly equivalent to the original \ac{pcfg} but describes the same relations in a more abstract and compact way. In contrast, \emph{expansion} of a \ac{pcfg} considers the symbols of the grammar as random variables in their own right, thereby endowing them with additional (possibly continuous) degrees of freedom. The resulting \ac{rbn} is therefore more powerful than the original \ac{pcfg}.
A \ac{pcfg} is abstracted to a discrete \ac{rbn} by defining the start/prior, transition, and structural distributions (\ref{eq:3Ejwxtzm}--\ref{eq:CJPJBcDP}) as
{
  \setlength{\abovedisplayskip}{2pt}
  \setlength{\belowdisplayskip}{2pt}
\begin{align}
  &\PriorDist(\NonTermVar{=}A)
    = \frac{W_{S \rightarrow A}}{\sum_{{}_{A^\prime}} \! W_{S \rightarrow A^\prime}} \Label[eq:KOrWOcmf]
  &
  &\NonTermTrans(\NonTermVar^\prime{=}B,\NonTermVar^\pprime{=}C \cond \NonTermVar{=}A)
  = \frac{W_{A \rightarrow BC}}{\sum_{{}_{B^\prime\!,C^\prime}} \! W_{A \rightarrow B^\prime C^\prime}} \label{eq:lGLGSXxQ} \\
  &\TermTrans(\TermVar{=}b \cond \NonTermVar{=}A)
    = \frac{W_{A \rightarrow b}}{\sum_{{}_{b^\prime}} W_{A \rightarrow b^\prime}} \Label[eq:tRhVdzpm]
  &
  &\StrucDist(\StrucVar \cond \NonTermVar{=}A)
  = \begin{cases}
    \frac{\sum_{{}_{B,C}} W_{A \rightarrow BC}}{\sum_{{}_X} W_{A \rightarrow X}} & \text{if } \StrucVar{=}\NonTermLabel \\[3mm]
    \frac{\sum_{{}_{b}} W_{A \rightarrow b}}{\sum_{{}_X} W_{A \rightarrow X}} & \text{if } \StrucVar{=}\TermLabel~,
  \end{cases} \label{eq:G7k2W2ko}
\end{align}%
}%
where $S$ is the grammar's start symbol, $A,B,C$ are non-terminal symbols, $b$ is a terminal symbol, $X$ is any right-hand side of a rule, $\StrucVar{=}\NonTermLabel$ and $\StrucVar{=}\TermLabel$ indicate a non-terminal and terminal transition, respectively, $W_{\cdot\,\rightarrow\,\cdot}$ is the weight of the corresponding rule, and rules that do not exist in the original \ac{pcfg} are taken to have zero weight.
In \emph{expansion}, the \ac{pcfg} is only used to define a ``skeleton'' for the \ac{rbn}, while the specific random variables and the concrete transition distributions need to be additionally specified. This means that the resulting \ac{rbn} model is more powerful than the original \ac{pcfg}, as the symbols may, for instance, be expanded to continuous random variables.

\subsection{Inference}
\label{sec:inference}

The two main goals of inference in \acp{rbn} are to 1)~train model parameters by maximising the marginal data likelihood and to 2)~compute posterior distributions or \glsreset{map}\ac{map} estimates of the network structure and non-terminal variables. In \acp{pcfg}, both is achieved by computing inside and outside probabilities \citep{manning1999}, which will be the starting point for our generalisation to continuous variables.

\textbf{Inside and Outside Probabilities:} We define inside and outside probabilities, $\Inside$ and $\Outside$, for \acp{rbn} in analogy to how they are defined for \acp{pcfg}, the only difference being that the variables may be continuous. We thus have
\begin{align}
  \colA{\Inside[i:k]} &:= p(\TermVarBlock_{i:k} \cond \NonTermVar_{i:k}) \Label[eq:9P164jUG]
  & \text{and}
  &&
     \colA{\Outside[i:k]} &:= p(\TermVarBlock_{0:i}, \NonTermVar_{i:k}, \TermVarBlock_{k:n})~, \label{eq:PBujnIdl}
\end{align}
where $n$ is the length of the sequence and $\TermVarBlock$ is fixed (and therefore omitted as argument on the left-hand side). That is, $\Inside[i:k]$ is the marginal likelihood of generating the sub-sequence $\TermVarBlock_{i:k}$ conditional on the respective non-terminal variable $\NonTermVar_{i:k}$, while $\Outside[i:k]$ is the marginal likelihood of generating the two sub-sequences $\TermVarBlock_{0:i}$ and $\TermVarBlock_{k:n}$ as well as the non-terminal variable $\NonTermVar_{i:k}$. In both cases, $\Inside$ and $\Outside$ are functions of the corresponding non-terminal variable with the structure and the remaining variables being marginalised out. Based on the inside and outside probabilities, the marginal data likelihood and the marginal posterior distributions over non-terminal variables are
\begin{align}
  p(\TermVarBlock) &= \!\!\int\!\! \colA{\Inside[0:n]} \, \PriorDist(\NonTermVar_{0:n}) \, d\NonTermVar_{0:n} \hspace{-0.5em}\Label[eq:ScKTv9k1]
  & \hspace{-0.5em}\text{and}\hspace{-0.5em}
  &&
     \widetilde{p}(\NonTermVar_{i:k} \cond \TermVarBlock) &= \frac{\colA{\Outside[i:k]} \, \colA{\Inside[i:k]}}{p(\TermVarBlock)}~,\hspace{-0.5em} \label{eq:NuTG01QR}
\end{align}
respectively. $\widetilde{p}(\NonTermVar_{i:k} \cond \TermVarBlock)$ is an \emph{unnormalised} probability distribution that specifies the probability of $\NonTermVar_{i:k}$ to exist via the normalisation constant $\int \widetilde{p}(\NonTermVar_{i:k} \cond \TermVarBlock)\,d\NonTermVar_{i:k}$, while the normalised version
corresponds to the marginal posterior distribution of $\NonTermVar_{i:k}$ for the case that it \emph{does} exist.

Inside probabilities are recursively computed bottom-up. For an \ac{rbn} in \ac{cnf} we start with the base case \eqref{eq:fDbELzNN} for single observations and then iterate \eqref{eq:2Q0hmTrM} to the top of the \ac{rbn} chart
\begin{align}
  \hspace{-0.5em}\colA{\Inside[i:i\shortplus1]}
  ={}& \StrucDist(\StrucVar_{i:i\shortplus1}{=}\TermLabel \cond \NonTermVar_{i:i\shortplus1}) \, \TermTrans(\TermVar_{i\shortplus1} \cond \NonTermVar_{i:i\shortplus1}) \label{eq:fDbELzNN} \\
  \colA{\Inside[i:k]}
  ={}&
       \StrucDist(\StrucVar_{i:k}{=}\NonTermLabel \cond \NonTermVar_{i:k})
       \sum_{\mathclap{j=i+1\hspace{1em}}}^{\mathclap{k-1}}
       \hspace{-0.1em}\int\hspace{-0.6em}\int\hspace{-0.15em}
       \NonTermTrans(\NonTermVar_{i:j},\NonTermVar_{j:k} \cond \NonTermVar_{i:k}) \,
       \colA{\Inside[i:j]} \,
       \colA{\Inside[j:k]} \,
       d\NonTermVar_{i:j} \, d\NonTermVar_{j:k}\mathrlap{\,.}
       \label{eq:2Q0hmTrM}
\end{align}
Outside probabilities are recursively computed top-down, while making use of the inside probabilities
\begin{align}
  \colA{\Outside[0:n]}
  ={}& \PriorDist(\NonTermVar_{0:n}) \label{eq:kfCjpgp9} \\
  \colA{\Outside[j:k]}
  ={}&
       \brackets[\Big]{
       \sum_{\mathclap{i=0}}^{j-1}
       \hspace{-0.1em}\int\hspace{-0.6em}\int\hspace{-0.15em}
       \StrucDist(\StrucVar_{i:k}{=}\NonTermLabel \cond \NonTermVar_{i:k}) \,
       \NonTermTrans(\NonTermVar_{i:j},\NonTermVar_{j:k} \cond \NonTermVar_{i:k}) \,
       \colA{\Outside[i:k]} \, \colA{\Inside[i:j]} \,
       d\NonTermVar_{i:j} \, d\NonTermVar_{i:k}
       }
       {}+{}
       \nonumber \newlineAlign*[-0mm]
       \brackets[\Big]{
       \sum_{\mathclap{l=k+1\hspace{1em}}}^{n}
       \hspace{-0.1em}\int\hspace{-0.6em}\int\hspace{-0.15em}
       \StrucDist(\StrucVar_{j:l}{=}\NonTermLabel \cond \NonTermVar_{j:l}) \,
       \NonTermTrans(\NonTermVar_{j:k},\NonTermVar_{k:l} \cond \NonTermVar_{j:l}) \,
       \colA{\Outside[j:l]} \, \colA{\Inside[k:l]} \,
       d\NonTermVar_{j:l} \, d\NonTermVar_{k:l}
       }\mathrlap{.} \label{eq:sjybKfDz}
\end{align}
As for \acp{pcfg}, the two terms in \eqref{eq:sjybKfDz} correspond to the possibility of $\NonTermVar_{j:k}$ being generated as the right or the left child, respectively.
The main conceptual difference to \acp{pcfg} is that we treat the discrete structural part (marginalised out by the sums) separately from the potentially continuous variables (marginalised out by the integrals).
For \acp{rbn} that are not in \ac{cnf},
the equations have to be adapted accordingly (see \appref{general_inside_outside} for the general case).

\textbf{Marginalisation:} Computing the marginal data likelihood \eqref{eq:ScKTv9k1} and the marginal posterior distributions over non-terminal variables \eqref{eq:NuTG01QR} requires to solve an exponential (w.r.t.~the length $n$ of the sequence) number of nested integrals in
(\ref{eq:fDbELzNN}--%
\ref{eq:sjybKfDz}),
which is generally intractable. However, for the special case of \aclp{grbn}, we provide an adaptive closed-form approximation in \secref{grbn}. Moreover, marginalising \emph{only} over the network structure for a fixed assignment of the non-terminal variables $\NonTermVarBlock$ is straight forward and allows for \glsreset{map}\ac{map} inference in general \acp{rbn}.

\textbf{\titleac{map} Inference:} For a fixed assignment of all non-terminal variables $\NonTermVarBlock$, we can compute the joint marginal likelihood $p(\NonTermVarBlock,\TermVarBlock)$ over observed terminal and latent non-terminal variables by only marginalising over the structure. This follows the same principle as above but uses the modified \emph{joint} inside and outside probabilities
\begin{align}
  \colA{\widehat{\Inside}_{i:k}}
  &:= p(\NonTermVarBlock_{i:k}, \TermVarBlock_{i:k} \cond \NonTermVar_{i:k}) \Label[eq:kL0XOP4c]
  & \text{and} &
  &
    \colA{\widehat{\Outside}_{j:k}}
  &:= p(\NonTermVarBlock_{0:j},\TermVarBlock_{0:j}, \NonTermVar_{j:k},  \NonTermVarBlock_{k:n},\TermVarBlock_{k:n})~, \label{eq:ytWkc3dC}
\end{align}
where all variables are fixed (and therefore omitted as arguments on the left-hand side). Analogously, the joint marginal likelihood and the marginal posterior probability of $\NonTermVar_{i:k}$ to exist then are
\begin{align}
  p(\NonTermVarBlock, \TermVarBlock) &= \colA{\widehat{\Inside}_{0:n}} \, \PriorDist(\NonTermVar_{0:n}) \Label[eq:xAfCaCDz]
  & \text{and} &
  &
    \widetilde{p}_{i:k}
  &= \frac{\colA{\widehat{\Outside}_{i:k}} \, \colA{\widehat{\Inside}_{i:k}}}{p(\NonTermVarBlock,\TermVarBlock)}~,\hspace{19ex}
\end{align}
where $\widetilde{p}_{i:k}$ is the probability of $\NonTermVar_{i:k}$ to exist for \emph{this specific} assignment of $\NonTermVarBlock$. The corresponding equations for the recursion differ from (\ref{eq:fDbELzNN}--\ref{eq:sjybKfDz}) only in that they do not integrate out the latent non-terminal variables (see \appref{joint_in_out}). As before, all computations can be efficiently performed via dynamic programming. Gradients w.r.t.~the variables and/or parameters are readily obtained from libraries such as \texttt{PyTorch} \citep{paszke2019}. Optimising the values of the latent non-terminal variables $\NonTermVarBlock$ via gradient descent yields \glsreset{map}\ac{map} estimates, while the structure is marginalised out. \Ac{map} estimates for the structure (i.e.~the best tree) conditional on an assignment for $\NonTermVarBlock$ can be computed (as for \acp{pcfg}) by replacing summation with maximisation \citep{goodman1999,huang2005}.

There are two caveats:~First, due to marginalising over multiple (exponentially many) network structures, $p(\NonTermVarBlock,\TermVarBlock)$ may be highly non-convex and optimising $\NonTermVarBlock$ via gradient descent is not guaranteed to find the global optimum. This is even the case for purely Gaussian \acp{rbn}, for which $p(\NonTermVarBlock,\TermVarBlock)$ is a mixture of Gaussians (one for each structure).
Second, we can optimise $\NonTermVarBlock$ while marginalising out the structure and we can optimise the structure for a fixed assignment of $\NonTermVarBlock$. However, successively optimising $\NonTermVarBlock$ and the structure is not equivalent to \emph{jointly} optimising both and the maximum of $p(\NonTermVarBlock,\TermVarBlock)$ may be unrelated to the maximum of the best structure (also see \figref{jJ4xGGDT}).
This means that generally, \emph{exact joint} \ac{map} inference over the latent variables \emph{and} the structure is hard.
For \aclp{grbn}, we provide an approximate solution below.

\subsection{\titleacp{grbn}}
\label{sec:grbn}

In a \glsreset{grbn}\ac{grbn}, the prior, non-terminal, and terminal distributions are linear Gaussians and the termination probability (structural distribution) is constant
\begin{align}
  \PriorDist(\NonTermVar)
  &:= \Norm*{\NonTermVar}{\mu_\PriorLabel}{\Sigma_\PriorLabel} && \text{[prior]} \label{eq:Bg7E21Fg} \\
  \NonTermTrans(\NonTermVar^\prime,\NonTermVar^\pprime \cond \NonTermVar)
  &:= \Norm*{\NonTermVar^\prime}{\NonTermVar}{\Sigma_{\NonTermLabel\mathrm{L}}} \, \Norm*{\NonTermVar^\pprime}{\NonTermVar}{\Sigma_{\NonTermLabel\mathrm{R}}} && \text{[non-terminal]} \label{eq:FZt5bnvr} \\
  \TermTrans(\TermVar \cond \NonTermVar)
  &:= \Norm*{\TermVar}{\NonTermVar}{\Sigma_\TermLabel} && \text{[terminal]} \label{eq:6JjalCAN} \\
  \StrucDist(\StrucVar{=}\TermLabel \cond \NonTermVar) &:= \TermProb~. && \text{[termination/structural]} \label{eq:uROCbA5z}
\end{align}
For clarity, we will show all derivations for \acp{grbn} in this basic form. For our evaluations and the application to music, we use a slightly extended version that includes linear transformations, mixtures of Gaussians, and multi-terminal transitions (\secref{grbn_music}). The derivations do not fundamentally change for the extended case (see \appref{grbn}). In \appref{example}, we show all calculations on a simple example.

\begin{figure}[t!]
\begin{minipage}[b]{0.47\linewidth}
  \centering
  \begin{tikzpicture}
    \node[inner sep=0,anchor=south west] (0,0) {\includegraphics[width=\linewidth]{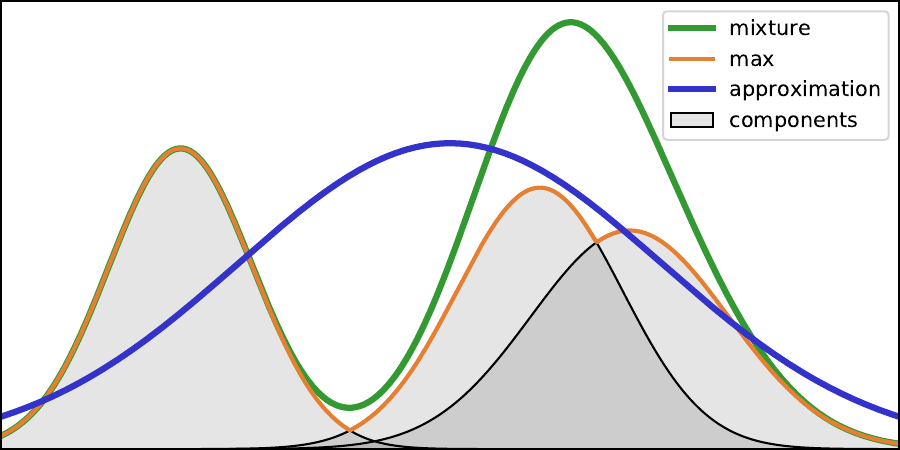}};
    \node at (1.323,2.22) {{\color{BurntOrange}\xmark}};
    \node at (3.3,2.25) {{\color{RoyalBlue}\xmark}};
    \node at (4.175,3.145) {{\color{OliveGreen}\xmark}};
  \end{tikzpicture}
  \caption{Three Gaussians components, the resulting mixture (green), maximum (orange), and moment-matching single Gaussian approximation (blue). Note that the maximum of the mixture ({\color{OliveGreen}\xmark}), the best component ({\color{BurntOrange}\xmark}), and the approximation ({\color{Blue}\xmark}) may be unrelated.}
  \label{fig:jJ4xGGDT}
  \end{minipage}
\hfill
  \begin{minipage}[b]{0.51\linewidth}
    \centering
    \tikzstyle{var}=[circle,draw,fill=white,inner sep=0pt,minimum size=22pt]%
    \tikzstyle{obs}=[fill=black!5]%
    \tikzstyle{edge}=[->,>=stealth]%
    \tikzstyle{plate}=[draw,rounded corners,inner sep=5pt]%
    \tikzstyle{gate}=[draw,dashed,inner sep=5pt]%
    \tikzstyle{factor}=[draw,fill=black,inner sep=0pt,minimum size=5pt]%
    \tikzstyle{con}=[{}-{Circle[scale=1]},shorten >=-2pt]%
    \tikzstyle{param}=[{Circle[scale=0.5]}-{stealth}]%
    \tikzstyle{paramcon}=[{Circle[scale=0.5]}-{Circle[scale=1]},shorten >=-2pt]%
    \def\scale{1.15}%
    \newcommand*{\plateindex}[2][1]{\scalebox{#1}{${}_{#2}$}}%
    \newcommand*{\GRBNonetau}{%
      \raisebox{-0\height}{\scalebox{\scale}{%
          \def\xscale{1.4}%
          \def\xscalecm{1.4cm}%
          \def\yscale{1}%
          \def\yscalecm{1cm}%
          \begin{tikzpicture}[xscale=\xscale,yscale=\yscale]
            \node[var] (alpha) at (0,0) {$\NonTermVar$};
            \node[] (lambda) at (1.8,0) {$\lambda$};
            \begin{scope}[xshift=-1cm,yshift=-2.1cm]
              \node[] (T) at (-1,1) {$W$};
              \node[var] (alpha_left) at (0,0) {$\NonTermVar^\prime$};
              \node[var] (alpha_right) at (1,0) {$\NonTermVar^\pprime$};
              \node[var] (tau') at (0,1) {$\TransVar$};
              \begin{scope}[yshift=0.5cm]
                \node[var,obs] (b) at (2,0) {$\TermVar$};
                \node[plate, fit=(b)] (b_plate) {};
                \node[anchor=south east] at (b_plate.south east) {\plateindex[0.9]{n}};
                \node[var] (n) at (lambda |- b) {$n$};
              \end{scope}
              \node[gate, anchor=south west, minimum height=2.2*\yscalecm, minimum width=3.7*\xscalecm] (a_gate) at (-0.5,-0.7) {};
              \node[anchor=east, minimum height=2.2*\yscalecm, minimum width=1.7*\xscalecm] (b_gate) at (a_gate.east) {};
              \draw[gate] (b_gate.north west) -- (b_gate.south west);
              \node[anchor=south west] at (a_gate.south west) {\plateindex[0.8]{\StrucVar=\NonTermLabel}};
              \node[anchor=south west] at (b_gate.south west) {\plateindex[0.8]{\StrucVar=\TermLabel}};
            \end{scope}
            \node[var] (sigma) at (alpha -| a_gate.north west) {$\StrucVar$};
            \incoming[][0.6]{alpha}
            \draw[edge,densely dotted] (alpha_left) -- (alpha_left |- a_gate.south);
            \draw[edge,densely dotted] (alpha_right) -- (alpha_right |- a_gate.south);
            \draw[edge] (alpha) -| (b);
            \draw[edge,con] (sigma) -- (a_gate.north west);
            \draw[edge] (alpha) -- (alpha_left);
            \draw[edge] (alpha) -- (alpha_right);
            \draw[edge] (tau') -- (alpha_left);
            \draw[param] (lambda) -- (n);
            \draw[con] (n) -- (b_plate);
            \draw[param] (T) -- (tau');
          \end{tikzpicture}
        }}}
    \GRBNonetau{}
    \caption{Graphical model of the \acl{grbn} for modelling music. The additional transposition variable $\TransVar$ is marginalised out during inference; the number of jointly generated observations $n$ is uniquely determined by the location in the parse chart.}
    \label{fig:xokxDDAz}
  \end{minipage}
\end{figure}

\textbf{Adaptive Approximation:} If the structure of a \ac{grbn} was fixed, all variables would be jointly Gaussian distributed as in a conventional Gaussian \acl{bn} \citep{koller_probabilistic_2009}. However, due to the unknown structure, we effectively have a mixture of exponentially many Gaussians, one for each possible structure. While in principle all integrals can be solved analytically, the exponential growth makes exact inference intractable. Therefore, our goal is to derive a parsing strategy that retains tractability by adaptively applying local approximations to the \aclp{gm} occurring in each recursion step.
We will here focus on the simplest case of approximating the mixtures with a single Gaussian (illustrated in \figref{jJ4xGGDT}, details in \appref{approx}), which can be efficiently computed in closed form \citep{bishop07,orguner2007}.
The inside and outside probabilities are thus represented by a simple Gaussian
{
\renewcommand*{\colB}[1]{}
\begin{align}
  \colA{\Inside[i:k]}
  &\approx c_{i:k}^{(\Inside)} \colB{\sum_l w_l} \, \Norm{\NonTermVar_{i:k}}{\mu_{i:k}^{(\Inside)\colB{l}}}{\Sigma_{i:k}^{(\Inside)\colB{l}}}
                        \Label[eq:3XIqlHPD]
                        &
  \colA{\Outside[j:k]}
  &\approx c_{j:k}^{(\Outside)} \colB{\sum_l w_l} \, \Norm{\NonTermVar_{j:k}}{\mu_{j:k}^{(\Outside)\colB{l}}}{\Sigma_{j:k}^{(\Outside)\colB{l}}}
                        \label{eq:aHhkDVDP}
\end{align}
}%
and this form is reestablished in each iteration by approximating the occurring mixtures. Consequently, the marginal posterior distributions over latent variables \eqref{eq:NuTG01QR} are also simple Gaussians and the marginal data likelihood \eqref{eq:ScKTv9k1} can be computed in closed form. This approximation scheme can be extended and refined by using existing methods for approximating each \acl{gm} by one with fewer components \citep{huber2008b,crouse2011}.

\textbf{Marginalisation:} In \eqref{eq:2Q0hmTrM} and \eqref{eq:sjybKfDz}, we have to integrate over products of Gaussian distributions to marginalise out the latent variables. To solve these integrals, we make use of the fact that the product of two Gaussians over a variable $x$ can be rewritten as \citep[see e.g.][]{petersen06}
\begin{align}
  \Norm{x}{\mu_1}{\Sigma_1} \, \Norm{x}{\mu_2}{\Sigma_2} ={}& \Norm{\mu_1}{\mu_2}{\Sigma_1+\Sigma_2} \, \Norm{x}{\bar\mu}{\bar\Sigma} \label{eq:P8MeIPHK}
\end{align}
with
\begin{align}
  \bar\Sigma :={}& \parens{\Sigma_1^{-1} + \Sigma_2^{-1}}^{-1}
  & \text{and} &
  & \bar\mu :={}& \bar\Sigma \left(\Sigma_1^{-1}\mu_1 + \Sigma_2^{-1}\mu_2\right)~.
\end{align}
Hence, when integrating over $x$, only the first term on the rhs.~of \eqref{eq:P8MeIPHK} remains.
A detailed step-by-step derivation of all results can be found in \appref{marginalisation}.
With the latent variables being marginalised out, \eqref{eq:2Q0hmTrM} and \eqref{eq:sjybKfDz} become simple mixtures of Gaussians that can be
easily approximated
to retain the simple analytic form
of the inside and outside probabilities.

\textbf{Tree Induction:} As described above, exact joint \ac{map} inference over the continuous latent variables and the structure is generally intractable. Moreover, the maximum of the approximate posterior does not necessarily coincide with the maximum of the exact posterior or that of a particular structure (see \figref{jJ4xGGDT}). Thus, first optimising $\NonTermVarBlock$ (based on the approximation) and then estimating the structure (conditional on the picked value of $\NonTermVarBlock$) may lead to arbitrarily bad results for tree induction.
Therefore, we leverage the adaptive character of our approximation scheme to compute local structure estimates in each step, before loosing relevant information due to further approximations. Specifically, during the bottom-up pass for computing inside probabilities, all structures are scored by the maximum of their marginal likelihood,
based on its current approximation \eqref{eq:3XIqlHPD}. The best overall structure is then selected (as usual) in a top-down pass (see \appref{tree_induction} and our example in \appref{example}).

\subsubsection{\titleacp{grbn} for Music}
\label{sec:grbn_music}

For the application to music, we slightly extend the basic \ac{grbn} discussed so far by introducing \emph{transpositions} and \emph{multi-terminal transitions} (changes in the equations highlighted in \colC{blue}). The corresponding graphical model of the \ac{rbn} cell is shown in \figref{xokxDDAz}. Furthermore, we describe how \acp{grbn} can be applied to \emph{categorical} data.

\textbf{Transpositions:} A transposition rotates the dimensions of the latent variable by a number of steps $\TransVar$ before generating the child. This is achieved by multiplying with an orthonormal transposition matrix \colC{$\TransMat_\TransVar$} that corresponds to the identity matrix with cyclicly rearranged columns. For the prior distribution, we assume a uniform weighting of all possible transpositions
\begin{align}
  \PriorDist(\NonTermVar)
  &:= \colC{\sum_{\TransVar=0}^{D-1} \frac{1}{D}} \, \Norm{\NonTermVar}{\colC{\TransMat_\TransVar} \, \mu_{\mathrm{p}}}{\Sigma_{\mathrm{P}}}~, && \text{[prior]}
\end{align}
where $D$ is the dimensionality of the data ($D=12$ for music in 12-tone equal temperament). For the non-terminal transitions, the probability for a specific transposition is determined by the weight parameter $W$
\begin{align}
  \NonTermTrans(\NonTermVar^\prime,\NonTermVar^\pprime \cond \NonTermVar)
  &:= \colC{\sum_{\TransVar=0}^{D-1} p(\TransVar \cond W)} \, \Norm{\NonTermVar^\prime}{\colC{\TransMat_\TransVar} \, \NonTermVar}{\Sigma_{\NonTermLabel\mathrm{L}}}\Norm{\NonTermVar^\pprime}{\NonTermVar}{\Sigma_{\NonTermLabel\mathrm{R}}}~. \label{eq:swzRuMxz}
\end{align}
Note that transpositions are only applied to the left child, because Western classical music is thought to be fundamentally goal directed \citep{lerdahl1983,rohrmeier2011,koelsch2013}. This means that the character of a section is largely determined by how it ends (the right child), which should also be reflected in the value of the parent node. In contrast, the role of the left child is to harmonically prepare the ending (or prepare a preparation to the ending etc). We therefore allow for arbitrary transpositions in the left child and we will see below that our model indeed captures the most important type of preparation in Western classical music: the cadential dominant-tonic progression.

\textbf{Multi-Terminal Transitions:} A multi-terminal transition generates multiple observed variables from a single latent variable. The variables are generated i.i.d.~and their number is governed by a Poisson distribution with rate parameter~$\lambda$
\begin{align}
  \TermTrans(\TermVar_{i:k} \cond \NonTermVar_{i:k})
  &:= \colC{\mathrm{Pois}(k-i-1 \cond \lambda) \prod_{\mathclap{j=i+1}}^k} \Norm{\TermVar_j}{\NonTermVar_{i:k}}{\Sigma_\TermLabel}~. && \text{[multi-terminal]} \label{eq:HIUzl7Mb}
\end{align}
Multi-terminal transitions do not conform to the \ac{cnf} assumed so far and we need to add the term
\begin{align}
  \colA{\Inside[i:k]}
  ={}& \cdots + \StrucDist(\StrucVar_{i:k}{=}\TermLabel \cond \NonTermVar_{i:k}) \, \TermTrans(\TermVar_{i:k} \cond \NonTermVar_{i:k}) \label{eq:P31WZcXh} \\
  ={}& \cdots + \TermProb \, \TermTrans(\TermVar_{i:k} \cond \NonTermVar_{i:k}) & \text{[for \acp{grbn}, see \eqref{eq:uROCbA5z}]}
\end{align}
to \eqref{eq:2Q0hmTrM} in order to account for the possibility to terminate from a higher-level variable. For $k=i+1$, this term becomes the base case \eqref{eq:fDbELzNN} of an \ac{rbn} in \ac{cnf}.

Multi-terminal transitions account for the situation where changes in the hierarchical structure occur at a lower rate than the time series is sampled. In between the structural changes, the data is assumed to be generated from the same model, which could also be more elaborate than i.i.d.~samples, as long as the relevant model parameters are captured by the \ac{rbn}'s latent variables.

\textbf{Categorical Data:} The observed variables of a \ac{grbn} are unconstrained real-valued, which poses a problem if the data are categorical. This situation is comparable to using \acp{gp} \citep{rasmussen2006} for classification and can be approached with similar methods. In our application to musical data, we observe one or more notes being played at any particular time and normalise these counts to obtain observations that correspond to the parameter of a categorical distribution. The natural likelihood function for this type of observations is a Dirichlet distribution. Therefore, we adapt the approach suggested in \citep{milios2018} for \acp{gp}, who assume a Dirichlet likelihood, which is then approximated by a Gaussian likelihood in log-space. Since an observation from a Dirichlet distribution corresponds to a normalised sample from independent Gamma distributions, each Gamma distribution can be separately approximated by a log-normal distribution, which results in a diagonal covariance matrix for the Gaussian likelihood in log-space. Matching the first and second moment yields \citep{milios2018}
\begin{align}
  \widetilde{\TermVar}_j^{(l)} &= \log \TermVar_j^{(l)} - \widetilde{\Sigma}_{ll}^{(j)}/2 & \text{and} && \widetilde{\Sigma}_{ll}^{(j)} &= \log\parens{1/\TermVar_j^{(l)}+1}~,
\end{align}
where $0<\TermVar_j^{(l)}<1$ is the $l^\text{th}$ element (normalised count) of the $j^\text{th}$ observation, $\widetilde{\TermVar}_j^{(l)}$ is the corresponding mean of the approximate Gaussian likelihood in log-space, and $\widetilde{\Sigma}_{ll}^{(j)}$ is the $l^\text{th}$ element on the diagonal of the covariance matrix for the $j^\text{th}$ observation. We thus have to replace $\widetilde{\TermVar}_j$ and $\widetilde{\Sigma}^{(j)}$ for $\TermVar_j$ and $\Sigma_\TermLabel$ in \eqref{eq:HIUzl7Mb}.

\section{Experiments}
\label{sec:LZ876DnR}

We performed a quantitative evaluation on synthetic data and applied our model to hierarchical music analysis of Bach preludes. We show that \acp{rbn} are superior to \ac{cpd} and \ac{hc} for tree induction and our method is able to infer fundamental harmonic principles of Western classical music.
Experiments were run on a 3.6 GHz Quad-Core Intel Core i7 processor with 32GB RAM. The model parameters were trained via gradient descent on the (approximate) marginal neg-log likelihood.

\subsection{Quantitative Evaluation on Tree Induction}
\label{sec:quant}

\begin{figure}[t!]
  \centering
  \begin{tabular}{@{\hspace{-1ex}}l@{}l@{\hspace{5ex}}l@{}l@{}}
    \vcenterbox{\includegraphics[width=0.407\textwidth,trim=4 0 304mm 2mm,clip]{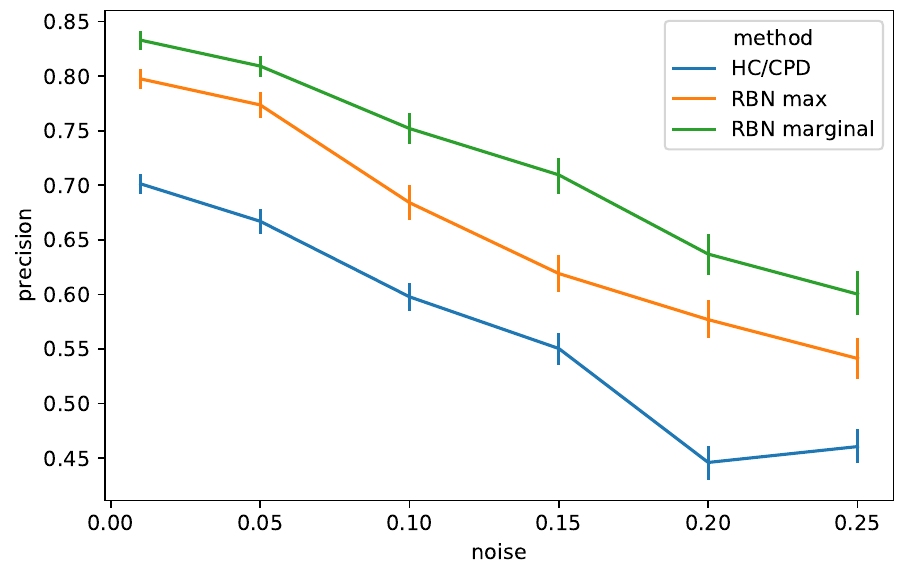}}
    &&
    \hspace*{-5.2mm}\raisebox{-0.41\height}[0mm][0mm]{\scalebox{0.905}{%
    \def\blacklength{-0.7}%
    \def\blackoffset{-0.4}%
    \begin{tikzpicture}[xscale=1,yscale=0.7]
      \foreach \x in {0,...,6} {
        \draw (\x,-0.01) -- (\x,-1) -- (\x+1,-1) -- (\x+1,-0.01);
      }
      \foreach \x in {1,3} {
        \draw[fill=black!20] (\x*3/5,0) -- (\x*3/5,\blackoffset) -- ({(\x+1)*3/5},\blackoffset) -- ({(\x+1)*3/5},0);
        \draw[fill=black,rounded corners=0.5mm] (\x*3/5,\blackoffset) -- (\x*3/5,\blacklength) -- ({(\x+1)*3/5},\blacklength) -- ({(\x+1)*3/5},\blackoffset);
      }
      \foreach \x in {1,3,5} {
        \draw[fill=black!20] (3+\x*4/7,0) -- (3+\x*4/7,\blackoffset) -- ({3+(\x+1)*4/7},\blackoffset) -- ({3+(\x+1)*4/7},0);
        \draw[fill=black,rounded corners=0.5mm] (3+\x*4/7,\blackoffset) -- (3+\x*4/7,\blacklength) -- ({3+(\x+1)*4/7},\blacklength) -- ({3+(\x+1)*4/7},\blackoffset);
      }
      \node[anchor=south west] at (-0.86cm,-6mm) {\includegraphics[width=7.8cm,trim=0 5mm 0 0,clip]{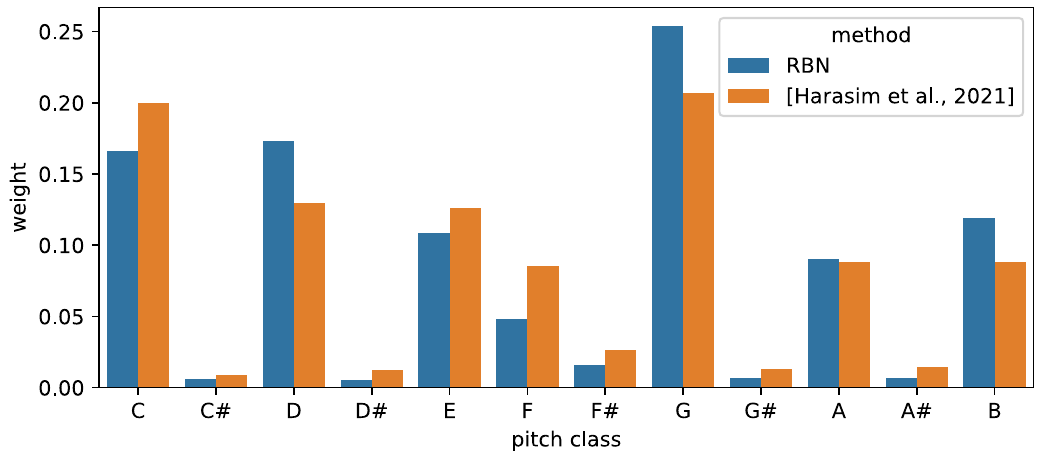}};
    \end{tikzpicture}%
    }}
    & \raisebox{-14mm}[0mm][0mm]{\textbf{(b)}}
    \\
    \hspace*{0.6ex}\vcenterbox{\includegraphics[width=0.4\textwidth,trim=154.5mm 0 153.5mm 2mm,clip]{./figures/plot_scores_l50_n500_m5_pretrained_RBN}}
    &~\raisebox{-13mm}[0mm][0mm]{\textbf{(a)}}
    &
      \hspace*{-5mm}\raisebox{-0.47\height}{\includegraphics[width=0.52\textwidth]{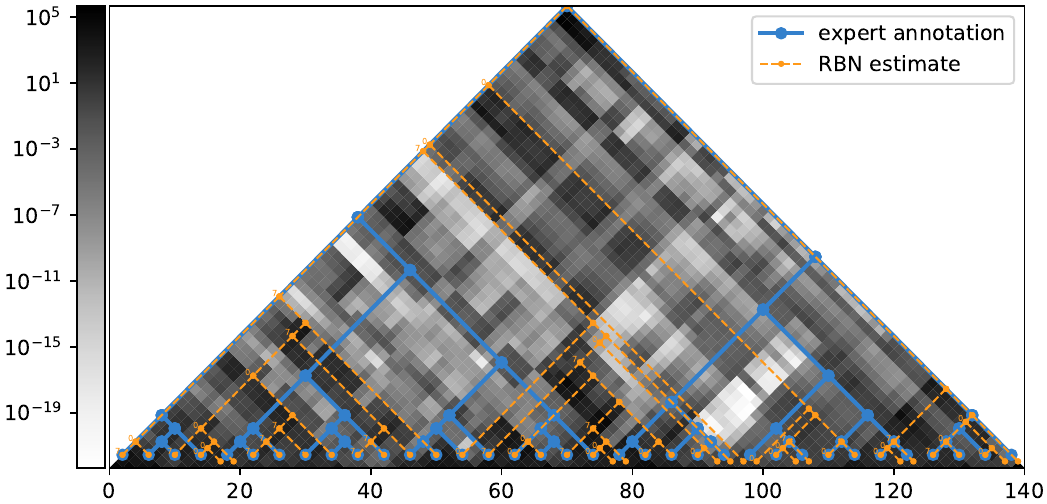}}
    & \raisebox{-13mm}[0mm][0mm]{\textbf{(c)}}
      \rlap{\phantom{\ac{hc}\ac{cpd}}}
  \end{tabular}
  \caption{
    \textbf{(a):}~Precision and recall w.r.t.~the ground-truth trees of 500 sequences for different noise levels for the baseline (blue) and the maximum and marginal \ac{rbn} estimates (orange, green); error bars indicate 95\% confidence intervals from bootstrapping (see \appref{lSWGmWpX} for technical details).
    \textbf{(b):}~Prior mean learned by our \ac{grbn} for music in comparison to recent values from the literature \citep{harasim2021}.
    \textbf{(c):}~Comparison of our model (orange/grey) to an expert annotation (blue) for Johann Sebastian Bach's Prelude No.~1 in C major, BWV 846. The greyscale indicates the  marginal probability of a node to exist at that particular location; the small numbers indicate the transposition in semitones for a left child; time is indicated in beats (quarter notes); the piece was divided into two-beat (half note) intervals. The plot follows the idea of \emph{scape plots} \citep{sapp2001,muller2012,lieckModelling2020}.
  }
  \label{fig:V9Y7WXs1}
\end{figure}

We performed a quantitative evaluation on synthetic data for the task of segmenting a noisy time series and inferring the underlying tree. For comparison, we used the best-performing \glsreset{cpd}\ac{cpd} method from the \texttt{ruptures} library \citep{truong2020} for segmenting the time series, combined with bottom-up \glsreset{hc}\ac{hc} for inferring the tree structure (``HC/CPD''). For details of the methodology, see \appref{lSWGmWpX}.

The evaluation results in \figref{V9Y7WXs1}(a) show that the \ac{rbn} tree estimates (\secref{grbn}) consistently outperform the one from \ac{hc}/\ac{cpd}, in terms of both precision and recall (and thus also in F1 measure).
The marginal node probabilities show an interesting performance pattern. They excel in terms of precision, which means that a node with high marginal probability is very likely to actually exist in the tree (low false-positive rate). However, they severely underestimate the overall node probabilities, which leads to recall falling far below the baseline. This means that a node with low marginal probability may in fact occur in the tree (high false-negative rate).

We think that the poor recall measure of the marginal probabilities is primarily due to (and the downside of) a fully Bayesian treatment that quantifies uncertainty. Even if the marginal probabilities have a maximum at the correct node location, probability mass will still spread around it and be allocated to a number of less probable locations. While this is the desired behaviour of a Bayesian method, it inevitably results in a lower recall value. The high precision value confirms that uncertainty is adequately quantified and not underestimated. That being said, the marginal probabilities provide an exceptionally rich basis for qualitative analyses. For instance, all ground-truth nodes are located at local maxima of the marginal probabilities and we can read off a number of other potential node locations, which essentially trace out the grid defined by the piece-wise constant segments (see \figref{PVpySY8K} in \appref{lSWGmWpX}).

\subsection{Hierarchical Music Analysis}
\label{sec:JM0tLiUM}

Harmonies in Western classical music exhibit a nested hierarchical structure that can be modeled by \acp{pcfg} operating on abstract chord symbols \citep{rohrmeier2011,rohrmeier_pearce_2018,harasimGeneralized2018,rohrmeier2020}. While these grammars can be applied to expert annotations of a musical score, hierarchical music analysis from the raw note level is an unsolved problem.
We trained a \ac{grbn} (\secref{grbn_music}) on the 24 major preludes of Johann Sebastian Bach's ``Wohltemperiertes Klavier I \& II'' (see \appref{music} for technical details and complete results).

Our first major finding is that the prior mean, shown in \figref{V9Y7WXs1}(b), corresponds to a major pitch profile (as could be expected from the training data) and is in excellent agreement with recent Bayesian estimates from the literature \citep{harasim2021}.
The fact that the major profile appears in the prior (i.e.~as the continuous equivalent of a grammar's start symbol) shows that our model picks up fundamentally important structures from the musical data.
Our second finding is that only two transpositions have non-zero weights: the identity with a weight of 78\% and the fifth scale degree (7 semitones) with a weight of 22\%. This corresponds to the left child being generated as the dominant of the parent and realises the most important harmonic preparation in Western classical music: the cadential dominant-tonic relation.
A closer inspection of the expert analysis (\figref{hYxyR3RZ} in \appref{music}) reveals that when considering the possible surface patterns (raw notes) of the labeled chords, most non-identity transitions can indeed be explained as (noisy) fifth transpositions.
The strong weight of fifth transpositions in our model is a highly non-trivial empirical confirmation of the established music theoretical insight that Baroque music is fundamentally driven by dominant-tonic relations.
While the estimated tree in \figref{V9Y7WXs1}(c) fails to reproduce the large-scale structure of the expert analysis (e.g.~the separation into two main parts), it accurately captures the measure-wise harmonic changes on the bottom level.

On the one hand, we see considerable room for improvement by integrating more advanced concepts, such as different modes (major/minor), diatonic in addition to chromatic transposition, or balancing of trees. On the other hand, our model was able to capture fundamental properties of Western classical music based on only 24 pieces. We therefore think that \aclp{grbn} are a highly promising approach for hierarchical music analysis from the raw note level, which should be further investigated.

\section{Conclusion}
\label{sec:VjrT4kdN}

We introduced \glsreset{rbn}\acp{rbn}, a novel class of probabilistic models that unifies the strengths of \glsreset{pcfg}\acp{pcfg} and \glsreset{dbn}\acp{dbn}, generalising both model classes.
We defined \acp{rbn} as a joint distribution over tree-structured Bayesian networks and their (discrete or continuous) variables and described how to perform inference over both the model structure and the variables by leveraging parsing methods for \acp{pcfg}. The provided formalisation connects with the methods for formal grammar as well as with the versatile notation for graphical models. On two data sets, we demonstrated the potential of \acp{rbn} for modelling nested hierarchical dependencies in real-valued time series and musical data.
The class of \acp{rbn} represents a substantial contribution to the machine learning toolkit by unifying two of the most important approaches for modelling sequential data and bears a large potential for further development and applications.

\begin{ack}
  This project has received funding from the European Research Council (ERC) under the European Union's Horizon 2020 research and innovation programme under grant agreement No 760081 – PMSB.
  This project was conducted at the Latour Chair in Digital and Cognitive Musicology, generously funded by Mr. Claude Latour.

\end{ack}

{
  \small
  \bibliography{literature.bib}
}

\clearpage

\appendix

\part{Appendix}

{\large\emph{Recursive Bayesian Networks: Generalising and Unifying Probabilistic Context-Free Grammars and Dynamic Bayesian Networks}}

\parttoc

\section{Theory}

\subsection{Transformation to \titleac{cnf}}
\label{app:W5B87VoS}

\newcommand*{\exnote}[1]{}

Transforming an \ac{rbn} into \ac{cnf} is done analogously to the procedure for \acp{pcfg}. We assume the original \ac{rbn} does not contain any epsilon productions, that is, a non-terminal variable always produces one or more other non-terminal and/or terminal variables. \exnote{ On the left we use the standard notation for conditional probability distribution \eqref{eq:2tAZjnlo}, on the right in [brackets], we use the consolidated notation from \eqref{eq:us8JMj6O} to highlight the parallels to \acp{pcfg}.}

\textbf{1) Eliminate terminal variables from mixed transitions:} This is done by introducing intermediate non-terminal variables. For each transition
\begin{align}
  &p(\NonTermVar^{(1)},\NonTermVar^{(2)},\ldots,\TermVar^{(1)},\TermVar^{(2)},\ldots \cond \NonTermVar)\exnote{&& [a \leadsto \NonTermVar^{(1)} \NonTermVar^{(2)} \cdots \TermVar^{(1)} \TermVar^{(2)} \cdots]}~,
\end{align}
with non-terminal variables $\NonTermVar^{(1)},\NonTermVar^{(2)},\ldots\in \SetNonTermVars$ and terminal variables $\TermVar^{(1)},\TermVar^{(2)},\ldots\in \SetTermVars$, we introduce new non-terminal variables $\NonTermVar_{\TermVar^{(1)}},\NonTermVar_{\TermVar^{(2)}},\ldots$ and replace the transition by
\begin{align}
  &p(\NonTermVar^{(1)},\NonTermVar^{(2)},\ldots,\NonTermVar_{\TermVar^{(1)}},\NonTermVar_{\TermVar^{(2)}},\ldots \cond \NonTermVar)~, \exnote{&& [a \leadsto \NonTermVar^{(1)} \NonTermVar^{(2)} \cdots \NonTermVar_{\TermVar^{(1)}} \NonTermVar_{\TermVar^{(2)}} \cdots]}
\end{align}
where the new non-terminals $\NonTermVar_{\TermVar^{(i)}}$ replace the original terminals $\TermVar^{(i)}$. We then add new deterministic transitions
\begin{align}
  &p(\TermVar^{(1)} \cond \NonTermVar_{\TermVar^{(1)}}) \exnote{&& [\NonTermVar_{\TermVar^{(1)}} \leadsto \TermVar^{(1)}]} \nonumber\\
  &p(\TermVar^{(2)} \cond \NonTermVar_{\TermVar^{(2)}}) \exnote{&&[\NonTermVar_{\TermVar^{(2)}} \leadsto \TermVar^{(2)}]} \\[-1mm]
  &\ \vdots \exnote{&&\ \vdots} \nonumber
\end{align}
that convert each non-terminal to its equivalent terminal variable. For the newly added non-terminals, there is only a single transition and hence a degenerate structural variable that can only take a single value.

\textbf{2) Eliminate more than two latent non-terminal variables:} This is done by introducing new non-terminals that capture combinations of multiple old non-terminals. Below, we show how the number of non-terminals can be reduced by one. Applying this procedure repeatedly allows for reducing the number of non-terminals from an arbitrary number down to two, as required for \ac{cnf}. A transition
\begin{align}
  p(\NonTermVar^{(1)},\NonTermVar^{(2)},\ldots,\NonTermVar^{(n)} \cond \NonTermVar)
\end{align}
that generates $n$ non-terminals $\NonTermVar^{(1)},\ldots,\NonTermVar^{(n)}$ is rewritten as
\begin{align}
 p(\NonTermVar^{(1)},\NonTermVar^{(2)},\ldots,\NonTermVar^{(n-1)} \cond \NonTermVar^\prime) \, p(\NonTermVar^\prime, \NonTermVar^{(n)} \cond \NonTermVar)~,
\end{align}
where we introduced the new non-terminal variable $\NonTermVar^\prime=(\NonTermVar^{(1)}, \NonTermVar^{(2)},\ldots, \NonTermVar^{(n-1)})$ that stores all the information from the first $n-1$ original non-terminals. The actual ``work'' is done by $p(\NonTermVar^\prime, \NonTermVar^{(n)} \cond \NonTermVar)$, which is the equivalent of the original $n$-fold transition. $p(\NonTermVar^{(1)},\NonTermVar^{(2)},\ldots,\NonTermVar^{(n-1)} \cond \NonTermVar^\prime)$ is a deterministic transition that just ``unpacks'' the information stored in $\NonTermVar^\prime$. Repeating this procedure to come to only pairwise transitions corresponds to a chain of these deterministic ``unpacking'' operations. As above, the newly added non-terminals have only a single possible transition.

\textbf{3) Eliminate unary cycles:} Unary cycles $p_{\text{cycle}}(\NonTermVar^\prime \cond \NonTermVar)$, where $\NonTermVar$ and $\NonTermVar^\prime$ are the \emph{same} non-terminal template variable ($\NonTermVar\equiv\NonTermVar^\prime$), are first transformed into unary transitions to a new non-terminal variable and then eliminated as described below. We define a new non-terminal variable $\bar\NonTermVar=(\NonTermVar^\prime,n)$, where $n>0$ represents the number of steps taken in the cycle before exiting it and $\NonTermVar^\prime$ is the value at the moment of exiting it.
The transition distribution to $\bar\NonTermVar$ is
\begin{align}
  p(\bar\NonTermVar \cond \NonTermVar) = p(\NonTermVar^\prime, n \cond \NonTermVar)
  ={}&
       \begin{cases}
         p(\StrucVar{\neq}\text{cycle} \cond \NonTermVar) \, p_{\text{cycle}}(\NonTermVar^\prime \cond \NonTermVar) & \text{if } n = 1 \\
         \int p(\StrucVar{=}\text{cycle} \cond \NonTermVar^\pprime) \, p_{\text{cycle}}(\NonTermVar^\prime \cond \NonTermVar^\pprime) \, p(\NonTermVar^\pprime, n-1 \cond \NonTermVar) \, d\NonTermVar^\pprime & \text{if } n > 1~,
       \end{cases}
\end{align}
where in the recursive case, the variables of all intermediate steps are successively marginalised out. In practical applications, if $p(\StrucVar{=}\text{cycle} \cond \NonTermVar^\pprime)<1$, the probability of remaining in the cycle decays exponentially and the recursion can be truncated after a number of steps. If on the other hand $p(\StrucVar{=}\text{cycle} \cond \NonTermVar^\pprime)\approx1$ so that truncating is not possible, one can work with the stationary distribution of the resulting Markov chain (i.e.~the Markov chain with transition distribution $p_{\text{cycle}}(\NonTermVar^\prime \cond \NonTermVar)$).

The structural probability to take a transition from $\NonTermVar$ to $\bar\NonTermVar$ is $p(\StrucVar{=}\text{cycle} \cond \NonTermVar)$, i.e.~the probability of entering the cycle in the first place. The \ac{rbn} cell of $\bar\NonTermVar$ is identical to that of $\NonTermVar$, except for the transition into the cycle, which is eliminated (the structural distribution thus has to be renormalised for the remaining transitions). The transitions use only the $\NonTermVar^\prime$-component of $\bar\NonTermVar$, ignoring the $n$-component. In this way, we have expresses the state after an arbitrary number of steps in the unary cycle as a distinct value of the new non-terminal variable $\bar\NonTermVar$.

\textbf{4) Eliminate unary transitions between non-terminal variables:} Unary transitions $p_{\text{unary}}(\NonTermVar^\prime \cond \NonTermVar)$, where $\NonTermVar$ and $\NonTermVar^\prime$ are \emph{different} non-terminal template variables, are transformed by treating $\NonTermVar^\prime$ as an intermediate variable and marginalising it out. All transitions $p^{(1)},\ldots,p^{(n)}$ from $\NonTermVar^\prime$ to some other variables (terminal and/or non-terminal)
\begin{align}
  &p^{(1)}(\ldots \cond \NonTermVar^\prime) \exnote{&& [ \NonTermVar^\prime \leadsto \NonTermVar^\pprime \cdots ]} \nonumber\\[-1mm]
  &\ \vdots \exnote{&&\ \vdots}\\[-1mm]
  &p^{(n)}(\ldots \cond \NonTermVar^\prime) \exnote{&& [ \NonTermVar^\prime \leadsto \NonTermVar^\ppprime \cdots ]} \nonumber
\end{align}
are replaced by a set of new transitions $p_1,\ldots,p_n$ from $\NonTermVar$ directly to the respective variables, with the intermediate variable $\NonTermVar^\prime$ marginalised out
\begin{align}
  p_1(\ldots \cond \NonTermVar) &= \int p^{(1)}(\ldots \cond \NonTermVar^\prime) \, p_{\text{unary}}(\NonTermVar^\prime \cond \NonTermVar) \, d\NonTermVar^\prime \nonumber\\[-1mm]
                              &\ \vdots \\[-1mm]
  p_n(\ldots \cond \NonTermVar) &= \int p^{(n)}(\ldots \cond \NonTermVar^\prime) \, p_{\text{unary}}(\NonTermVar^\prime \cond \NonTermVar) \, d\NonTermVar^\prime~. \nonumber
\end{align}
The intermediate variable $\NonTermVar^\prime$ and its \ac{rbn} cell is eliminated if it was only reachable via $\NonTermVar$.
The new transitions $p_1,\ldots,p_n$ are merged into the cell of $\NonTermVar$, while the original transition $p_{\text{unary}}$ to $\NonTermVar^\prime$ is removed. This requires redefining the structural distribution $p(\StrucVar \cond \NonTermVar)$ such that the probability mass $p(\StrucVar{=}\text{unary} \cond \NonTermVar)$ that was formerly assigned to $p_{\text{unary}}$ is now split among the new transitions $p_1$ to $p_n$ according to the structural distribution $p(\StrucVar^\prime \cond \NonTermVar^\prime)$ of $\NonTermVar^\prime$. Specifically, for a new transition $p_i$, we define
\begin{align}
  p(\StrucVar{=}i \cond \NonTermVar) &:={} p(\StrucVar{=}\text{unary} \cond \NonTermVar) \, p(\StrucVar^\prime{=}i \cond \NonTermVar^\prime)~.
\end{align}

\subsection{Relation to \acp{pcfg}}
\label{app:PdZJXFH8}

As described in \secref{definition}, a \ac{pcfg} can be rewritten as an \ac{rbn} by \emph{abstraction} or \emph{expansion}, where abstraction produces an equivalent \ac{rbn} that describes the same relations in a more abstract and compact way, while expansion produces a more general \ac{rbn} using the original \ac{pcfg} as a skeleton. We describe the two procedures in detail below and use the following definition of a \ac{pcfg}:
\begin{definition}[\titleac{pcfg}]
  A \ac{pcfg} is a tuple $(N,T,S,R,W)$ of
  \begin{align}
    N &: \text{non-terminal symbols}
    &
      T &: \text{terminal symbols}
    &
      S &: \text{start symbol} \\
    R &: \text{production rules}\in N \times (N \cup T)^*
    &
      W &: \text{rule weights}~.
  \end{align}
\end{definition}

\subsubsection{Abstraction of a \acs{pcfg}}

\begin{theorem}
  A \ac{pcfg} in \ac{cnf} can be \emph{abstracted} to an equivalent discrete \ac{rbn} in \ac{cnf} with one latent (non-terminal) template variable $\NonTermVar$ and one observed (terminal) template variable $\TermVar$ by defining the prior, transition, and structural distributions as
  \begin{align}
    &\PriorDist(\NonTermVar{=}A)
      = \frac{W_{S \rightarrow A}}{\sum_{{}_{A^\prime}} \! W_{S \rightarrow A^\prime}} &\eqref{eq:KOrWOcmf}&
    &
    &\NonTermTrans(\NonTermVar^\prime{=}B,\NonTermVar^\pprime{=}C \cond \NonTermVar{=}A)
      = \frac{W_{A \rightarrow BC}}{\sum_{{}_{B^\prime\!,C^\prime}} \! W_{A \rightarrow B^\prime C^\prime}} \tag{\ref{eq:lGLGSXxQ}} \\
    &\TermTrans(\TermVar{=}b \cond \NonTermVar{=}A)
      = \frac{W_{A \rightarrow b}}{\sum_{{}_{b^\prime}} W_{A \rightarrow b^\prime}} &\eqref{eq:tRhVdzpm}&
    &
    &\StrucDist(\StrucVar \cond \NonTermVar{=}A)
      = \begin{cases}
        \frac{\sum_{{}_{B,C}} W_{A \rightarrow BC}}{\sum_{{}_X} W_{A \rightarrow X}} & \text{if } \StrucVar{=}\NonTermLabel \\[3mm]
        \frac{\sum_{{}_{b}} W_{A \rightarrow b}}{\sum_{{}_X} W_{A \rightarrow X}} & \text{if } \StrucVar{=}\TermLabel~,
      \end{cases} \tag{\ref{eq:G7k2W2ko}}
  \end{align}
where $A,B,C\in N$ are non-terminal symbols of the \ac{pcfg}, $b\in T$ is a terminal symbol, $X\in N^2\cup T$ is any right-hand side of a rule, $\StrucVar{=}\NonTermLabel$ and $\StrucVar{=}\TermLabel$ indicate a non-terminal and terminal transition, respectively, $W_{A \rightarrow X}$ is the weight of the corresponding \ac{pcfg} rule, and rules that do not exist in the original \ac{pcfg} are taken to have zero weight.
\end{theorem}
To show equivalence, we need to prove that the transition probabilities from a given non-terminal symbol $A$ are the same in the original \ac{pcfg} and the new \ac{rbn}.
\begin{proof}
  In the \ac{rbn}, the probability for a non-terminal transition $A \rightarrow B\,C$ is
  \begin{align}
    P(A \rightarrow B\,C)
    &=
      \StrucDist(\StrucVar{=}\NonTermLabel \cond \NonTermVar{=}A) \,
      \NonTermTrans(\NonTermVar^\prime{=}B,\NonTermVar^\pprime{=}C \cond \NonTermVar{=}A) \\
    &=
      \frac{\sum_{{}_{B^\prime\!,C^\prime}} \! W_{A \rightarrow B^\prime C^\prime}}{\sum_{{}_X} W_{A \rightarrow X}}
      \frac{W_{A \rightarrow BC}}{\sum_{{}_{B^\prime\!,C^\prime}} \! W_{A \rightarrow B^\prime C^\prime}} \\
    &=
      \frac{W_{A \rightarrow BC}}{\sum_{{}_X} W_{A \rightarrow X}}
  \end{align}
  and that for a terminal transition $A \rightarrow b$ is
  \begin{align}
    P(A \rightarrow b)
    &=
      \StrucDist(\StrucVar{=}\TermLabel \cond \NonTermVar{=}A) \,
      \TermTrans(\TermVar{=}b \cond \NonTermVar{=}A) \\
    &=
      \frac{\sum_{{}_{b^\prime}} W_{A \rightarrow b^\prime}}{\sum_{{}_X} W_{A \rightarrow X}}
      \frac{W_{A \rightarrow b}}{\sum_{{}_{b^\prime}} W_{A \rightarrow b^\prime}} \\
    &=
      \frac{W_{A \rightarrow b}}{\sum_{{}_X} W_{A \rightarrow X}}~,
  \end{align}
  which matches the corresponding probabilities in the \ac{pcfg}, gained by normalising the respective weights.
\end{proof}
Conversely, any discrete \ac{rbn} can be rewritten as a \ac{pcfg}.
\begin{theorem}
  A discrete \ac{rbn} with $n$ latent non-terminal template variables $\NonTermVar_1,\ldots,\NonTermVar_n$, $m$ observed terminal template variables $\TermVar_1,\ldots,\TermVar_m$, and a prior $\PriorDist(\NonTermVar_1)$ over $\NonTermVar_1$ can be rewritten as a \ac{pcfg} with
  \begin{align}
    N
    :={}& \NonTermVar_1 \oplus \cdots \oplus \NonTermVar_n \\
    T
    :={}& \TermVar_1 \oplus \cdots \oplus \TermVar_m \\
    W_{A \rightarrow X}
    :={}&
          \begin{cases}
            \PriorDist(X) & \text{if } A=S \land X \in \NonTermVar_1 \\
            p(\StrucVar{=}i \cond A)\,p_i(X \cond A) & \text{if a matching transition exists in the \ac{rbn}} \\
            0 & \text{else}~,
          \end{cases} \label{eq:DjfVRhoK}
  \end{align}
  where $\cdot\,\oplus\,\cdot$ concatenates the value ranges of the respective variables, $X\in \NonTermVar_i$ denote that the value $X$ is in the value range of the \ac{rbn} variable $\NonTermVar_i$, and the second case in \eqref{eq:DjfVRhoK} requires there be a transition $p_i(x_1,\ldots,x_k\cond \NonTermVar_i)$ such that $A\in \NonTermVar_i$ and $X \in x_1\oplus\cdots\oplus x_k$.
\end{theorem}

\subsubsection{Expansion of a \acs{pcfg}}

\emph{Expansion} of a \ac{pcfg} to an \ac{rbn} uses the \ac{pcfg} as a ``skeleton'' to define the number of template variables and the structural transitions. The domains and transitions for the variables need to be added, which results in an \ac{rbn} that is more powerful than the original \ac{pcfg}. Specifically, we have
\begin{align}
  \SetNonTermVars :={}& \{\NonTermVar_A \cond A \in N\} & \text{and} && \SetTermVars :={}& \{\TermVar_b \cond b \in T\}
\end{align}
for the sets of latent non-terminal and observed terminal template variables and
\begin{align}
  p(\StrucVar_A{=}X \cond \NonTermVar_A) &= \frac{W_{A \rightarrow X}}{\sum_{X^\prime} W_{A \rightarrow X^\prime}} & \text{with} && A \in N \text{ and } X, X^\prime \in (N \cup T)^*
\end{align}
for the structural transitions. Additional, we have to define the domain for each of the non-terminal and terminal variables in $\SetNonTermVars$ and $\SetTermVars$, and for each rule $A \rightarrow X_1\,X_2\ldots$ from the original \ac{pcfg}, we have to define a concrete transition distribution $p(\NonTermOrTermVar_{X_1},\NonTermOrTermVar_{X_2},\ldots \cond \NonTermVar_A)$ for the \ac{rbn} (where $\NonTermOrTermVar_{X_1}, \NonTermOrTermVar_{X_2}, \ldots \in \SetNonTermVars\cup\SetTermVars$ are non-terminal or terminal variables in the \ac{rbn}, respectively, depending on whether $X_1,X_2,\ldots\in N\cup T$ are non-terminal or terminal symbols in the \ac{pcfg}).

Expansion of a \ac{pcfg} into an \ac{rbn} seems appealing if a simple \ac{pcfg} can be used to describe the \emph{type} of variables (as opposed to their values) as well as the structure of the generative process. The actual transitions on the variables' values may then take place on a sub-symbolic/continuous level, which cannot be described by a \ac{pcfg}.

\subsection{General Inside and Outside Probabilities}
\label{app:general_inside_outside}

\subsubsection{Inside Probabilities}

The inside probability
\begin{align}
  \colA{\Inside[i:k]}
  ={}& p(\TermVarBlock_{i:k} \cond \NonTermVar_{i:k})
\end{align}
is the probability of generating the observed terminal variables $\TermVarBlock_{i:k}$ from the latent non-terminal variable $\NonTermVar_{i:k}$. This means that we need to marginalise over all possible paths of generation. Transitions may directly generate observed variables, but they may also generate lower-level non-terminals, in which case we have to recurse using the respective inside probabilities from those variables.

Let $\SetTransDist_\NonTermVar\subseteq\SetTransDist$ be the set of possible transitions from the latent non-terminal template variable $\NonTermVar\in\SetNonTermVars$ (of which $\NonTermVar_{i:k}$ is one specific instantiation), with $p(\StrucVar_{i:k}{=}\TransDist \cond \NonTermVar_{i:k})$ being the probability for the transition $\TransDist \in \SetTransDist_\NonTermVar$ to be selected. This constitutes the first sum in \eqref{eq:eSXAvxdQ} below, which marginalises over the transitions.

The transition $\TransDist$ generates $\Arity$ new non-terminal and/or terminal variables, where $\Arity$ is the arity of $\TransDist$. These may be located at different positions in the parse chart, depending on which part of the observed variables $\TermVarBlock_{i:k}$ is generated from them. That is, the variables' locations in the parse chart are not known during generation and are determined in hindsight once all observed variables are generated; thus, they \emph{are} known for parsing. We denote the respective splitting points by $j_1,\ldots,j_{\Arity-1}$ (they have to fulfill the condition $i<j_1<\ldots<j_{\Arity-1}<k$) and the respective variables by $\NonTermOrTermVar_{i:j_1}, \ldots, \NonTermOrTermVar_{j_{\Arity-1}:k}\in\SetNonTermVars\cup\SetTermVars$. The second multi-sum in \eqref{eq:eSXAvxdQ} aggregates the probabilities of the different splitting possibilities, that is, of all valid assignments of $j_1,\ldots,j_{\Arity-1}$ ($\Arity-1$ degrees of freedom). For instance, a transition of arity $\Arity=2$ has one free splitting point $j_1$ to sum over.

Some of the generated variables may be observed/terminal variables, for which nothing more needs to be done as they directly constitute the respective part of $\TermVarBlock_{i:k}$. For the subset of non-terminal variables, which we denote by $\{\NonTermOrTermVar_{j:j^\prime}\in\SetNonTermVars\}$, we need to insert their respective inside probabilities and marginalise them out. This constitutes the product and multi-integral in \eqref{eq:eSXAvxdQ}.

The general form of the inside probabilities then is
\begin{align}
  \colA{\Inside[i:k]}
  ={}&
       \sum_{\mathclap{\TransDist \in \SetTransDist_\NonTermVar}}
       \StrucDist(\StrucVar_{i:k}{=}\TransDist \cond \NonTermVar_{i:k})
       \hspace{1ex}\multisum_{\mathclap{i<j_1<\ldots<j_{\Arity-1}<k}}
       \nonumber\newlineAlign*[0mm]
       \idotsint\limits_{\mathclap{\{\NonTermOrTermVar_{j:j^\prime}\in\SetNonTermVars\}}}
       p_{\TransDist}(\NonTermOrTermVar_{i:j_1}, \ldots, \NonTermOrTermVar_{j_{\Arity-1}:k} \cond \NonTermVar_{i:k})
       \hspace{2ex}
       \prod_{\mathclap{\{\NonTermOrTermVar_{j:j^\prime}\in\SetNonTermVars\}}}
       \colA{\Inside[j:j^\prime][\NonTermOrTermVar]}~. \label{eq:eSXAvxdQ}
\end{align}
The concrete \acp{rbn} considered in the paper have only two transitions, one non-terminal transition of arity two and one terminal transition of arity one (for \ac{cnf}) or more (for the extended \acp{grbn} used in the quantitative evaluation and for modelling music).
For non-terminal transition of arity two, the multi-sum in \eqref{eq:eSXAvxdQ} reduces to a single sum and the multi-integral to a double integral, which gives us \eqref{eq:2Q0hmTrM}. For the terminal transition, \eqref{eq:eSXAvxdQ} simplifies to \eqref{eq:fDbELzNN} or the extended version \eqref{eq:P31WZcXh}, respectively.

\subsubsection{Outside Probabilities}

The outside probability
\begin{align}
  \colA{\Outside[j:j^\prime]}
  ={}&
       p(\TermVarBlock_{0:j}, \NonTermVar_{j:j^\prime}, \TermVarBlock_{j^\prime:n})
\end{align}
is the joint probability of generating the latent non-terminal variable $\NonTermVar_{j:j^\prime}$ as well as the prefix and suffix of observed terminal variables,  $\TermVarBlock_{0:j}$ and $\TermVarBlock_{j^\prime:n}$, respectively. For this, we now have to consider all possible ways how $\NonTermVar_{j:j^\prime}$ as well as the prefix and suffix could have been generated from a parent non-terminal $\bar\NonTermVar$ ($\NonTermVar$ and $\bar\NonTermVar$ may correspond to the same template variable or to two different ones).

Let $\SetTransDist_\NonTermVar^{-1}\subseteq\SetTransDist$ denote the set of transitions that include $\NonTermVar$ as one of the generated variables. Importantly, if $\NonTermVar$ appears multiple times in the generated variables of a particular transition, these different options of generating $\NonTermVar$ are represented as multiple distinct entries in $\SetTransDist_\NonTermVar^{-1}$, one for each occurrence. The first sum in \eqref{eq:o9LzbyCz} runs over these different possibilities of generating $\NonTermVar$.

For a transition $\TransDist\in\SetTransDist_\NonTermVar^{-1}$ of arity $\Arity$, let $j_0,\ldots,j_\Arity$ be the splitting points, including the start and end point $j_0$ and $j_\Arity$ of the parent variable $\bar\NonTermVar_{j_0:j_\Arity}$, which have to fulfill the condition $0\le j_0<\ldots<j_\Arity\le n$ (where $n$ is the length of the sequence). One pair of adjacent splitting points $(j_m,j_{m\shortplus1})$ corresponds to the occurrence of $\NonTermVar_{j:j^\prime}$, where $m$ is the position (starting at zero) at which $\NonTermVar$ appears in the generated variables of the particular transition $\TransDist$. We therefore have the additional constraints $j_m=j$ and $j_{m\shortplus1}=j^\prime$, resulting in $\Arity-1$ remaining free indices to sum over (as for the inside probabilities above). This corresponds to the second multi-sum in \eqref{eq:o9LzbyCz}.

The set of non-terminal variables generated from the parent $\bar\NonTermVar_{j_0:j_\Arity}$, excluding $\NonTermVar_{j:j^\prime}$, is denoted by $\{\NonTermOrTermVar_{l:l^\prime}\in\SetNonTermVars\}\setminus \NonTermVar_{j:j^\prime}$. Together with the directly generated terminal variables, these generate part of the prefix and suffix, $\TermVarBlock_{j_0:j}$ and $\TermVarBlock_{j^\prime:j_\Arity}$. The remaining prefix and suffix, $\TermVarBlock_{0:j_0}$ and $\TermVarBlock_{j_\Arity:n}$, are generated from the parent variable  $\bar\NonTermVar_{j_0:j_\Arity}$. For the parent, we recurse via its outside probability $\Outside[j_0:j_\Arity][\bar\NonTermVar]$, while for the newly generated non-terminal variables (except $\NonTermVar_{j:j^\prime}$), we have to use the respective inside probability $\Inside[l:l^\prime][\NonTermOrTermVar]$ in \eqref{eq:o9LzbyCz}. Additionally, we have to marginalise out the parent (first integral) and the newly generated non-terminal variables (second multi-integral).

The general outside probabilities then are
\begin{align}
  \colA{\Outside[j:j^\prime]}
  ={}&
       \sum_{\mathclap{\TransDist \in \SetTransDist_\NonTermVar^{-1}}}
       \hspace{4ex}
       \multisum_{\mathclap{\substack{0\le j_0<\ldots<j_\Arity\le n\\j_m=j \land j_{m\shortplus1}=j^\prime}}}
  \hspace{4ex}
  \int\limits_{\bar\NonTermVar_{j_0:j_\Arity}}
  \hspace{3ex}
  \idotsint\limits_{\mathclap{\{\NonTermOrTermVar_{l:l^\prime}\in\SetNonTermVars\}\setminus \NonTermVar_{j:j^\prime}}}
  p_{\mathrm{s}_a}(\StrucVar_{j_0:j_\Arity}{=}\TransDist \cond \bar\NonTermVar_{j_0:j_\Arity}) \label{eq:o9LzbyCz}
  \newlineAlign*[0mm]
  p_{\TransDist}(\NonTermOrTermVar_{j_0:j_1},\ldots,\NonTermVar_{j:j^\prime},\ldots,\NonTermOrTermVar_{j_{\Arity-1}:j_\Arity} \cond \bar\NonTermVar_{j_0:j_\Arity}) \,
  \colA{\Outside[j_0:j_\Arity][\bar\NonTermVar]}
  \hspace{1ex}
  \prod_{\mathclap{\{\NonTermOrTermVar_{l:l^\prime}\in\SetNonTermVars\}\setminus \NonTermVar_{j:j^\prime}}}
  \colA{\Inside[l:l^\prime][\NonTermOrTermVar]}~. \nonumber
\end{align}
For a non-terminal transition of arity two, as we have it in the paper, the multi-sum in \eqref{eq:o9LzbyCz} reduces to a single sum and $\{\NonTermOrTermVar_{l:l^\prime}\in\SetNonTermVars\}\setminus \NonTermVar_{j:j^\prime}$ contains only a single non-terminal, the second child. Importantly, $\SetTransDist_\NonTermVar^{-1}$ has two elements, one for $\NonTermVar_{j:j^\prime}$ being generated as the right child and one for it being generated as the left child, which gives us \eqref{eq:sjybKfDz}.

\subsubsection{Joint Inside and Outside Probabilities}
\label{app:joint_in_out}

The joint inside and outside probabilities \eqref{eq:kL0XOP4c} and \eqref{eq:ytWkc3dC} for an \ac{rbn} in \ac{cnf} are computed analogously to (\ref{eq:fDbELzNN}--\ref{eq:sjybKfDz}) for the normal inside and outside probabilities, that is,
\begin{align}
  \hspace{-0.5em}\colA{\widehat{\Inside}_{i:i\shortplus1}}
  ={}& \StrucDist(\StrucVar_{i:i\shortplus1}{=}\TermLabel \cond \NonTermVar_{i:i\shortplus1}) \, \TermTrans(\TermVar_{i\shortplus1} \cond \NonTermVar_{i:i\shortplus1}) \\
  \colA{\widehat{\Inside}_{i:k}}
  ={}&
       \StrucDist(\StrucVar_{i:k}{=}\NonTermLabel \cond \NonTermVar_{i:k})
       \sum_{\mathclap{j=i+1\hspace{1em}}}^{\mathclap{k-1}}
       \NonTermTrans(\NonTermVar_{i:j},\NonTermVar_{j:k} \cond \NonTermVar_{i:k}) \,
       \colA{\widehat{\Inside}_{i:j}} \,
       \colA{\widehat{\Inside}_{j:k}} \\
  \colA{\widehat{\Outside}_{0:n}}
  ={}& \PriorDist(\NonTermVar_{0:n}) \\
  \colA{\widehat{\Outside}_{j:k}}
  ={}&
       \brackets[\Big]{
       \sum_{\mathclap{i=0}}^{j-1}
       \StrucDist(\StrucVar_{i:k}{=}\NonTermLabel \cond \NonTermVar_{i:k}) \,
       \NonTermTrans(\NonTermVar_{i:j},\NonTermVar_{j:k} \cond \NonTermVar_{i:k}) \,
       \colA{\widehat{\Outside}_{i:k}} \, \colA{\widehat{\Inside}_{i:j}}
       }
       {}+{}
       \nonumber \newlineAlign*[-0mm]
       \brackets[\Big]{
       \sum_{\mathclap{l=k+1\hspace{1em}}}^{n}
       \StrucDist(\StrucVar_{j:l}{=}\NonTermLabel \cond \NonTermVar_{j:l}) \,
       \NonTermTrans(\NonTermVar_{j:k},\NonTermVar_{k:l} \cond \NonTermVar_{j:l}) \,
       \colA{\widehat{\Outside}_{j:l}} \, \colA{\widehat{\Inside}_{k:l}}
       }~.
\end{align}
This differs from (\ref{eq:fDbELzNN}--\ref{eq:sjybKfDz}) only by dropping the integrals and dependencies on the non-terminal variables (as their values are now fixed). Joint inside and outside probabilities for the general case are obtained from \eqref{eq:eSXAvxdQ} and \eqref{eq:o9LzbyCz} analogously, i.e.~again by dropping the integrals and dependencies on the non-terminal variables.

\subsection{\titleacp{grbn}}
\label{app:grbn}

In the following, we present derivations for the extended case of \acp{grbn}, described in \secref{grbn_music}, which includes linear transformations $\colC{\TransMat}$ for the left child. For this, we will make use of the fact that a normal distribution over a transformed variable $\colC{\TransMat} x$ can be rewritten as
\begin{align}
  \Norm{\colC{\TransMat} x}{\mu}{\Sigma}
  &= \frac{1}{\bars*{\det*{\colC{\TransMat}}}} \Norm{x}{\colC{\TransMat^{-1}}\mu}{\colC{\TransMat^{-1}}\Sigma\,\colC{{T^\top}^{-1}}} \\
  &= \Norm{x}{\colC{\TransMat^\top}\mu}{\colC{\TransMat^\top}\Sigma\,\colC{\TransMat}}~, \label{eq:L2ANPyRd}
\end{align}
where $\bars{\det*{\colC{\TransMat}}}$ is the absolute value of the determinant of $\colC{\TransMat}$ and in \eqref{eq:L2ANPyRd} we made use of the fact that in our case, the transformation matrices are orthonormal, so that $\colC{\TransMat^{-1}}=\colC{\TransMat^\top}$ and $\bars{\det*{\colC{\TransMat}}}=1$.

Note that for an implementation, some of the results should be rewritten in order to minimise the number of matrix inverses that need to be taken. In particular, the identity
\begin{align}
  \parens{\Sigma_1^{-1} + \Sigma_2^{-1}}^{-1}
  &= \Sigma_1\parens{\Sigma_1 + \Sigma_2}^{-1}\Sigma_2
\end{align}
is useful for the implementation, but we omit it in our derivation for clarity.

\subsubsection{Marginalisation}
\label{app:marginalisation}

For the inside probability $\colA{\Inside[i:k]}$, the integral in \eqref{eq:2Q0hmTrM} is
{
  \renewcommand*{\colB}[1]{}
\begin{align}
  \iint
  &
    \NonTermTrans(\NonTermVar_{i:j},\NonTermVar_{j:k} \cond \NonTermVar_{i:k}) \,
    \colA{\Inside[i:j]} \,
    \colA{\Inside[j:k]} \,
    d\NonTermVar_{i:j} \, d\NonTermVar_{j:k} \nonumber \\
  ={}&
       \colC{\sum_\TransVar w_\TransVar}
       \colB{\sum_{l,m} w_l \, w_m} \,
       c_{i:j}^{(\Inside)} \,
       c_{j:k}^{(\Inside)}
       \iint
       \Norm{\NonTermVar_{i:j}}{\colC{\TransMat_\TransVar}\NonTermVar_{i:k}}{\Sigma_{\NonTermLabel\mathrm{L}}} \,
       \Norm{\NonTermVar_{j:k}}{\NonTermVar_{i:k}}{\Sigma_{\NonTermLabel\mathrm{R}}}
       \nonumber\newlineAlign*[20mm]
       \Norm{\NonTermVar_{i:j}}{\mu_{i:j}^{(\Inside)\colB{l}}}{\Sigma_{i:j}^{(\Inside)\colB{l}}} \,
       \Norm{\NonTermVar_{j:k}}{\mu_{j:k}^{(\Inside)\colB{m}}}{\Sigma_{j:k}^{(\Inside)\colB{m}}} \,
       d\NonTermVar_{i:j} \, d\NonTermVar_{j:k} \label{eq:ut0S43vr} \\
  ={}&
       \colC{\sum_\TransVar w_\TransVar}
       \colB{\sum_{l,m} w_l \, w_m} \,
       c_{i:j}^{(\Inside)} \,
       c_{j:k}^{(\Inside)} \,
       \Norm{\colC{\TransMat_\TransVar}\NonTermVar_{i:k}}{\mu_{i:j}^{(\Inside)\colB{l}}}{\Sigma_{\NonTermLabel\mathrm{L}}+\Sigma_{i:j}^{(\Inside)\colB{l}}} \,
       \Norm{\NonTermVar_{i:k}}{\mu_{j:k}^{(\Inside)\colB{m}}}{\Sigma_{\NonTermLabel\mathrm{R}}+\Sigma_{j:k}^{(\Inside)\colB{m}}} \label{eq:GzXNN2fD} \\
  ={}&
       \colC{\sum_\TransVar w_\TransVar}
       \colB{\sum_{l,m} w_l \, w_m} \,
       c_{i:j}^{(\Inside)} \,
       c_{j:k}^{(\Inside)} \,
       \Norm{\NonTermVar_{i:k}}{\colC{\TransMat_\TransVar^\top}\mu_{i:j}^{(\Inside)\colB{l}}}{\colC{\TransMat_\TransVar^\top}\brackets{\Sigma_{\NonTermLabel\mathrm{L}}+\Sigma_{i:j}^{(\Inside)\colB{l}}}\colC{\TransMat_\TransVar}} \,
       \Norm{\NonTermVar_{i:k}}{\mu_{j:k}^{(\Inside)\colB{m}}}{\Sigma_{\NonTermLabel\mathrm{R}}+\Sigma_{j:k}^{(\Inside)\colB{m}}} \label{eq:6hOufLzN} \\
  ={}&
       \colC{\sum_\TransVar w_\TransVar}
       \colB{\sum_{l,m} w_l \, w_m} \,
       c_{i:j:k}^{(\Inside)} \,
       \Norm{\NonTermVar_{i:k}}{\mu_{i:j:k}^{(\Inside)\colB{lm}}}{\Sigma_{i:j:k}^{(\Inside)\colB{lm}}} \label{eq:62Q4A7pD}
\end{align}
}%
with
{
\renewcommand*{\colB}[1]{}
\begin{align}
  c_{i:j:k}^{(\Inside)}
  :={}& c_{i:j}^{(\Inside)} \,
       c_{j:k}^{(\Inside)} \,
       \Norm{\colC{\TransMat_\TransVar^\top}\mu_{i:j}^{(\Inside)\colB{l}}}{\mu_{j:k}^{(\Inside)\colB{m}}}{\colC{\TransMat_\TransVar^\top}\brackets{\Sigma_{\NonTermLabel\mathrm{L}}+\Sigma_{i:j}^{(\Inside)\colB{l}}}\colC{\TransMat_\TransVar}+\Sigma_{\NonTermLabel\mathrm{R}}+\Sigma_{j:k}^{(\Inside)\colB{m}}} \label{eq:OuIhYpEh} \\
  \mu_{i:j:k}^{(\Inside)\colB{lm}}
  :={}& \Sigma_{i:j:k}^{(\Inside)\colB{lm}} \brackets*{
        \colC{\TransMat_\TransVar^\top}\parens*{\Sigma_{\NonTermLabel\mathrm{L}} + \Sigma_{i:j}^{(\Inside)\colB{l}}}^{-1}
        \mu_{i:j}^{(\Inside)\colB{l}}
        +
        \parens*{\Sigma_{\NonTermLabel\mathrm{R}} + \Sigma_{j:k}^{(\Inside)\colB{m}}}^{-1}
        \mu_{j:k}^{(\Inside)\colB{m}}
        } \label{eq:60dfWImN} \\
  \Sigma_{i:j:k}^{(\Inside)\colB{lm}}
  :={}& \brackets*{
        \parens*{\colC{\TransMat_\TransVar^\top}\brackets{\Sigma_{\NonTermLabel\mathrm{L}} + \Sigma_{i:j}^{(\Inside)\colB{l}}}\colC{\TransMat_\TransVar}}^{-1}
        +
        \parens*{\Sigma_{\NonTermLabel\mathrm{R}} + \Sigma_{j:k}^{(\Inside)\colB{m}}}^{-1}
        }^{-1}~, \label{eq:1fhmMSM1}
\end{align}
}%
where in \eqref{eq:ut0S43vr} we inserted \eqref{eq:3XIqlHPD} and \eqref{eq:swzRuMxz};
in \eqref{eq:GzXNN2fD} we used \eqref{eq:P8MeIPHK} twice to rewrite the pairwise products of Gaussians over $\NonTermVar_{i:j}$ and $\NonTermVar_{j:k}$ and marginalise them out; in \eqref{eq:6hOufLzN} we used \eqref{eq:L2ANPyRd} to rewrite the transformation; and in \eqref{eq:62Q4A7pD} we used \eqref{eq:P8MeIPHK} a third time to rewrite the resulting product as a single Gaussian over $\NonTermVar_{i:k}$.

For the outside probability $\colA{\Outside[j:k]}$, the integrals in \eqref{eq:sjybKfDz} for $\NonTermVar_{j:k}$ being generated as the right child are
{
\renewcommand*{\colB}[1]{}
\begin{align}
  \iint&
         \NonTermTrans(\NonTermVar_{i:j},\NonTermVar_{j:k} \cond \NonTermVar_{i:k}) \,
         \colA{\Outside[i:k]} \, \colA{\Inside[i:j]} \,
         d\NonTermVar_{i:j} \, d\NonTermVar_{i:k} \label{eq:hdGfmqOd} \\
  ={}&
       \colC{\sum_\TransVar w_\TransVar}
       c_{i:k}^{(\Outside)}
       c_{i:j}^{(\Inside)}
       \colB{\sum_{l,m} w_l \, w_m}
       \iint
       \Norm{\NonTermVar_{i:j}}{\colC{\TransMat_\TransVar}\NonTermVar_{i:k}}{\Sigma_{\NonTermLabel\mathrm{L}}} \,
       \Norm{\NonTermVar_{j:k}}{\NonTermVar_{i:k}}{\Sigma_{\NonTermLabel\mathrm{R}}}
       \nonumber\newlineAlign*[30mm]
       \Norm{\NonTermVar_{i:k}}{\mu_{i:k}^{(\Outside)\colB{l}}}{\Sigma_{i:k}^{(\Outside)\colB{l}}} \,
       \Norm{\NonTermVar_{i:j}}{\mu_{i:j}^{(\Inside)\colB{m}}}{\Sigma_{i:j}^{(\Inside)\colB{m}}} \,
       d\NonTermVar_{i:j} \, d\NonTermVar_{i:k} \label{eq:QywtiDbz} \\
  ={}&
       \colC{\sum_\TransVar w_\TransVar}
       c_{i:k}^{(\Outside)}
       c_{i:j}^{(\Inside)}
       \colB{\sum_{l,m} w_l \, w_m}
       \int
       \Norm{\NonTermVar_{j:k}}{\NonTermVar_{i:k}}{\Sigma_{\NonTermLabel\mathrm{R}}} \,
       \Norm{\NonTermVar_{i:k}}{\mu_{i:k}^{(\Outside)\colB{l}}}{\Sigma_{i:k}^{(\Outside)\colB{l}}} \,
       \Norm{\colC{\TransMat_\TransVar}\NonTermVar_{i:k}}{\mu_{i:j}^{(\Inside)\colB{m}}}{\Sigma^{(1)}} \,
       d\NonTermVar_{i:k} \label{eq:08DC8IOL} \\
  ={}&
       \colC{\sum_\TransVar w_\TransVar}
       c_{i:k}^{(\Outside)}
       c_{i:j}^{(\Inside)}
       \colB{\sum_{l,m} w_l \, w_m}
       \int
       \Norm{\NonTermVar_{j:k}}{\NonTermVar_{i:k}}{\Sigma_{\NonTermLabel\mathrm{R}}} \,
       \Norm{\NonTermVar_{i:k}}{\mu_{i:k}^{(\Outside)\colB{l}}}{\Sigma_{i:k}^{(\Outside)\colB{l}}} \,
       \Norm{\NonTermVar_{i:k}}{\colC{\TransMat_\TransVar^\top}\mu_{i:j}^{(\Inside)\colB{m}}}{\colC{\TransMat_\TransVar^\top}\Sigma^{(1)}\colC{\TransMat_\TransVar}} \,
       d\NonTermVar_{i:k} \label{eq:oq6SZsqT} \\
  ={}&
       \colC{\sum_\TransVar w_\TransVar}
       c_{i:k}^{(\Outside)}
       c_{i:j}^{(\Inside)}
       \colB{\sum_{l,m} w_l \, w_m}
       \int
       \Norm{\NonTermVar_{j:k}}{\NonTermVar_{i:k}}{\Sigma_{\NonTermLabel\mathrm{R}}} \,
       \Norm{\NonTermVar_{i:k}}{\mu^{(2)}}{\Sigma^{(2)}} \,
       \Norm{\mu_{i:k}^{(\Outside)\colB{l}}}{\colC{\TransMat_\TransVar^\top}\mu_{i:j}^{(\Inside)\colB{m}}}{\Sigma^{(3)}} \,
       d\NonTermVar_{i:k} \label{eq:MjsgwXC8} \\
  ={}&
       \colC{\sum_\TransVar w_\TransVar}
       c_{i:k}^{(\Outside)}
       c_{i:j}^{(\Inside)}
       \colB{\sum_{l,m} w_l \, w_m}
       \Norm{\mu_{i:k}^{(\Outside)\colB{l}}}{\colC{\TransMat_\TransVar^\top}\mu_{i:j}^{(\Inside)\colB{m}}}{\Sigma^{(3)}} \,
       \Norm{\NonTermVar_{j:k}}{\mu^{(2)}}{\Sigma^{(4)}} \label{eq:7tTRq6w2} \\
  ={}&
       \colC{\sum_\TransVar w_\TransVar} \,
       \colB{\sum_{l,m} w_l \, w_m}
       c_{i:j:k}^{(\Outside)} \,
       \Norm{\NonTermVar_{j:k}}{\mu_{i:j:k}^{(\Outside)}}{\Sigma_{i:j:k}^{(\Outside)}} \label{eq:AXn5wXJa}
\end{align}
}%
with
{
\renewcommand*{\colB}[1]{}
\begin{align}
  \Sigma^{(1)} &= \Sigma_{\NonTermLabel\mathrm{L}} + \Sigma_{i:j}^{(\Inside)\colB{m}} &
                                                                          \Sigma^{(2)} &= \brackets[\Big]{\parens[\big]{\Sigma_{i:k}^{(\Outside)\colB{l}}}^{\!-1}+\parens[\big]{\colC{\TransMat_\TransVar^\top}\Sigma^{(1)}\colC{\TransMat_\TransVar}}^{\!-1}}^{-1} \\
  \Sigma^{(3)} &= \Sigma_{i:k}^{(\Outside)\colB{l}} + \colC{\TransMat_\TransVar^\top}\Sigma^{(1)}\colC{\TransMat_\TransVar} &
  \mu^{(2)} &= \Sigma^{(2)} \brackets[\Big]{
                \parens[\big]{\Sigma_{i:k}^{(\Outside)\colB{l}}}^{\!-1}\mu_{i:k}^{(\Outside)\colB{l}}
                +
                \colC{\TransMat_\TransVar^\top}\parens[\big]{\Sigma^{(1)}}^{\!-1}\mu_{i:j}^{(\Inside)\colB{m}}
              } \\
  \Sigma^{(4)} &= \Sigma_{\NonTermLabel\mathrm{R}} + \Sigma^{(2)}
\end{align}
}%
and
{
\renewcommand*{\colB}[1]{}
\begin{align}
  c_{i:j:k}^{(\Outside)} &= c_{i:k}^{(\Outside)} \, c_{i:j}^{(\Inside)} \, \Norm{\mu_{i:k}^{(\Outside)\colB{l}}}{\colC{\TransMat_\TransVar^\top}\mu_{i:j}^{(\Inside)\colB{m}}}{\Sigma_{i:k}^{(\Outside)\colB{l}} + \colC{\TransMat_\TransVar^\top}\brackets{\Sigma_{\NonTermLabel\mathrm{L}} + \Sigma_{i:j}^{(\Inside)\colB{m}}}\colC{\TransMat_\TransVar}} \label{eq:mGzWaHXL} \\
  \mu_{i:j:k}^{(\Outside)} &= \brackets[\Big]{\parens[\big]{\Sigma_{i:k}^{(\Outside)\colB{l}}}^{\!-1}+\parens[\big]{\colC{\TransMat_\TransVar^\top}\brackets{\Sigma_{\NonTermLabel\mathrm{L}} + \Sigma_{i:j}^{(\Inside)\colB{m}}}\colC{\TransMat_\TransVar}}^{\!-1}}^{-1}
              \brackets[\Big]{
              \parens[\big]{\Sigma_{i:k}^{(\Outside)\colB{l}}}^{\!-1}\mu_{i:k}^{(\Outside)\colB{l}}
              +
              \colC{\TransMat_\TransVar^\top}\parens[\big]{\Sigma_{\NonTermLabel\mathrm{L}} + \Sigma_{i:j}^{(\Inside)\colB{m}}}^{\!-1}\mu_{i:j}^{(\Inside)\colB{m}}
              } \\
  \Sigma_{i:j:k}^{(\Outside)} &= \Sigma_{\NonTermLabel\mathrm{R}} + \brackets[\Big]{\parens[\big]{\Sigma_{i:k}^{(\Outside)\colB{l}}}^{\!-1}+\parens[\big]{\colC{\TransMat_\TransVar^\top}\brackets{\Sigma_{\NonTermLabel\mathrm{L}} + \Sigma_{i:j}^{(\Inside)\colB{m}}}\colC{\TransMat_\TransVar}}^{\!-1}}^{-1}~, \label{eq:AQuwyg6n}
\end{align}
}%
where in \eqref{eq:hdGfmqOd} we took the constant termination probability \eqref{eq:uROCbA5z} out of the integral and dropped it; in \eqref{eq:QywtiDbz} we inserted \eqref{eq:3XIqlHPD}, \eqref{eq:aHhkDVDP} and \eqref{eq:swzRuMxz};
in \eqref{eq:08DC8IOL} we applied \eqref{eq:P8MeIPHK} to marginalise out $\NonTermVar_{i:j}$; in \eqref{eq:oq6SZsqT} we used \eqref{eq:L2ANPyRd} to rewrite the transformation; in \eqref{eq:MjsgwXC8} and \eqref{eq:7tTRq6w2} we used \eqref{eq:P8MeIPHK} twice to marginalise out $\NonTermVar_{i:k}$; and in \eqref{eq:AXn5wXJa} we rewrote the final result using (\ref{eq:mGzWaHXL}--\ref{eq:AQuwyg6n}). Due to the asymmetric terms in the outside probabilities, the result is somewhat more complex than for the inside probabilities.

Analogously, the integrals in \eqref{eq:sjybKfDz} for $\NonTermVar_{j:k}$ being generated as the left child are
{
\renewcommand*{\colB}[1]{}
\begin{align}
  \iint&
         \NonTermTrans(\NonTermVar_{j:k},\NonTermVar_{k:l} \cond \NonTermVar_{j:l}) \,
         \colA{\Outside[j:l]} \, \colA{\Inside[k:l]} \,
         d\NonTermVar_{j:l} \, d\NonTermVar_{k:l} \\
  ={}&
       \colC{\sum_\TransVar w_\TransVar}
       c_{j:l}^{(\Outside)}
       c_{k:l}^{(\Inside)}
       \colB{\sum_{m,q} w_m \, w_q}
       \iint
       \Norm{\NonTermVar_{j:k}}{\colC{\TransMat_\TransVar}\NonTermVar_{j:l}}{\Sigma_{\NonTermLabel\mathrm{L}}} \,
       \Norm{\NonTermVar_{k:l}}{\NonTermVar_{j:l}}{\Sigma_{\NonTermLabel\mathrm{R}}}
       \nonumber\newlineAlign*[30mm]
       \Norm{\NonTermVar_{j:l}}{\mu_{j:l}^{(\Outside)\colB{m}}}{\Sigma_{j:l}^{(\Outside)\colB{m}}} \,
       \Norm{\NonTermVar_{k:l}}{\mu_{k:l}^{(\Inside)\colB{q}}}{\Sigma_{k:l}^{(\Inside)\colB{q}}} \,
       d\NonTermVar_{j:l} \, d\NonTermVar_{k:l} \\
  ={}&
       \colC{\sum_\TransVar w_\TransVar}
       c_{j:l}^{(\Outside)}
       c_{k:l}^{(\Inside)}
       \colB{\sum_{m,q} w_m \, w_q}
       \int
       \Norm{\NonTermVar_{j:k}}{\colC{\TransMat_\TransVar}\NonTermVar_{j:l}}{\Sigma_{\NonTermLabel\mathrm{L}}} \,
       \Norm{\NonTermVar_{j:l}}{\mu_{j:l}^{(\Outside)\colB{m}}}{\Sigma_{j:l}^{(\Outside)\colB{m}}} \,
       \Norm{\NonTermVar_{j:l}}{\mu_{k:l}^{(\Inside)\colB{q}}}{\Sigma^{(1^\prime)}} \,
       d\NonTermVar_{j:l} \\
  ={}&
       \colC{\sum_\TransVar w_\TransVar}
       c_{j:l}^{(\Outside)}
       c_{k:l}^{(\Inside)}
       \colB{\sum_{m,q} w_m \, w_q}
       \int
       \Norm{\colC{\TransMat_\TransVar^\top}\NonTermVar_{j:k}}{\NonTermVar_{j:l}}{\colC{\TransMat_\TransVar^\top}\Sigma_{\NonTermLabel\mathrm{L}}\colC{\TransMat_\TransVar}} \,
       \Norm{\NonTermVar_{j:l}}{\mu^{(2^\prime)}}{\Sigma^{(2^\prime)}}
       \Norm{\mu_{j:l}^{(\Outside)\colB{m}}}{\mu_{k:l}^{(\Inside)\colB{q}}}{\Sigma^{(3^\prime)}} \,
       d\NonTermVar_{j:l} \\
  ={}&
       \colC{\sum_\TransVar w_\TransVar}
       c_{j:l}^{(\Outside)}
       c_{k:l}^{(\Inside)}
       \colB{\sum_{m,q} w_m \, w_q}
       \Norm{\mu_{j:l}^{(\Outside)\colB{m}}}{\mu_{k:l}^{(\Inside)\colB{q}}}{\Sigma^{(3^\prime)}} \,
       \Norm{\colC{\TransMat_\TransVar^\top}\NonTermVar_{j:k}}{\mu^{(2^\prime)}}{\Sigma^{(4^\prime)}} \\
  ={}&
       \colC{\sum_\TransVar w_\TransVar}
       c_{j:l}^{(\Outside)}
       c_{k:l}^{(\Inside)}
       \colB{\sum_{m,q} w_m \, w_q}
       \Norm{\mu_{j:l}^{(\Outside)\colB{m}}}{\mu_{k:l}^{(\Inside)\colB{q}}}{\Sigma^{(3^\prime)}} \,
       \Norm{\NonTermVar_{j:k}}{\colC{\TransMat_\TransVar}\mu^{(2^\prime)}}{\colC{\TransMat_\TransVar}\Sigma^{(4^\prime)}\colC{\TransMat_\TransVar^\top}} \\
  ={}&
       \colC{\sum_\TransVar w_\TransVar} \,
       \colB{\sum_{m,q} w_m \, w_q}
       c_{j:k:l}^{(\Outside)} \,
       \Norm{\NonTermVar_{j:k}}{\mu_{j:k:l}^{(\Outside)}}{\Sigma_{j:k:l}^{(\Outside)}}
\end{align}
}%
with
{
\renewcommand*{\colB}[1]{}
\begin{align}
  \Sigma^{(1^\prime)} &= \Sigma_{\NonTermLabel\mathrm{R}} + \Sigma_{k:l}^{(\Inside)\colB{q}} &
                                                                                 \Sigma^{(2^\prime)} &= \brackets[\Big]{\parens[\big]{\Sigma_{j:l}^{(\Outside)\colB{m}}}^{\!-1}+\parens[\big]{\Sigma^{(1^\prime)}}^{\!-1}}^{-1} \\
  \Sigma^{(3^\prime)} &= \Sigma_{j:l}^{(\Outside)\colB{m}} + \Sigma^{(1^\prime)} &
  \mu^{(2^\prime)} &= \Sigma^{(2^\prime)} \brackets[\Big]{
                     \parens[\big]{\Sigma_{j:l}^{(\Outside)\colB{m}}}^{\!-1}\mu_{j:l}^{(\Outside)\colB{m}}
                     +
                     \parens[\big]{\Sigma^{(1^\prime)}}^{\!-1}\mu_{k:l}^{(\Inside)\colB{q}}
                     } \\
  \Sigma^{(4^\prime)} &= \colC{\TransMat_\TransVar^\top}\Sigma_{\NonTermLabel\mathrm{L}}\colC{\TransMat_\TransVar} + \Sigma^{(2^\prime)}
\end{align}
}%
and
{
\renewcommand*{\colB}[1]{}
\begin{align}
  c_{j:k:l}^{(\Outside)} &= c_{j:l}^{(\Outside)} \, c_{k:l}^{(\Inside)} \, \Norm{\mu_{j:l}^{(\Outside)\colB{m}}}{\mu_{k:l}^{(\Inside)\colB{q}}}{\Sigma_{j:l}^{(\Outside)\colB{m}} + \Sigma_{\NonTermLabel\mathrm{R}} + \Sigma_{k:l}^{(\Inside)\colB{q}}} \\
  \mu_{j:k:l}^{(\Outside)} &= \colC{\TransMat_\TransVar}\brackets[\Big]{\parens[\big]{\Sigma_{j:l}^{(\Outside)\colB{m}}}^{\!-1}+\parens[\big]{\Sigma_{\NonTermLabel\mathrm{R}} + \Sigma_{k:l}^{(\Inside)\colB{q}}}^{\!-1}}^{-1} \brackets[\Big]{
                     \parens[\big]{\Sigma_{j:l}^{(\Outside)\colB{m}}}^{\!-1}\mu_{j:l}^{(\Outside)\colB{m}}
                     +
                     \parens[\big]{\Sigma_{\NonTermLabel\mathrm{R}} + \Sigma_{k:l}^{(\Inside)\colB{q}}}^{\!-1}\mu_{k:l}^{(\Inside)\colB{q}}
                     } \\
  \Sigma_{j:k:l}^{(\Outside)} &= \Sigma_{\NonTermLabel\mathrm{L}} + \colC{\TransMat_\TransVar}\brackets[\Big]{\parens[\big]{\Sigma_{j:l}^{(\Outside)\colB{m}}}^{\!-1}+\parens[\big]{\Sigma_{\NonTermLabel\mathrm{R}} + \Sigma_{k:l}^{(\Inside)\colB{q}}}^{\!-1}}^{-1}\colC{\TransMat_\TransVar^\top}~.
\end{align}
}%

\subsubsection{Approximation}
\label{app:approx}

A \glsreset{gm}\acl{gm} distribution $p(x)$ with normalised mixture weights $c_i$, means $\mu_i$, and covariance matrices $\Sigma_i$ can be approximated with a single Gaussian as
\begin{align}
  \widehat{p}(x) &= \Norm{x}{\widehat{\mu}}{\widehat{\Sigma}}
  & \text{with} &&
                  \widehat{\mu} &= \sum_i c_i \, \mu_i
  & \text{and} &&
                  \widehat{\Sigma} &= \sum_i c_i \brackets[\Big]{\Sigma_i + (\mu_i - \widehat{\mu})(\mu_i - \widehat{\mu})^\top}~. \label{eq:MBXuP9Jn}
\end{align}
The approximation $\widehat{p}(x)$ matches the first and second moments of $p(x)$ and minimises the \ac{kld} $\DKL\brackets*{p(x) \ccond \widehat{p}(x)}$ \citep{orguner2007,bishop07}. This direction of the \ac{kld} is the one used e.g.~in expectation propagation, not the one used in e.g.~variational methods \citep{bishop07}. That means, $\widehat{p}(x)$ will adequately represent the support and uncertainty of $p(x)$ (e.g.~it will be non-zero wherever $p(x)$ is non-zero). On the other hand, a value of $x$ may have a high probability in $\widehat{p}(x)$ even though in $p(x)$ it has not (also see \figref{jJ4xGGDT}).

\subsubsection{Tree Induction}
\label{app:tree_induction}

Exact joint optimisation of the structure and the continuous latent variables is intractable. We therefore choose the best tree for a \ac{grbn} based the maximum of the (approximated) inside probability. Inserting \eqref{eq:uROCbA5z} and \eqref{eq:62Q4A7pD} into \eqref{eq:2Q0hmTrM}, we have
\begin{align}
  \colA{\Inside[i:k]}
  ={}&
       (1 - \TermProb)
       \sum_{\mathclap{j=i+1\hspace{1em}}}^{\mathclap{k-1}}
       \colC{\sum_\TransVar w_\TransVar} \,
       c_{i:j:k}^{(\Inside)} \,
       \Norm{\NonTermVar_{i:k}}{\mu_{i:j:k}^{(\Inside)}}{\Sigma_{i:j:k}^{(\Inside)}}~, \label{eq:ekKyUZau}
\end{align}
which is maximised by taking the mode of the Gaussian and maximising over $j$ and (if using transpositions) $\TransVar$
\begin{align}
  \max_{(\NonTermVar_{i:k},j,\TransVar)} \colC{w_\TransVar} \, c_{i:j:k}^{(\Inside)} \, \Norm{\NonTermVar_{i:k}}{\mu_{i:j:k}^{(\Inside)}}{\Sigma_{i:j:k}^{(\Inside)}}
  &= \max_{(j,\TransVar)} \colC{w_\TransVar} \, c_{i:j:k}^{(\Inside)} \, \det*{2\pi\Sigma_{i:j:k}^{(\Inside)}}^{-\frac{1}{2}} \label{eq:oJrD1TsI}
\end{align}
If we have multi-terminal transitions (or more generally other possible transitions), we also have to maximise over the different possible transitions. For each non-terminal variable, we compute and store the best choice during bottom-up computations of the inside probabilities. Afterwards, we can construct the best tree by starting at the root node and recursively picking the best structure top-down.

\section{Example}
\label{app:example}

\newcommand{\StartEndToCenterWidth}[2]{
  \pgfmathsetmacro{\CenterCoordUnscaled}{(#1 + #2)/2}
  \pgfmathsetmacro{\CenterCoord}{sqrt(2)*\CenterCoordUnscaled}
  \pgfmathsetmacro{\WidthCoordUnscaled}{#2 - #1}
  \pgfmathsetmacro{\WidthCoord}{sqrt(2)*\WidthCoordUnscaled/2}
}

\NewDocumentCommand{\DrawDiamond}{ O{} O{\CenterCoord,\WidthCoord} O{\SqrtTwo} }{
  \draw[#1] ($(#2)+(#3,0)$) -- ++(-#3,#3) -- ++(-#3,-#3) -- ++(#3,-#3) -- cycle;
}

\NewDocumentCommand{\DrawSquare}{ O{} O{\CenterCoord,\WidthCoord} O{0.9*\SqrtTwo} }{
  \draw[#1] ($(#2)+(#3,#3)$) -- ++(-#3-#3,0) -- ++(0,-#3-#3) -- ++(#3+#3,0) -- cycle;
}

\NewDocumentCommand{\FillChart}{ s m o o }{
  \IfNoValueTF{#3}{
    \def\MinStart{0}
  }{
    \def\MinStart{#3}
  }
  \IfNoValueTF{#4}{
    \def\MaxEnd{\SequenceLength}
  }{
    \def\MaxEnd{#4}
  }
  \pgfmathsetmacro{\MaxEndMinusOne}{int(\MaxEnd-1)}
  \foreach \Start in {\MinStart,...,\MaxEndMinusOne} {
    \pgfmathsetmacro{\StartPlusOne}{int(\Start+1)}
    \foreach \End in {\StartPlusOne,...,\MaxEnd} {
      \StartEndToCenterWidth{\Start}{\End}
      \IfBooleanTF{#1}{
        \ifthenelse{\equal{\Start}{\MinStart}\AND\equal{\End}{\MaxEnd}}{}{#2}
      }{
        #2
      }
    }
  }
}

\newcommand*{\FillSequence}[1]{
  \pgfmathsetmacro{\SequenceLengthMinusOne}{int(\SequenceLength-1)}
  \foreach \Start in {0,...,\SequenceLengthMinusOne} {
    \pgfmathsetmacro{\End}{int(\Start+1)}
    \StartEndToCenterWidth{\Start}{\End}
    #1
  }
}

\newcommand*{\AddBorderLabels}{
  \pgfmathsetmacro{\SequenceLengthMinusOne}{int(\SequenceLength-1)}
  \foreach \End in {1,...,\SequenceLength} {
    \StartEndToCenterWidth{0}{\End}
    \node at ($(\CenterCoord,\WidthCoord)+0.8*(-\SqrtTwo,\SqrtTwo)$) {\End};
  }
  \foreach \Start in {0,...,\SequenceLengthMinusOne} {
    \StartEndToCenterWidth{\Start}{\SequenceLength}
    \node at ($(\CenterCoord,\WidthCoord)+0.8*(\SqrtTwo,\SqrtTwo)$) {\Start};
  }
  \pgfmathsetmacro{\End}{\SequenceLength/2+0.5}
  \StartEndToCenterWidth{0}{\End}
  \node at ($(\CenterCoord,\WidthCoord)+1.2*(-\SqrtTwo,\SqrtTwo)$) {\rotatebox{45}{span end}};
  \pgfmathsetmacro{\Start}{\SequenceLength/2-0.5}
  \StartEndToCenterWidth{\Start}{\SequenceLength}
  \node at ($(\CenterCoord,\WidthCoord)+1.2*(\SqrtTwo,\SqrtTwo)$) {\rotatebox{-45}{span start}};
}

\NewDocumentCommand{\DrawNonTermTrans}{ m m m O{lat_var,emphnode} O{edge,emphedge} }{
  \node[#4] at (NonTerm_#1_#2) {\DrawNonTermLabel{#1}{#2}};
  \node[#4] at (NonTerm_#2_#3) {\DrawNonTermLabel{#2}{#3}};
  \draw[#5] (NonTerm_#1_#3) -- (NonTerm_#1_#2);
  \draw[#5] (NonTerm_#1_#3) -- (NonTerm_#2_#3);
}
\NewDocumentCommand{\DrawTermTrans}{ m O{obs_var,emphnode} O{edge,emphedge} }{
  \pgfmathsetmacro{\IgiBQkByXAtQliEBLbJD}{int(#1-1)}
  \node[#2] at (Term_#1) {\DrawTermLabel{#1}};
  \draw[#3] (NonTerm_\IgiBQkByXAtQliEBLbJD_#1) -- (Term_#1);
}
\NewDocumentCommand{\EmphNonTerminal}{ m m O{lat_var,emphnode} }{
  \node[#3] at (NonTerm_#1_#2) {\DrawNonTermLabel{#1}{#2}};
}
\NewDocumentCommand{\EmphTerminal}{ m O{obs_var,emphnode} }{
  \node[#2] at (Term_#1) {\DrawTermLabel{#1}};
}
\NewDocumentCommand{\BoxChart}{ O{opacity=0} O{1.2*\SqrtTwo} O{1.2*\SqrtTwo} O{1.2*\SqrtTwo+1.3} O{1.2*\SqrtTwo} O{0} O{\SequenceLength} }{
  \pgfmathsetmacro{\txJgFTHkOHLNcqOUUCHT}{int(#6+1)}
  \pgfmathsetmacro{\kZtfxeUgWGESuOlkJEhs}{int(#7-1)}
  \StartEndToCenterWidth{#6}{\txJgFTHkOHLNcqOUUCHT}
  \coordinate (BottomLeft) at ({\CenterCoord-(#2)},{\WidthCoord-(#4)});
  \StartEndToCenterWidth{\kZtfxeUgWGESuOlkJEhs}{#7}
  \coordinate (BottomRight) at ({\CenterCoord+(#3)},{\WidthCoord-(#4)});
  \StartEndToCenterWidth{#6}{#7}
  \coordinate (Top) at (\CenterCoord,{\WidthCoord+(#5)});
  \draw[#1] (BottomLeft) -- (BottomRight) |- (Top) -| cycle;
}

\newcommand*{\DrawNonTermLabelWithIndex}[2]{$\NonTermVar_{#1:#2}$}
\newcommand*{\DrawNonTermLabelWithoutIndex}[2]{$\NonTermVar$}
\newcommand*{\DrawTermLabelWithIndex}[1]{$\TermVar_{#1}$}
\newcommand*{\DrawTermLabelWithoutIndex}[1]{$\TermVar$}

\newcommand*{\DrawNonTermLabel}{\DrawNonTermLabelWithIndex}
\newcommand*{\DrawTermLabel}{\DrawTermLabelWithIndex}

\definecolor{almostwhite}{HTML}{FEFEFE}
\definecolor{mygrey}{HTML}{DDDDDD}

\tikzstyle{var}=[circle,draw,inner sep=0pt,minimum size=22pt]
\tikzstyle{lat_var}=[var,fill=almostwhite]
\tikzstyle{obs_var}=[var,fill=mygrey]
\tikzstyle{ghost_var}=[var,fill=almostwhite,draw=mygrey,text=mygrey]
\tikzstyle{factor}=[draw,fill=black,inner sep=0pt,minimum size=5pt]

\tikzstyle{edge}=[->,>=stealth]
\tikzstyle{con}=[-*,shorten >=-2.5pt]

\tikzstyle{plate}=[draw,rounded corners,inner sep=13pt]
\tikzstyle{gate}=[draw,dashed,inner sep=5pt]

\tikzstyle{emph}=[line width=1pt,draw=RoyalBlue]
\tikzstyle{emphnode}=[emph]
\tikzstyle{emphedge}=[emph]

\def\SequenceLength{4}

\pgfmathsetmacro{\SqrtTwo}{sqrt(2)/2}

\begin{figure}
  \centering
  \begin{tikzpicture}[scale=1]
    \BoxChart
    \FillChart{
      \DrawDiamond[fill=almostwhite,draw=mygrey]
      \node[ghost_var] (NonTerm_\Start_\End) at (\CenterCoord, \WidthCoord) {\DrawNonTermLabel{\Start}{\End}};
    }
    \FillSequence{
      \node[obs_var,emphnode,minimum size=7ex] (Term_\End) at (\CenterCoord, \WidthCoord-1.5) {$\TermVar_{\End}{=}\ifthenelse{\equal{\End}{4}}{0}{\Start}$};
    }
    \EmphNonTerminal{0}{4}
    \DrawNonTermTrans{0}{3}{4}
    \DrawNonTermTrans{0}{1}{3}
    \DrawNonTermTrans{1}{2}{3}
    \FillSequence{\DrawTermTrans{\End}[opacity=0]}
  \end{tikzpicture}
  \caption{Parse chart for a sequence of length $n=4$ with best tree estimate (see text for details).}
  \label{fig:example_chart}
\end{figure}
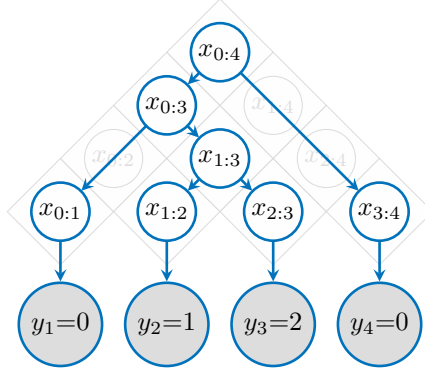

In this section, we present the complete calculations for the inside probabilities, the tree estimate, and the marginal likelihood for a basic \ac{grbn} (no transpositions or multi-terminal transitions) on a simple example sequence of length $n=4$ (also see \figref{example_chart}).
We assume parameters
\begin{align}
  \mu_\PriorLabel&=0 & \Sigma_\PriorLabel&=\Sigma_{\NonTermLabel\mathrm{L}}=\Sigma_{\NonTermLabel\mathrm{R}}=\Sigma_\TermLabel=1 & \TermProb&=1/2
\end{align}
in (\ref{eq:Bg7E21Fg}--\ref{eq:uROCbA5z}) and a scalar sequence
\begin{align}
  \TermVarBlock = (\TermVar_1, \TermVar_2, \TermVar_3, \TermVar_4) = (0, 1, 2, 0)~.
\end{align}

\subsection{Preliminaries}

Inside probabilities are approximated with a single Gaussian
\begin{align}
  \colA{\Inside[i:k]}
  &\approx c_{i:k}^{(\Inside)} \, \Norm{\NonTermVar_{i:k}}{\mu_{i:k}^{(\Inside)}}{\Sigma_{i:k}^{(\Inside)}}~, \tag{\ref{eq:3XIqlHPD}}
\end{align}
specified by $c_{i:k}^{(\Inside)}$, $\mu_{i:k}^{(\Inside)}$, and $\Sigma_{i:k}^{(\Inside)}$, which are the relevant quantities to be computed.

At the bottom level, we use \eqref{eq:fDbELzNN} for the base case and insert \eqref{eq:6JjalCAN} and \eqref{eq:uROCbA5z} to obtain
\begin{align}
  \colA{\Inside[i:i\shortplus1]}
  ={}& \TermProb \, \Norm*{\TermVar_{i\shortplus1}}{\NonTermVar_{i:i\shortplus1}}{\Sigma_\TermLabel}~, \label{eq:wQzJ249Z}
\end{align}
where we can directly read off $c_{i:k}^{(\Inside)}$, $\mu_{i:k}^{(\Inside)}$, and $\Sigma_{i:k}^{(\Inside)}$.

For the higher levels, we have to use \eqref{eq:2Q0hmTrM} for the recursive case, where inserting \eqref{eq:62Q4A7pD} to solve the integrals in closed form gives
\begin{align}
  \colA{\Inside[i:k]}
  ={}&
       (1-\TermProb) \,
       \sum_{\mathclap{j=i+1\hspace{1em}}}^{\mathclap{k-1}}
       c_{i:j:k}^{(\Inside)} \,
       \Norm{\NonTermVar_{i:k}}{\mu_{i:j:k}^{(\Inside)}}{\Sigma_{i:j:k}^{(\Inside)}} \label{eq:iVoZRgeO}
\end{align}
with parameters given by (\ref{eq:OuIhYpEh}--\ref{eq:1fhmMSM1}) as
\begin{align}
  c_{i:j:k}^{(\Inside)}
  ={}& c_{i:j}^{(\Inside)} \,
       c_{j:k}^{(\Inside)} \,
       \Norm{\mu_{i:j}^{(\Inside)}}{\mu_{j:k}^{(\Inside)}}{1+\Sigma_{i:j}^{(\Inside)}+1+\Sigma_{j:k}^{(\Inside)}} \label{eq:zGWz8B6J} \\
  \mu_{i:j:k}^{(\Inside)}
  ={}& \Sigma_{i:j:k}^{(\Inside)} \brackets*{
       \parens*{1 + \Sigma_{i:j}^{(\Inside)}}^{-1}
       \mu_{i:j}^{(\Inside)}
       +
       \parens*{1 + \Sigma_{j:k}^{(\Inside)}}^{-1}
       \mu_{j:k}^{(\Inside)}
       } \\
  \Sigma_{i:j:k}^{(\Inside)}
  ={}& \brackets*{
       \parens*{1 + \Sigma_{i:j}^{(\Inside)}}^{-1}
       +
       \parens*{1 + \Sigma_{j:k}^{(\Inside)}}^{-1}
       }^{-1}~, \label{eq:ai15QxkH}
\end{align}
where we already inserted $\Sigma_{\NonTermLabel\mathrm{L}}=\Sigma_{\NonTermLabel\mathrm{R}}=1$.

If the sum in \eqref{eq:iVoZRgeO} has only a single term, we immediately get
\begin{align}
  c_{i:k}^{(\Inside)} &= (1-\TermProb) \, c_{i:j:k}^{(\Inside)}
  & \mu_{i:k}^{(\Inside)} &= \mu_{i:j:k}^{(\Inside)}
  & \Sigma_{i:k}^{(\Inside)} &= \Sigma_{i:j:k}^{(\Inside)}~. \label{eq:Ee31cgLj}
\end{align}
If there is more than one term in the sum in \eqref{eq:iVoZRgeO}, this means that there are multiple splitting options that are marginalised out and we therefore need to do two things.

First, we have to identify the best splitting option to be able to compute the tree estimate. This is done by using \eqref{eq:oJrD1TsI} and comparing the values of
\begin{align}
  \frac{c_{i:j:k}^{(\Inside)}}{\sqrt{\Sigma_{i:j:k}^{(\Inside)}}}~, \label{eq:4u62IdQI}
\end{align}
where $\det*{\Sigma_{i:j:k}^{(\Inside)}}=\Sigma_{i:j:k}^{(\Inside)}$ in the scalar case and we left out shared constant factors.

Second, we have to approximate the resulting mixture with a single Gaussian using \eqref{eq:MBXuP9Jn}, where the mixture weights have to be normalised. For the univariate/scalar case considered here, we then get
\begin{align}
  \mu_{i:k}^{(\Inside)} &= \frac{\sum_{j=i+1}^{k-1}  c_{i:j:k}^{(\Inside)} \, \mu_{i:j:k}^{(\Inside)}}{\sum_{j=i+1}^{k-1} c_{i:j:k}^{(\Inside)}} \label{eq:vTIYC7ZO} \\
  \Sigma_{i:k}^{(\Inside)} &= \frac{\sum_{j=i+1}^{k-1} c_{i:j:k}^{(\Inside)} \brackets[\Big]{\Sigma_{i:j:k}^{(\Inside)} + (\mu_{i:j:k}^{(\Inside)} - \mu_{i:k}^{(\Inside)})^2}}{\sum_{j=i+1}^{k-1} c_{i:j:k}^{(\Inside)}} \\
  c_{i:k}^{(\Inside)} &= (1-\TermProb) \sum_{j=i+1}^{k-1} c_{i:j:k}^{(\Inside)}~. \label{eq:PTjQVesn}
\end{align}

Finally, the marginal likelihood \eqref{eq:ScKTv9k1} is obtained as
\begin{align}
  p(\TermVarBlock)
  &= \int \colA{\Inside[0:n]} \, \PriorDist(\NonTermVar_{0:n}) \, d\NonTermVar_{0:n} \\
  &\approx \int c_{0:n}^{(\Inside)} \, \Norm{\NonTermVar_{0:n}}{\mu_{0:n}^{(\Inside)}}{\Sigma_{0:n}^{(\Inside)}} \, \Norm{\NonTermVar_{0:n}}{0}{1} \, d\NonTermVar_{0:n} \\
  &= \int c_{0:n}^{(\Inside)} \, \Norm{\mu_{0:n}^{(\Inside)}}{0}{\Sigma_{0:n}^{(\Inside)}+1} \, \Norm{\NonTermVar_{0:n}}{\bar\mu}{\bar\Sigma} \, d\NonTermVar_{0:n} \\
  &= c_{0:n}^{(\Inside)} \, \Norm{\mu_{0:n}^{(\Inside)}}{0}{\Sigma_{0:n}^{(\Inside)}+1} \label{eq:AMYB4ozh}
\end{align}
where we have used \eqref{eq:P8MeIPHK} to rewrite the product of Gaussians and inserted $\mu_\PriorLabel=0$ and $\Sigma_\PriorLabel=1$.

The normal distribution is defined as
\begin{align}
  \Norm{x}{\mu}{\Sigma}
  &= \frac{1}{\sqrt{2\pi\det*{\Sigma}}}\exp\brackets*{-\frac{1}{2}(x-\mu)^\top\Sigma^{-1}(x-\mu)} \\
  &= \frac{1}{\sqrt{2\pi\Sigma}}\exp\brackets*{-\frac{1}{2}\frac{(x-\mu)^2}{\Sigma}}~,
\end{align}
where the second line is for the scalar case.

\subsection{Calculations}

We start with the inside probabilities at the bottom level for the latent variables $\NonTermVar_{0:1}$, $\NonTermVar_{1:2}$, $\NonTermVar_{2:3}$, $\NonTermVar_{3:4}$ and from \eqref{eq:wQzJ249Z} we read off (without any approximations)
\begin{align}
  c_{i:i+1}^{(\Inside)} &= \TermProb = 1/2
  & \mu_{i:i+1}^{(\Inside)} &= \TermVar_{i+1}
  & \Sigma_{i:i+1}^{(\Inside)} &= \Sigma_\TermLabel = 1
\end{align}
with
\begin{align}
  \mu_{0:1}^{(\Inside)}&=0 & \mu_{1:2}^{(\Inside)}&=1 & \mu_{2:3}^{(\Inside)}&=2 & \mu_{3:4}^{(\Inside)}&=0~.
\end{align}

Next, we compute the inside probabilities on the first level for the variables $\NonTermVar_{0:2}$, $\NonTermVar_{1:3}$, $\NonTermVar_{2:4}$. The only possible splitting option is for $j=i+1$ and from \eqref{eq:Ee31cgLj} we get (again without approximation)
{
  \renewcommand*{\colB}[1]{}
  \renewcommand*{\colC}[1]{}
  \begin{align}
    c_{i:i+2}^{(\Inside)}
    ={}& (1-\TermProb) \,
         \TermProb^2 \,
         \Norm{\TermVar_{i+1}}{\TermVar_{i+2}}{4},
    & \mu_{i:i+2}^{(\Inside)\colB{lm}} ={}& \parens{\TermVar_{i+1}+\TermVar_{i+2}}/2,
    & \Sigma_{i:i+2}^{(\Inside)\colB{lm}} ={}& 1
  \end{align}
}%
and hence
\begin{align}
  c_{0:2}^{(\Inside)} &= \frac{1}{2^4e^\frac{1}{8}\sqrt{2\pi}} \approx 2.20\cdot10^{-2}
  & \mu_{0:2}^{(\Inside)} &= 0.5
  & \Sigma_{0:2}^{(\Inside)} &= 1 \\
  c_{1:3}^{(\Inside)} &= \frac{1}{2^4e^\frac{1}{8}\sqrt{2\pi}} \approx 2.20\cdot10^{-2}
  & \mu_{1:3}^{(\Inside)} &= 1.5
  & \Sigma_{1:3}^{(\Inside)} &= 1 \\
  c_{2:4}^{(\Inside)} &= \frac{1}{2^4e^\frac{1}{2}\sqrt{2\pi}} \approx 1.51\cdot10^{-2}
  & \mu_{2:4}^{(\Inside)} &= 1
  & \Sigma_{2:4}^{(\Inside)} &= 1~.
\end{align}

Turning to the values for $\NonTermVar_{0:3}$ and $\NonTermVar_{1:4}$, we now have two terms in the sum in \eqref{eq:iVoZRgeO}, which means that we need to evaluate the best split and approximate the mixture. The corresponding parameters of the mixture are given by (\ref{eq:zGWz8B6J}--\ref{eq:ai15QxkH}) as
{
  \renewcommand*{\colB}[1]{}
  \renewcommand*{\colC}[1]{}
  \begin{align}
    c_{i:j:k}^{(\Inside)}
    ={}& c_{i:j}^{(\Inside)} \,
         c_{j:k}^{(\Inside)} \,
         \Norm{\colC{\TransMat_\TransVar^\top}\mu_{i:j}^{(\Inside)\colB{l}}}{\mu_{j:k}^{(\Inside)\colB{m}}}{4}~,
    & \mu_{i:j:k}^{(\Inside)\colB{lm}}
    ={}& (\mu_{i:j}^{(\Inside)} + \mu_{j:k}^{(\Inside)})/2~,
    & \Sigma_{i:j:k}^{(\Inside)\colB{lm}}
    ={}& 1~,
  \end{align}
}%
which results in
{
  \renewcommand*{\colB}[1]{}
  \renewcommand*{\colC}[1]{}
  \begin{align}
    c_{0:1:3}^{(\Inside)}
    &= \frac{1}{2^7\pi e^\frac{13}{32}}
      \approx 1.66\cdot10^{-3}
    &\mu_{0:1:3}^{(\Inside)} &=
                               3/4
    & \Sigma_{0:1:3}^{(\Inside)} &= 1 \\
    c_{0:2:3}^{(\Inside)}
    &= \frac{1}{2^7\pi e^\frac{13}{32}}
      \approx 1.66\cdot10^{-3}
    &\mu_{0:2:3}^{(\Inside)} &=
                               5/4
    & \Sigma_{0:2:3}^{(\Inside)} &= 1
  \end{align}
  and
  \begin{align}
    c_{1:2:4}^{(\Inside)}
    &= \frac{1}{2^7\pi e^\frac{1}{2}}
      \approx 1.51\cdot10^{-3}
    &\mu_{1:2:4}^{(\Inside)} &=
                               1
    & \Sigma_{1:2:4}^{(\Inside)} &= 1 \\
    c_{1:3:4}^{(\Inside)}
    &= \frac{1}{2^7\pi e^\frac{13}{32}}
      \approx 1.66\cdot10^{-3}
    &\mu_{1:3:4}^{(\Inside)} &=
                               3/4
    & \Sigma_{1:3:4}^{(\Inside)} &= 1~.
  \end{align}
}%

To identify the best split for each variable based on \eqref{eq:4u62IdQI}, we see (all variances are equal) from
\begin{align}
  c_{0:1:3}^{(\Inside)} &= c_{0:2:3}^{(\Inside)} & \text{and} && c_{1:2:4}^{(\Inside)} &< c_{1:3:4}^{(\Inside)}~,
\end{align}
that for $\NonTermVar_{0:3}$ both splits are equally well and for $\NonTermVar_{0:3}$ the split $\NonTermVar_{1:4}\rightarrow(\NonTermVar_{1:3}, \NonTermVar_{3:4})$ at $j=3$ is better. This is intuitively clear, since generating $(\TermVar_2,\TermVar_3)=(1,2)$ from the same non-terminal variable $\NonTermVar_{1:3}=1.5$ is more likely than generating $(\TermVar_3,\TermVar_4)=(2,0)$ from $\NonTermVar_{2:4}=1$, given that in both cases the values are generated from a Gaussian with variance 1.

We approximate the mixtures with a single Gaussian with parameters given by (\ref{eq:vTIYC7ZO}--\ref{eq:PTjQVesn}) as
\begin{align}
  c_{0:3}^{(\Inside)} &\approx 1.66\cdot10^{-3}
  & \mu_{0:3}^{(\Inside)}
  &= 1
  & \Sigma_{0:3}^{(\Inside)} &= \frac{17}{16} \\
  c_{1:4}^{(\Inside)} &\approx 1.58\cdot10^{-3}
  & \mu_{1:4}^{(\Inside)}
  &\approx 0.869
  & \Sigma_{1:4}^{(\Inside)} &\approx 1.016~.
\end{align}

Finally, we have the inside probability for the root variable $\NonTermVar_{0:4}$ with three terms in the sum in \eqref{eq:iVoZRgeO} with parameters
{
  \begin{align}
    c_{i:j:k}^{(\Inside)}
    ={}& c_{i:j}^{(\Inside)} \,
          c_{j:k}^{(\Inside)} \,
          \Norm{\mu_{i:j}^{(\Inside)}}{\mu_{j:k}^{(\Inside)}}{1+\Sigma_{i:j}^{(\Inside)}+1+\Sigma_{j:k}^{(\Inside)}} \\
    \mu_{i:j:k}^{(\Inside)}
    ={}& \Sigma_{i:j:k}^{(\Inside)} \brackets*{
          \parens*{1 + \Sigma_{i:j}^{(\Inside)}}^{-1}
          \mu_{i:j}^{(\Inside)}
          +
          \parens*{1 + \Sigma_{j:k}^{(\Inside)}}^{-1}
          \mu_{j:k}^{(\Inside)}
          } \\
    \Sigma_{i:j:k}^{(\Inside)}
    ={}& \brackets*{
          \parens*{1 + \Sigma_{i:j}^{(\Inside)}}^{-1}
          +
          \parens*{1 + \Sigma_{j:k}^{(\Inside)}}^{-1}
          }^{-1}
  \end{align}
}
and hence
\begin{align}
  c_{0:1:4}^{(\Inside)} &\approx 1.43\cdot10^{-4}
  & \mu_{0:1:4}^{(\Inside)} &\approx 0.433
  &\Sigma_{0:1:4}^{(\Inside)} &\approx 1.004 \\
  c_{0:2:4}^{(\Inside)} &\approx 6.38\cdot10^{-5}
  & \mu_{0:2:4}^{(\Inside)} &= 0.75
  &\Sigma_{0:2:4}^{(\Inside)} &= 1 \\
  c_{0:3:4}^{(\Inside)} &\approx 1.45\cdot10^{-4}
  & \mu_{0:3:4}^{(\Inside)} &\approx 0.492
  &\Sigma_{0:3:4}^{(\Inside)} &\approx 1.015~.
\end{align}

For the splitting options, \eqref{eq:4u62IdQI} gives
\begin{align}
  \frac{c_{0:1:4}^{(\Inside)}}{\sqrt{\Sigma_{0:1:4}^{(\Inside)}}} &\approx 1.427\cdot10^{-4}
  & \frac{c_{0:2:4}^{(\Inside)}}{\sqrt{\Sigma_{0:2:4}^{(\Inside)}}} &\approx 6.38\cdot10^{-5}
  & \frac{c_{0:3:4}^{(\Inside)}}{\sqrt{\Sigma_{0:3:4}^{(\Inside)}}} &\approx 1.439\cdot10^{-4}
\end{align}
and we see that the split $\NonTermVar_{0:4}\rightarrow(\NonTermVar_{0:3}, \NonTermVar_{3:4})$ for $j=3$ is the best one. Intuitively, this makes sense because it splits between $\TermVar_3$ and $\TermVar_4$, which is the biggest step. Not we have only minor differences between the split options, because for simplicity we have chosen our variance parameters with a value of 1, which is relatively large compared to the spread of the values. Choosing smaller variances would result in more prominent splitting preferences.

We can now construct the full tree by also picking the best split for $\NonTermVar_{0:3}$, which is a tie between splitting at $j=1$ and $j=2$, so we can choose either one (in practice one might consider random tie breaking to avoid biases due to variable order). The resulting tree is shown in \figref{example_chart}.

The parameters for the inside probability of $\NonTermVar_{0:4}$, given by approximating the three \acl{gm} components, are
\begin{align}
  c_{0:4}^{(\Inside)} &\approx 1.76\cdot10^{-4}
  & \mu_{0:4}^{(\Inside)} &= 0.515
  & \Sigma_{0:4}^{(\Inside)} &= 1.021~.
\end{align}
Based on \eqref{eq:AMYB4ozh} this results in a marginal likelihood of
\begin{align}
  p(\TermVarBlock) &\approx 4.63\cdot10^{-5}~.
\end{align}

\section{Experiments}

\subsection{Details for Quantitative Evaluation}
\label{app:lSWGmWpX}

\begin{figure}[tp!]
  \centering
  \includegraphics[width=0.485\linewidth]{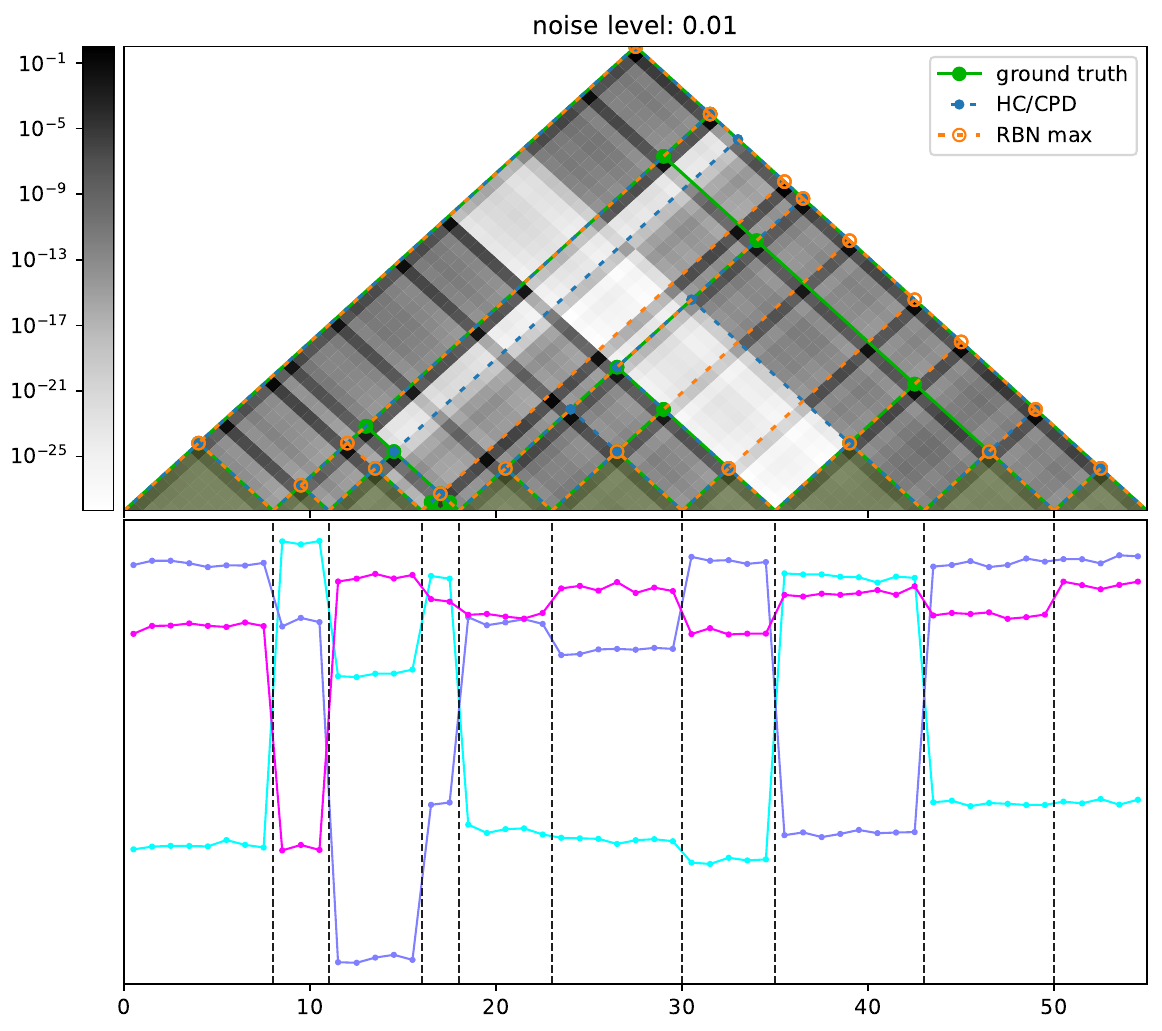}
  \includegraphics[width=0.5\linewidth]{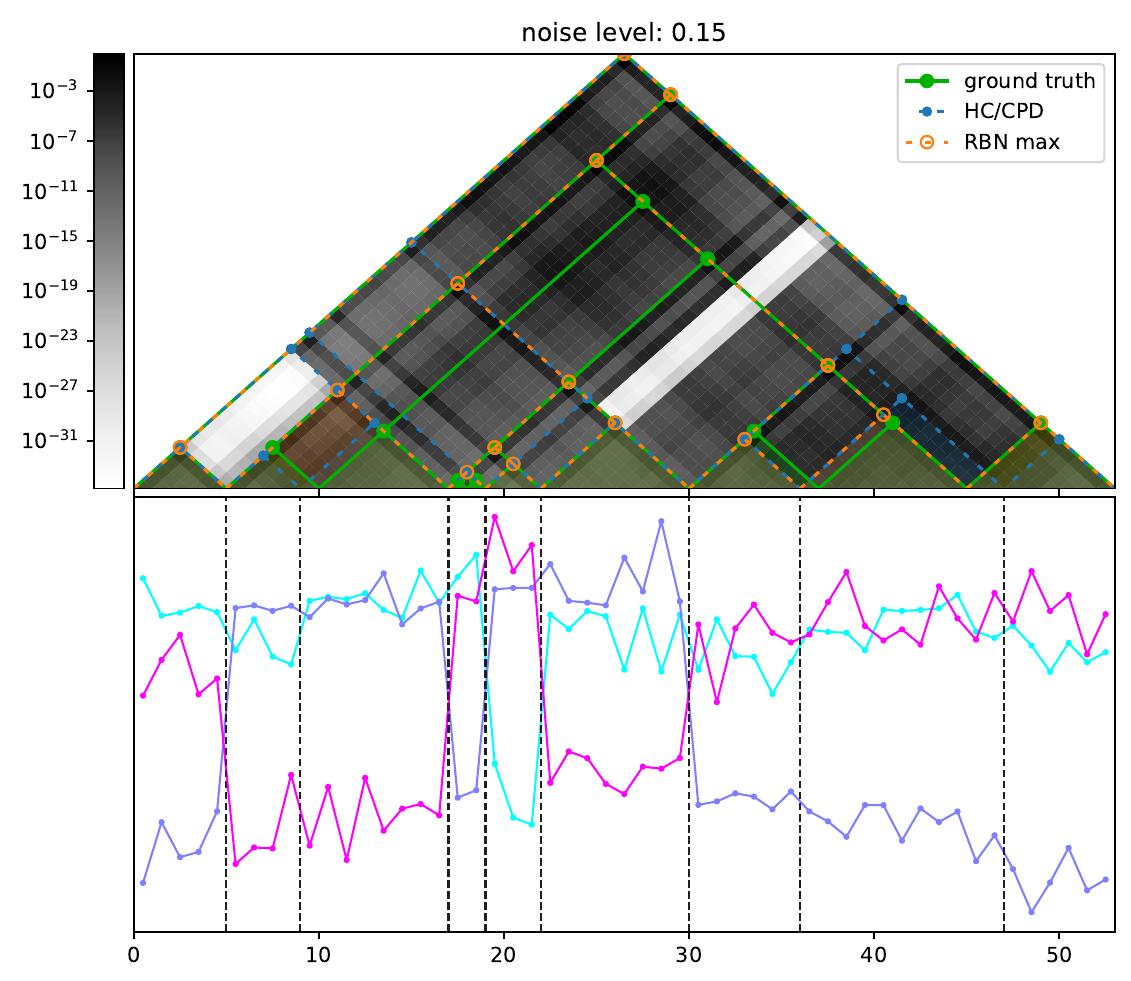}
  \caption{Example of the synthetic data used in the quantitative evaluation with noise levels of 0.01 (left) and 0.15 (right). \textbf{(bottom):}~Generated three-dimensional time series; vertical dashed lines indicate the segments identified by the \glsreset{cpd}\ac{cpd} method. \textbf{(top):}~The ground-truth tree (green), the tree estimate from \glsreset{hc}\ac{hc} based on the \ac{cpd} segmentation (blue), and the tree estimate of the \ac{rbn} (orange). The grey scale indicates the marginal node probabilities based on the \ac{rbn}.}
  \label{fig:PVpySY8K}
\end{figure}

\begin{figure}[tp!]
  \centering
  \includegraphics[width=\linewidth,angle=0]{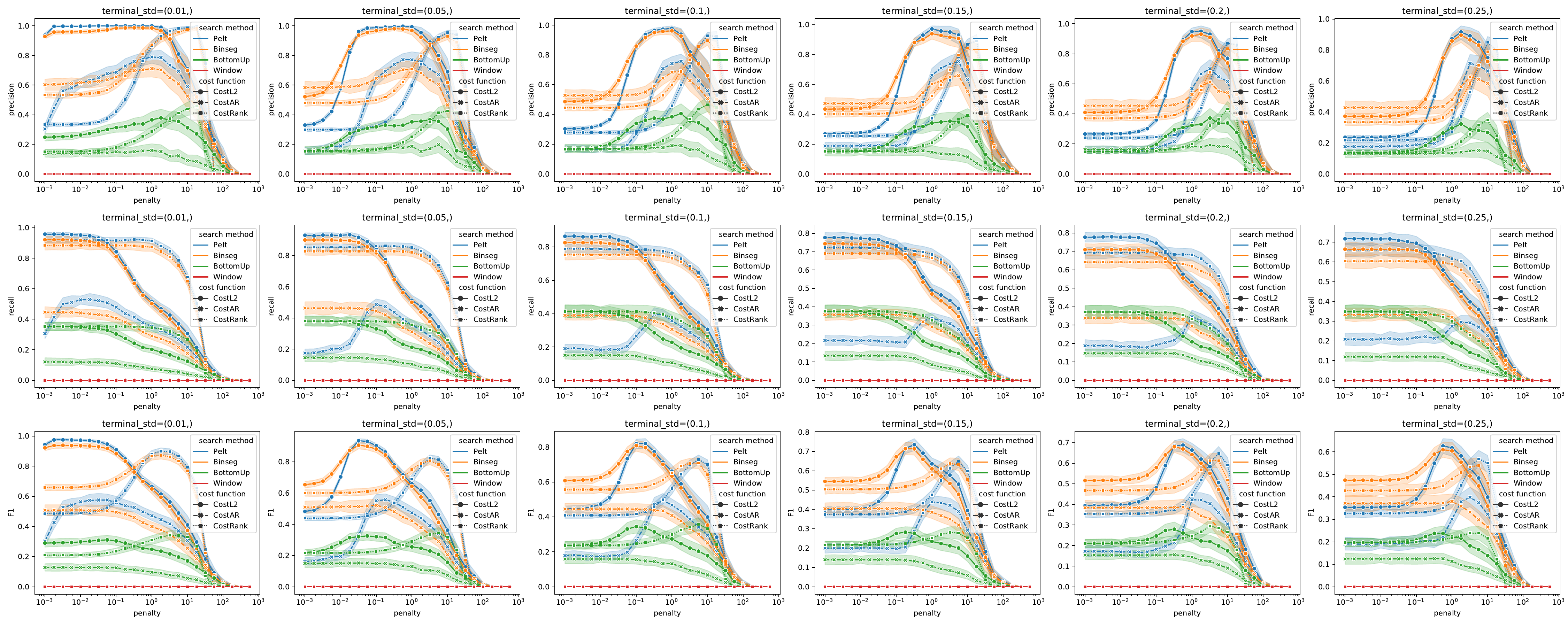}
  \caption{Grid search for best parameters and \ac{cpd} method, based on the F1 score w.r.t.~the ground-truth change points. The combination of \texttt{Pelt} search method with \texttt{CostL2} cost function performed best (with optimal parameters selected) in all cases.}
  \label{fig:xrXXXt9K}
\end{figure}

We performed a quantitative evaluation on synthetic data for the task of segmenting a noisy time series and inferring the underlying tree. We used the \acl{grbn} for music (\secref{grbn_music}) with some simplifications: 1)~data were continuous and not categorical, 2)~they had only three dimensions instead of twelve, 3)~the prior distribution did not have any transpositions, 4)~the left child could only be transposed by zero or one step. The data were sampled from this model, using zero prior mean, $\Sigma_{\mathrm{p}}=\mathbb{1}$, $\Sigma_{\NonTermLabel\mathrm{L}}=\Sigma_{\NonTermLabel\mathrm{R}}=\mathbb{1}\cdot0.1^2$, $\lambda=5$, equal weights $W_0=W_1=0.5$ for transposition by zero or one, and $\Sigma_\TermLabel=\mathbb{1}\cdot\mathtt{noise}^2$ with different noise levels $\{0.01, 0.05, 0.1, 0.15, 0.2, 0.25\}$. We used a terminal probability of $\TermProb=0.6$ for sampling and rejected any sequences with a length outside the range of 50--55. An example of the data is shown in \figref{PVpySY8K}.

For comparison, we used the best-performing \glsreset{cpd}\ac{cpd} method from the \texttt{ruptures} library \citep{truong2020} for segmenting the time series, combined with bottom-up \glsreset{hc}\ac{hc} for inferring the tree structure (``HC/CPD''). For each noise level, we selected the \ac{cpd} method and parameters with best F1 score based on the ground-truth segments of 100 training sequences (see \figref{xrXXXt9K}). In \ac{hc}, pairs of adjacent segments with the smallest Euclidean/L2 distance between their mean values were successively combined to construct the tree. For the \ac{rbn}, all parameters were trained from scratch by minimising the marginal likelihood of the observations of only 10 training sequences (i.e.~no ground-truth information and less training data than for the baseline was used), separately for each noise level.

The models were evaluated on 500 test sequences by computing their precision and recall w.r.t.~the ground-truth trees. Each possible node was treated as a separate binary classification task and the results reflect the number of correctly predicted nodes. Note that due to the strongly unbalanced class distribution (many more possible node locations than actual nodes in the tree) precision and recall or the combined F1 score are the appropriate performance metrics (as opposed to e.g.~accuracy). For the \ac{rbn} they were computed in two different ways: 1)~based on the best-tree estimate (``RBN max'') and 2)~based on the marginal node probabilities \eqref{eq:NuTG01QR} (``RBN marginal'').

Precision and recall  are computed from the true positive (TP), false positive (FP), and false negative (FN) rates
\begin{align}
  \mathrm{recall} &= \frac{\mathrm{TP}}{\mathrm{TP} + \mathrm{FN}} \\
  \mathrm{precision} &= \frac{\mathrm{TP}}{\mathrm{TP} + \mathrm{FP}} \\
  F1 &= 2 \frac{\mathrm{precision} \cdot \mathrm{recall}}{\mathrm{precision} + \mathrm{recall}}
\end{align}
For the single-tree estimates (baseline model and  best-tree estimate from \acp{rbn}) we compared the ground-truth and estimated tree node-by-node to count correctly predicted nodes (TP), nodes that are in the prediction but not the ground-truth (FP), and nodes that are in the ground-truth but not the prediction (FN). For the marginal node  probabilities, we computed the corresponding rates by counting all nodes in the ground-truth tree (TP+FN), summing the marginal probabilities  over all nodes in the ground-truth tree (TP), and summing the marginal probabilities over all possible nodes (TP+FP).

\subsection{Hierarchical Music Analysis}
\label{app:music}

\begin{figure*}[tp!]
  \centering
  \includegraphics[width=\linewidth]{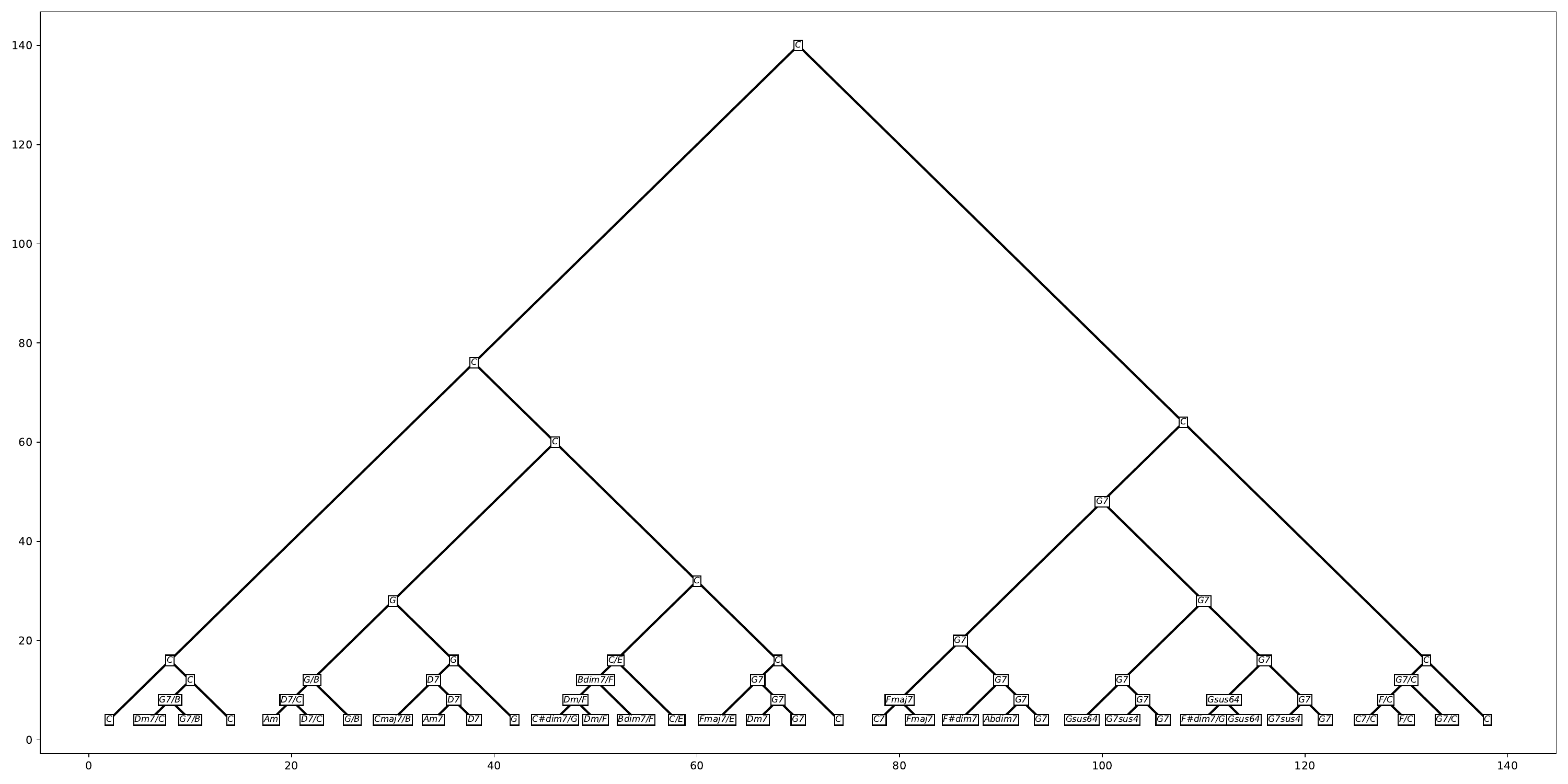}
  \caption{Harmonic analysis of Johann Sebastian Bach's Prelude No.~1 in C major, BWV 846}
  \label{fig:hYxyR3RZ}
\end{figure*}

The scores were pre-processed by computing \acp{pcd}, as used for the identification of musical keys \citep{krumhansl1982,temperley2008,albrecht2014}, using the \texttt{pitchscapes} library \citep{lieckModelling2020}. We used a resolution of 70 equally spaced time slices per piece, resulting in sequences of 12-dimensional categorical distributions.
The tree shown in \figref{V9Y7WXs1}(c) for Johann Sebastian Bach's Prelude No.~1 in C major, BWV 846, corresponds to a harmonic expert analysis performed by the authors. \figref{hYxyR3RZ} shows the annotated tree with additional chord labels, which are provided in a simplified notation commonly used in Jazz lead sheets to be more accessible to a broad audience. In \tabref{results}, we list the results for all 24 preludes. For a better interpretation of the model and the presented results, there are two relevant points to consider.

\subsubsection{Chromatic versus Diatonic Transposition}

It is interesting to look in more detail at what musical aspects the model can or cannot represent. In a nutshell, it \emph{can} represent chromatic transposition but \emph{cannot} represent diatonic transposition, which has a number of consequences, as described in the following.

The transpositions of the left child perform a cyclic rotation of the corresponding probabilities in the pitch-class distribution represented by the latent variable, which corresponds to a \emph{chromatic} transposition. This determines not only which pitch classes have a significant probability to occur (the in-scale tones) but also the specific weights. For instance, the tonic and fifth scale degree typically have the highest weights. A transposition by 5 or 7 semitones from a current major key (say C major) thus corresponds to a modulation to the sub-dominant (F major) or dominant (G major) key, respectively. This \emph{includes} adaptation of the fourth and seventh scale degree of the target key, respectively (B$\rightarrow$B$\flat$ for F major; F$\rightarrow$F$\sharp$ for G major), as well as the correct assignment of strong weights to the tonic and fifth scale degree.

However, \emph{diatonic} transposition cannot be represented in this way. For instance, to represent a modulation from C major to A minor, the model has two options that are both far from optimal. 1)~It can choose not to apply a chromatic transposition, which ensures that all in-scale tones are correctly represented (i.e.~they have significant weight). This, however, means that the relative weights are not appropriate for A minor. In particular, the strong weights on the tonic and fifth scale degree are not present and, instead, the third and seventh scale degree (C and G, the former tonic and fifth scale degree) have disproportionally strong weight. Correcting these weights has to occur through the Gaussian transitions, which can only be explained with a relatively high transition variance. 2)~The second option would be to perform a chromatic transposition by 9 semitones, which ensures that the strongest weights remain on the tonic and fifth scale degree of the new key. However, three out-of-scale tones (C$\sharp$, F$\sharp$, G$\sharp$) now have a high weight, while the respective in-scale tones do not. Again, this has to be corrected for by the Gaussian transition noise at a potentially even higher cost than in the first case.

This is a highly plausible explanation for why we only see non-zero weights for the identity and (chromatic) transposition by a fifth in our experiments. Any diatonic modulations are best explained by reweighting using via Gaussian transition noise without a transposition, rather than by a chromatic transposition, which would require an even stronger reweighting (except for modulation to the sub-dominant and dominant key, which can be appropriately explained by a chromatic transposition).

\subsubsection{Chord Labels}

It is important to note that the chord labels in the expert annotation convey significantly more information than just what pitch classes can be expected to occur in the respective section. For example, the very same pitch-class distribution of G--C--E could amongst others be labeled as a C major chord in second inversion, a G major chord with 64-suspension (Gsus64), or an A minor seventh chord with omitted root, which might be easily confused by a musically untrained annotator. Which of these labels is correct depends in many cases on the context, such as how a chord resolves to the next one. While these differences are important from a musical perspective (they express a different experience of the same musical events), our model was trained to only predict pitch-class distributions. Therefore, in its current state, it cannot reproduce these distinctions, but we expect future versions to significantly improve in this respect.

{
  \def\height{0.25\linewidth}
  \newcommand*{\pieceentry}[4]{%
    Johann Sebastian Bach&#1&#2&#3\\\nopagebreak
    \multicolumn{4}{c}{%
      \includegraphics[height=\height]{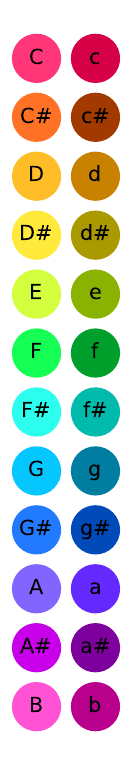}%
      \includegraphics[height=\height]{./figures/\detokenize{#4}}%
    }\\
    \otoprule%
  }
  \begin{longtable}[c]{llll}
    \caption{Results for all major preludes in Johann Sebastian Bach's ``Wohltemperiertes Klavier I \& II''. \textbf{(left):}~Expected value of the latent variables, i.e.~the mean of \eqref{eq:NuTG01QR}, colour-coded using a key-finding algorithm from the \texttt{pitchscapes} library \citep{lieckModelling2020}. \textbf{(right):}~Marginal node probability, i.e.~the normalisation of \eqref{eq:NuTG01QR} as well as the \ac{rbn} tree estimate.\label{tab:results}}\\

    \toprule
    \endfirsthead

    \toprule
    \endhead

    \multicolumn{4}{c}{(continued on next page)}\\
    \bottomrule
    \endfoot

    \multicolumn{4}{c}{End of \tabref{results}}\\
    \bottomrule\bottomrule
    \endlastfoot

    \pieceentry{Prelude No.~1 in C major}{BWV 846}{Wohltemperiertes Klavier I}{210606-Prelude_No_1_BWV_846_in_C_Major} \\
    \pieceentry{Prelude No.~3 in C$\sharp$ major}{BWV 848}{Wohltemperiertes Klavier I}{242986-Prelude_No_3_BWV_848_in_C_Major} \\
    \pieceentry{Prelude No.~5 in D major}{BWV 850}{Wohltemperiertes Klavier I}{274576-Prelude_No_5_BWV_850_in_D_Major} \\
    \pieceentry{Prelude No.~7 in E$\flat$ major}{BWV 852}{Wohltemperiertes Klavier I}{316381-Prelude_No_7_BWV_852_in_E_Major} \\
    \pieceentry{Prelude No.~9 in E major}{BWV 854}{Wohltemperiertes Klavier I}{366016-Prelude_No_9_BWV_854_in_E_Major} \\
    \pieceentry{Prelude No.~11 in F major}{BWV 856}{Wohltemperiertes Klavier I}{420466-Prelude_No_11_BWV_856_in_F_Major} \\
    \pieceentry{Prelude No.~13 in F$\sharp$ major}{BWV 858}{Wohltemperiertes Klavier I}{480961-Prelude_No_13_BWV_858_in_F_Major} \\
    \pieceentry{Prelude No.~15 in G major}{BWV 860}{Wohltemperiertes Klavier I}{535841-Prelude_No_15_BWV_860_in_G_Major} \\
    \pieceentry{Prelude No.~17 in A$\flat$ major}{BWV 862}{Wohltemperiertes Klavier I}{598056-Prelude_No_17_BWV_862_in_A_Major} \\
    \pieceentry{Prelude No.~19 in A major}{BWV 864}{Wohltemperiertes Klavier I}{667506-Prelude_No_19_BWV_864_in_A_Major} \\
    \pieceentry{Prelude No.~21 in B$\flat$ major}{BWV 866}{Wohltemperiertes Klavier I}{734101-Prelude_No_21_BWV_866_in_B_Major} \\
    \pieceentry{Prelude No.~23 in B major}{BWV 868}{Wohltemperiertes Klavier I}{824336-Prelude_No_23_BWV_868_in_B_Major} \\
    \pieceentry{Prelude No.~1 in C major}{BWV 870}{Wohltemperiertes Klavier II}{897876-Prelude_No_1_BWV_870_in_C_Major} \\
    \pieceentry{Prelude No.~3 in C$\sharp$ major}{BWV 872}{Wohltemperiertes Klavier II}{971581-Prelude_No_3_BWV_872_in_C_Major} \\
    \pieceentry{Prelude No.~5 in D major}{BWV 874}{Wohltemperiertes Klavier II}{1066276-Prelude_No_5_BWV_874_in_D_Major} \\
    \pieceentry{Prelude No.~7 in E$\flat$ major}{BWV 876}{Wohltemperiertes Klavier II}{1133561-Prelude_No_7_BWV_876_in_E_Major} \\
    \pieceentry{Prelude No.~9 in E major}{BWV 878}{Wohltemperiertes Klavier II}{1207206-Prelude_No_9_BWV_878_in_E_Major} \\
    \pieceentry{Prelude No.~11 in F major}{BWV 880}{Wohltemperiertes Klavier II}{1293351-Prelude_No_11_BWV_880_in_F_Major} \\
    \pieceentry{Prelude No.~13 in F$\sharp$ major}{BWV 882}{Wohltemperiertes Klavier II}{1392041-Prelude_No_13_BWV_882_in_F_Major} \\
    \pieceentry{Prelude No.~15 in G major}{BWV 884}{Wohltemperiertes Klavier II}{1497201-Prelude_No_15_BWV_884_in_G_Major} \\
    \pieceentry{Prelude No.~17 in A$\flat$ major}{BWV 886}{Wohltemperiertes Klavier II}{1593551-Prelude_No_17_BWV_886_in_A_Major} \\
    \pieceentry{Prelude No.~19 in A major}{BWV 888}{Wohltemperiertes Klavier II}{1704556-Prelude_No_19_BWV_888_in_A_Major} \\
    \pieceentry{Prelude No.~21 in B$\flat$ major}{BWV 890}{Wohltemperiertes Klavier II}{1822451-Prelude_No_21_BWV_890_in_B_Major} \\
    \pieceentry{Prelude No.~23 in B major}{BWV 892}{Wohltemperiertes Klavier II}{1938106-Prelude_No_23_BWV_892_in_B_Major}
  \end{longtable}
}

\end{document}